\pdfoutput=1
\documentclass[10pt]{article}
\usepackage[text={5.5in,9in},centering]{geometry}
\usepackage[T1]{fontenc}
\usepackage{times}
\usepackage{microtype}
\usepackage{graphicx}
\usepackage[round,authoryear]{natbib}
\usepackage{amsmath,amssymb,amsthm,mathtools}
\usepackage{booktabs}
\usepackage{tabularx}
\usepackage{multirow}
\usepackage{xcolor}
\usepackage{xurl}
\usepackage{listings}
\usepackage{tcolorbox}\tcbuselibrary{listings,skins}
\usepackage{capt-of}
\lstdefinestyle{promptbase}{basicstyle=\scriptsize\ttfamily, breaklines=true, columns=fullflexible, frame=none, backgroundcolor={}, aboveskip=0pt, belowskip=0pt, moredelim=[is][\color{black!55}\itshape]{|}{|}}
\lstdefinestyle{prompt-cpol}{style=promptbase, moredelim=[s][\color{cpol}\bfseries]{\{}{\}}}
\lstdefinestyle{prompt-cwm}{style=promptbase, moredelim=[s][\color{cwm!80!black}\bfseries]{\{}{\}}}
\newtcblisting{promptbox}[2]{listing only, enhanced, colback=#1!3, colframe=#1!72!black, boxrule=0.6pt, arc=2pt,
  title={#2}, fonttitle=\sffamily\bfseries\small, coltitle=white, colbacktitle=#1!78!black, toptitle=1pt, bottomtitle=1pt,
  left=5pt, right=5pt, top=2pt, bottom=2pt, before skip=6pt, after skip=6pt, listing options={style=prompt-#1}}
\newcommand{\sigup}{\textsuperscript{\fontsize{6.5}{7}\selectfont$\dagger$}}\newcommand{\sigdn}{\textsuperscript{\fontsize{6.5}{7}\selectfont$\ddagger$}}
\usepackage{tikz}
\usetikzlibrary{arrows.meta,positioning,fit,calc,shapes.geometric,backgrounds}
\usepackage{algorithm}
\usepackage{algorithmic}

\newcommand{\wmppo}{WM-PPO}
\newcommand{\qppo}{Q-PPO}
\newcommand{\sopo}{SO-PPO}
\newcommand{\steppo}{Turn-ratio PPO}
\definecolor{cpol}{RGB}{31,90,166}
\definecolor{cwm}{RGB}{77,154,77}
\definecolor{ccri}{RGB}{224,130,20}
\definecolor{creal}{RGB}{90,90,90}

\newcommand{\V}{V_\psi}
\newcommand{\pol}{\pi_\theta}
\newcommand{\wm}{p_\phi}
\newcommand{\qhat}{\hat q}

\usepackage[hidelinks]{hyperref}   %
\hypersetup{pdftitle={Learning to Act with World Models by Valuing What Comes Next}, pdfauthor={Chenghao Gao, Zhi Zhang, Dian Jin, Yihe Pan, Doudou Zhou}}
\usepackage[section,above]{placeins}   %
\makeatletter\let\wmppo@acc@orig\@addtocurcol\makeatother
\usepackage{flafter}
\makeatletter\let\wmppo@acc@fl\@addtocurcol\let\@addtocurcol\wmppo@acc@orig
\newcommand{\appendixfloatsafter}{\let\@addtocurcol\wmppo@acc@fl}\makeatother
\title{\textbf{Learning to Act with World Models by Valuing What Comes Next}}
\usepackage{authblk}

\author[1,*]{Chenghao Gao}
\author[2,*]{Zhi Zhang}
\author[1]{Dian Jin}
\author[3]{Yihe Pan}
\author[1,{}\textdagger]{Doudou Zhou}
\affil[1]{Department of Statistics and Data Science, National University of Singapore}
\affil[2]{Department of Statistics and Data Science, University of California, Los Angeles}
\affil[3]{Department of Data Science, Fudan University}
\affil[*]{Equal contribution}
\affil[{}\textdagger]{Corresponding author: \texttt{doudouzhou@nus.edu.sg}}
\date{}

\begin{document}
\maketitle
\begin{abstract}
A sparse terminal reward tells an agent little about its steps. Much of the missing information is implicit in the environment's dynamics, because each state transition can change the conditions for success long before the reward arrives. A world model trained on real transitions can learn to predict what each action changes, but a predicted observation does not by itself show whether success has become more likely. With a binary terminal reward, PPO's critic makes this judgment by estimating the current policy's probability of success, but only for states the agent reaches. Our method, \wmppo{}, elegantly brings the two together. For each action the policy samples, executed or not, the world model predicts the next observation and the critic values the predicted successor. In the policy channel, comparisons among these values give the policy auxiliary advantages; in the value channel, their mean gives the critic an auxiliary target. A shared gate weights both channels according to how reliably comparisons with similar margins predicted real outcomes earlier in training. The policy, critic and world model all learn from the agent's own interaction, with no teacher or privileged information, and only the policy is used at inference. With a comparable number of real environment steps, \wmppo{} raises success from PPO's 63.5\% to 87.5\% on ALFWorld and from 39.3\% to 52.9\% on ScienceWorld. It matches the strongest PPO refinements we tested on ALFWorld and exceeds the best of them by 6.5 points on ScienceWorld. The trained policy more often completes prerequisite steps before placing an object or committing to an answer.
\end{abstract}

\section{Introduction}
\label{sec:intro}

Agents trained with reinforcement learning on multi-step tasks often receive only sparse terminal rewards, which provide limited feedback for learning. Yet each state transition can change the conditions for success long before the reward arrives. A needed object is found, an operation's precondition is met, or a command is rejected, and the next observations record all these changes. When tasked with cleaning a plate and putting it in a cabinet, for example, an agent may find the plate and pick it up, but put it away without first cleaning it at the sink. The final reward marks the whole attempt as a failure. It does not say which steps helped, or which other action would have left the agent better placed to succeed.

World models learn environment dynamics by predicting how an agent's actions change the environment. What a predicted change means for success depends on the task and on what the current policy can accomplish from the resulting state. In text environments, an action's contribution to success is not necessarily reflected in how much the observation text changes. A transition that produces a very different textual observation may leave the chances of success unchanged: opening a cabinet, for example, may reveal a long list of objects unrelated to the goal. Conversely, a small change in the environment may establish a critical prerequisite or make the goal unreachable. Heating a mug, for instance, may produce only a brief confirmation, yet satisfy a necessary condition for a task that requires a hot mug. We therefore evaluate predicted transitions by their effect on eventual success.

Fortunately, PPO~\citep{ppo} already learns such a valuation through its critic. The critic estimates the expected cumulative reward from a state under the current policy, and its evaluation evolves as the policy learns. When agents must learn from sparse terminal success rewards, the critic offers feedback before the task outcome is known. Its values estimate the probability of eventual success, and changes across a transition reveal how an action affects that probability. While PPO uses these estimates only through the advantages of executed actions, we apply the critic to the next states predicted by a world model for the policy's candidate actions (Figure~\ref{fig:pipeline}). These estimates, which we call \emph{transition values}, let us compare the policy's candidates through their predicted consequences. Diagnostics on ALFWorld~\citep{alfworld} support this idea: selecting among the policy's candidates using these values achieves higher success than direct policy sampling (Section~\ref{sec:diagnostics}).

We therefore introduce \wmppo{}, which incorporates this feedback through policy and value channels. At each state along the main training trajectory, the policy samples several candidate actions and executes the first as usual. The policy channel turns candidate comparisons into auxiliary advantages in PPO's clipped objective, while the value channel uses the mean transition value as an additional one-step target for the critic at the current state. These critic updates also affect the advantages used in subsequent policy updates. Since value estimation in LLM reasoning can be biased and yield unreliable comparisons between candidate steps~\citep{vcppo,vineppo}, \wmppo{} calibrates both channels against real outcomes. For each training task, it collects one main episode and three branches, each trying a different action at a randomly chosen step. A shared gate uses earlier branch outcomes to weight both channels according to the margin between transition values (Section~\ref{sec:method}). The design stays elegant: the world model is the only learned component added to PPO and reuses the real transitions collected during training. Real returns remain the critic's main supervision. Training requires no teacher model or privileged information, and the policy acts alone at inference.

Prior work also draws on signals beyond terminal rewards to improve language agents (Table~\ref{tab:related} in Section~\ref{sec:related}). Some methods add observation-prediction losses during RL~\citep{echo,paw,dgswm}, while others learn environment dynamics before downstream policy training~\citep{earlyexp,spaiwm,rwml}. Distillation methods obtain finer feedback from teacher distributions or from models given hindsight observations or reference solutions~\citep{gkd,ocsd,vaspo}. WMA, ITP and CoMAP also pair actions with predicted consequences, but use them at decision time: the predictions select, inform or revise the next action, so the world model remains part of the deployed agent~\citep{wma,itp,comap}. In WMA, the policy stays frozen, and the comparisons guide each decision rather than training the policy. Internalizing the Future~\citep{itf} learns from teacher-generated foresight traces. Its policy expresses confidence in an imagined plan before generating an action, and this confidence is calibrated against task outcomes to construct step-level advantages. ProAct's MC-GRPO~\citep{proact} compares actions sampled at the same state using values estimated from simulator rollouts. \wmppo{} instead evaluates candidates through their predicted consequences using a world model and PPO's existing critic. Both networks learn from the agent's own interactions, turning predicted state changes into feedback for policy and value learning. The policy thus learns from this information during training, and no world model is needed at inference.

We evaluate \wmppo{} on the simpler ALFWorld and on the harder ScienceWorld~\citep{scienceworld}, whose tasks often require establishing intermediate conditions before the agent commits to an answer. The comparisons test whether learning from predicted consequences improves success for a comparable amount of real interaction (Section~\ref{sec:experiments}). They cover PPO, its multi-turn refinements and group-based methods; structural controls that change one component at a time; and auxiliary supervision from observation prediction, agreement and teachers. The main table compares 17 methods and variants, and the interaction-efficiency analysis places \wmppo{} among 23 other methods and settings on ALFWorld and 20 on ScienceWorld. We then analyze how the two channels affect learning, and how the trained policy advances a task (Section~\ref{sec:understanding}).

Our contributions are threefold:
\begin{itemize}
\item \textbf{Learning from dynamics by valuing predicted consequences.} We show that valuing world-model predictions with PPO's critic turns information implicit in environment dynamics into feedback that supplements sparse terminal rewards. Diagnostics demonstrate that these transition values support both action selection and policy improvement.
\item \textbf{Calibrating value-based feedback with real outcomes.} To account for noisy critic estimates, we use outcomes of real branches to calibrate a shared gate that weights both channels by how often comparisons with similar margins agreed with real outcomes earlier in training. The critic's values thus reach both the policy and the critic itself as a stable signal grounded in real outcomes.
\item \textbf{\wmppo{}: transition values for policy and critic learning.} We introduce two complementary learning channels: comparisons among transition values provide auxiliary policy advantages, while their mean supplies an auxiliary critic target. Only the policy is needed at inference. On ALFWorld and ScienceWorld, \wmppo{} reaches PPO's success levels with fewer real interaction steps and achieves higher success at comparable interaction budgets.
\end{itemize}

\section{From Predicted Consequences to Learning Signals}
\label{sec:diagnostics}

We first relate state transitions to changes in the probability of success, then test whether predicted consequences can guide a policy update.

\paragraph{Setting.} A task is a natural-language goal in a text environment. At step $t$ the agent is in state $s_t$, by which we mean the content of its prompt: the goal, its five most recent observations and actions, the current observation, and the actions or objects currently available. The policy $\pol$ samples a short response $y_t$, from which one action $a_t$ is parsed. The environment returns the next observation $o_{t+1}$, and the next state $s_{t+1}$ adds $a_t$ and $o_{t+1}$ to the history and refreshes the list of available actions or objects. The value of a state under the policy $\pi$ is the expected return that follows it,
\begin{equation}
V^\pi(s_t)=\mathbb E_\pi\Bigl[\,\sum_{k\ge 0}\gamma^{k}\,r_{t+k+1}\Bigm| s_t\Bigr],
\label{eq:value}
\end{equation}
where $\gamma$ is the discount factor, $r$ denotes reward, and we treat $s_t$ as sufficient for predicting returns. In the environments we study, the only reward is the final reward $R\in\{0,1\}$, given at the last step: 1 if the task was completed within 40 steps and 0 otherwise. With $\gamma=1$, the value is then the probability of eventually succeeding,
\begin{equation}
V^\pi(s_t)=\mathbb E_\pi\bigl[R\bigm| s_t\bigr]=\Pr\nolimits_\pi\bigl(\text{success}\bigm| s_t\bigr),
\label{eq:value-prob}
\end{equation}
and we write $V^\pi(s)=R$ for a state that ends the episode. The superscript marks that these values belong to the current policy $\pi$, so they can change as the policy learns. An intermediate transition from $s_t$ to $s_{t+1}$ receives no reward, yet it changes this probability by $V^\pi(s_{t+1})-V^\pi(s_t)$. The observation records what changed in the environment; the difference in state values measures how that change affects the current policy's probability of success. The probability of success after an action is $Q^\pi(s_t,a)=\mathbb E[V^\pi(s_{t+1})\mid s_t,a_t=a]$, and the action's advantage is the expected change in state value caused by the action,
\begin{equation}
A^\pi(s_t,a)=Q^\pi(s_t,a)-V^\pi(s_t)=\mathbb E\bigl[V^\pi(s_{t+1})-V^\pi(s_t)\bigm| s_t,a_t=a\bigr].
\label{eq:adv-transition}
\end{equation}
At a fixed state, the difference between two actions' advantages is therefore the difference between their expected success probabilities. PPO estimates this advantage only for the executed action, through the critic's values along the executed path (Appendix~\ref{app:gae}). Estimating it for the alternatives requires their transitions, which a world model can predict.

The critic $\V(s_t)$ estimates $V^\pi(s_t)$, and a world model $\wm(o_{t+1}\mid x_t, a_t)$ predicts the next observation from its own view $x_t$ of the history (Section~\ref{sec:method-eval}). All models are LoRA adapters~\citep{lora} on Qwen3-4B-Instruct-2507~\citep{qwen3}. The diagnostics use ALFWorld~\citep{alfworld} and the policy and critic of a PPO run continued from the common warm-up of Section~\ref{sec:setup} to iteration 24 (53.6\% success on the 274 evaluation games); all evaluations are paired by game.

Equation~\eqref{eq:adv-transition} suggests three tests. The critic must order the successors of different actions by their probability of success; predicted successors must preserve these differences for actions that were not taken; and updates built on the comparisons must improve the policy when it acts alone. We test these conditions before introducing the online method.

\subsection{Does the critic see the value of a state transition?}
\label{sec:diag-real}

\begin{figure}[t]
\centering
\includegraphics[width=\linewidth]{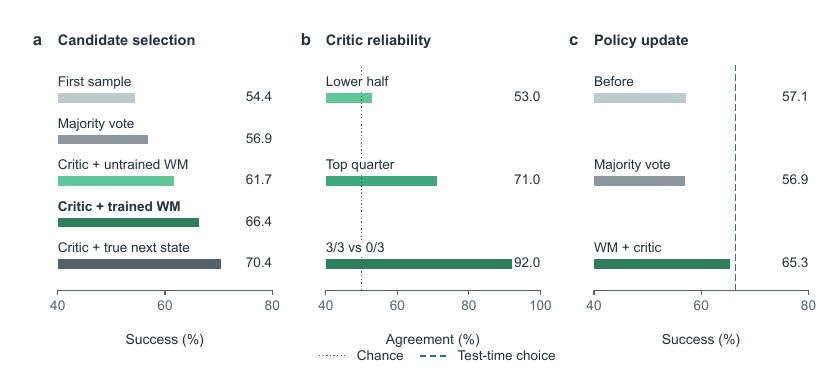}
\caption{\textbf{Valuing successors and learning from their comparisons on ALFWorld} (PPO policy and critic after 24 iterations; 274 evaluation games). \textbf{(a)}~Success when the agent executes one of four sampled candidates, chosen by majority vote or by the critic's value of the candidate's next state; the next state is predicted by an untrained or a trained world model, or obtained from the environment. \textbf{(b)}~Agreement of the critic's ordering with estimated success probabilities for candidate pairs at the same state (195 pairs), split by the gap between the critic's two values. The last bar contains the pairs in which one candidate succeeded in all three continuations and the other in none; dotted line: chance. \textbf{(c)}~Policy-only success after one offline PPO-clip epoch whose labels come from the world model and the critic or from majority vote (temperature 1, two episodes per game); dashed line: test-time selection with the trained world model from (a). Horizontal axes start at 40\%.}
\label{fig:diagnostics}
\end{figure}

We first test whether the critic can compare actions at the same state by valuing their real successors. At every step, the policy samples four candidate actions. Each distinct candidate is executed for one step in a copy of the environment, and the critic values each resulting next state. The agent then executes the candidate with the highest-valued successor, which is expected to leave the greatest chance of success (Equation~\eqref{eq:adv-transition}). The policy is held fixed.

Selecting by the critic's value of the real successor raises success from 56.9\% (majority vote over the four candidates) to 70.4\%, a gain of 13.5 points (SE 2.8, $p<10^{-4}$; Figure~\ref{fig:diagnostics}a); executing a single sampled candidate gives 54.4\%. Most of this gain comes from valuing intermediate successors: a control that selects a candidate only when it completes the task immediately, and otherwise uses majority vote, reaches 59.5\%. The critic therefore carries information about intermediate transitions that the agent's own choices do not use.

The reliability of this ranking depends on the gap between the critic's values. Prior work reports noisy and initially biased value estimates that rank candidate next steps near chance~\citep{vineppo,vcppo,vapo}. At randomly chosen states, we executed each distinct candidate and estimated its success probability from three continuations. Of the 195 candidate pairs whose estimates differ, the critic orders 57\% correctly, and the median gap between its two values is 0.008. Ranking accuracy increases from 53\% in the lower half of the critic's value gaps to 71\% in the top quarter. Among the 12 pairs for which one candidate succeeded in all three continuations and the other in none, the critic ranked 11 correctly (Figure~\ref{fig:diagnostics}b). Large value gaps identify the comparisons the critic orders most reliably. The reliability gate uses real outcomes to determine how much weight each margin should receive (Section~\ref{sec:method-gate}).

\subsection{Do predicted state transitions preserve this value?}
\label{sec:diag-pred}

We next test whether learned dynamics make these values available for candidates that are not executed. A world model replaces the one-step execution: a separate LoRA adapter on the same base model, trained with a next-observation loss on transitions that the same policy generated on 1,000 training games. For this diagnostic only, its data also include one-step executions of candidates the policy did not choose (48k transitions, 44\% of them for such candidates); the world model of \wmppo{} is trained on executed transitions alone. At each step, the world model greedily predicts the next observation of every distinct candidate, and the critic values each predicted next state. The agent selects the candidate with the highest value and executes it in the environment; the other candidates are never executed.

Choosing by predicted consequences reaches 66.4\%, 9.5 points above majority vote (SE 2.5, $p<0.001$; Figure~\ref{fig:diagnostics}a) and close to the 70.4\% obtained with the true consequences. With the untrained base model as the world model, selection reaches 61.7\%; the trained model is 4.7 points higher. Two controls examine why selection with true consequences performs better. When the true next observation is combined with the action list from before the step, as in the world model's successor, selection reaches 65.3\%. When newly revealed objects are removed from the true observation, it reaches 59.1\% (Appendix~\ref{app:diag}). Both the refreshed action list and the newly revealed content contribute to the gain. The predictions are exact for 99.7\% of the held-out observations that follow from what the agent has already seen, but for only 26\% of those that reveal objects it has not seen; for the latter, they still give a prior over what will be found (Appendix~\ref{app:diag}). In this diagnostic, predicted transitions preserve most of the selection gain obtained with real ones.

\subsection{Can the policy learn from these values?}
\label{sec:diag-learn}

The final test asks whether the comparisons produced by the world model and critic can improve the policy itself. Starting from the same checkpoint, the policy played 500 new training games, executing the first of four samples at each step. At the 8,042 decision points with at least two distinct candidates, the candidate $\hat a_t$ preferred by the world model and the critic served as a label, and each of the four samples $a_t^{(1)},\dots,a_t^{(K)}$ ($K=4$) received the advantage
\begin{equation}
\tilde A_{t,k} = z_{t,k}-\bar z_t,\qquad z_{t,k}=\mathbf 1\bigl[a_t^{(k)} = \hat a_t\bigr],\qquad \bar z_t=\frac{1}{K}\sum_{k'=1}^{K}z_{t,k'}.
\label{eq:offline-adv}
\end{equation}
These advantages sum to zero across the samples at each decision point. One epoch of PPO-clip on these samples used no task reward. The updated policy was then evaluated alone, at temperature 1, twice on the 274 games.

The update raises policy-only success from 57.1\% to 65.3\% (+8.2 points, SE 2.0, $p=0.0003$; Figure~\ref{fig:diagnostics}c). Using majority-vote labels on the same data for the same number of updates gives 56.9\%, no change from 57.1\%, although the two sets of labels agree at 44\% of the decision points. With the data and update count held fixed, the consequence-based labels supply the information that improves the policy. The gain persists when the updated policy acts alone, without the world model or the critic.

\paragraph{From diagnostics to an online method.} Starting from a fixed policy, the diagnostics show that the critic can compare actions by valuing their successors, that the world model extends this evaluation to unexecuted candidates, and that the policy learns from the resulting comparisons. Reliability depends on the value margin, so the comparisons must be calibrated against real outcomes. As the policy improves, it visits new states and proposes new actions, requiring the world model, critic and reliability estimates to be updated. \wmppo{} brings these steps into online training and uses the transition values for both candidate comparisons and critic targets.

\section{\wmppo{}}
\label{sec:method}

\begin{figure}[!tp]
\centering
\includegraphics[width=\linewidth]{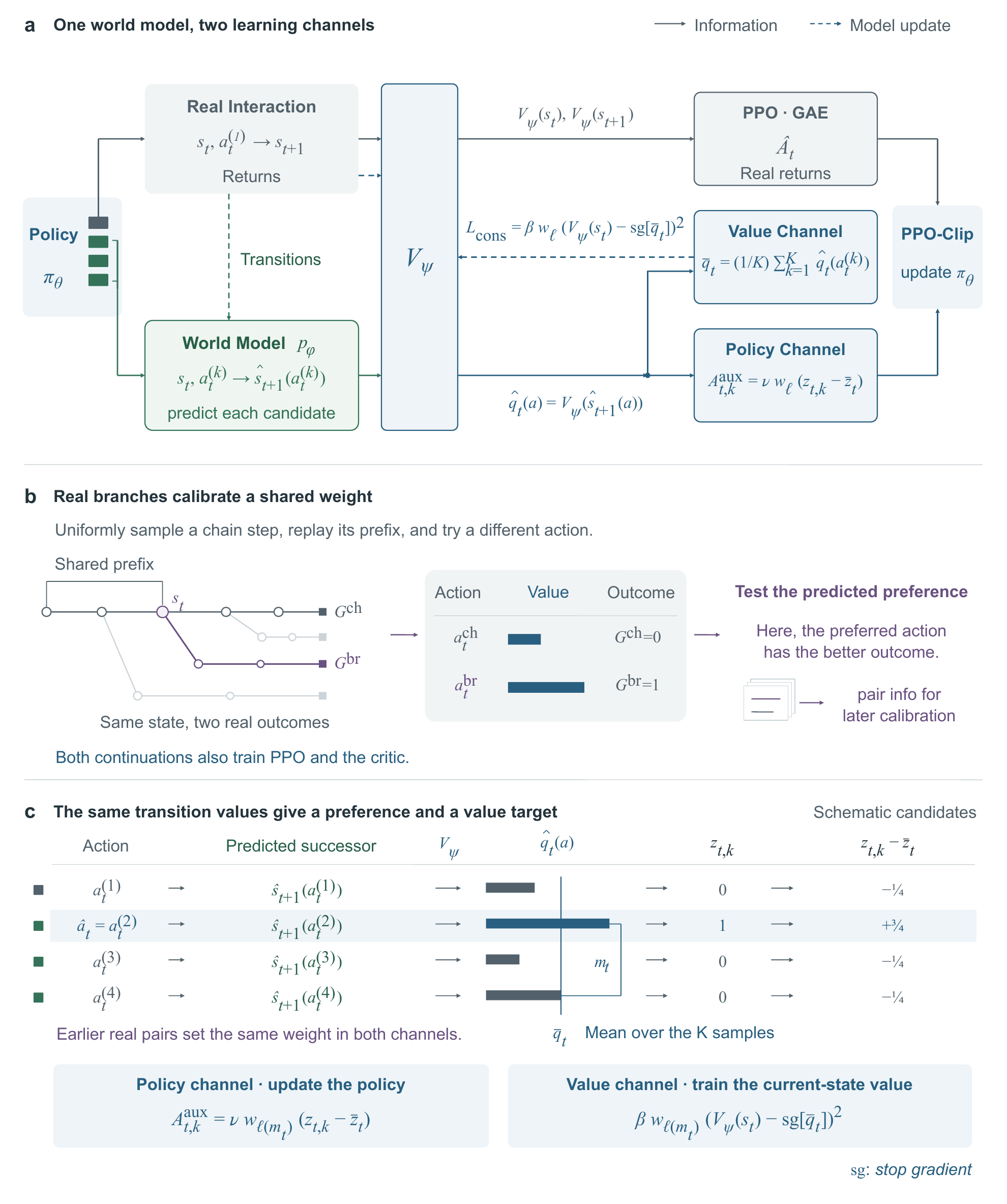}
\caption{\textbf{Transition values for policy and value learning.} \textbf{a},~Real state transitions train the world model; real returns are the main supervision for the critic. The policy samples $K$ responses and executes the first. The world model predicts each candidate's next observation. The same critic $\V$ values real states for PPO's advantages $\hat A_t$ and predicted successors for transition values $\qhat_t(a)$ (Equation~\eqref{eq:score}; 1 or 0 at predicted endings). Comparisons supply auxiliary advantages $A^{\mathrm{aux}}_{t,k}$ in PPO-clip (policy channel). The mean transition value $\bar q_t$ is held fixed as a target for $\V(s_t)$ during the critic update (value channel). Both use the weight $w_\ell$ of the state's margin bin $\ell(m_t)$. Solid arrows carry information, and dashed arrows update models. At evaluation, the policy acts alone. \textbf{b},~Replaying a prefix lets chain and branch actions be tested from the same state; their outcomes calibrate each margin bin's weight for both channels (Figure~\ref{fig:gate}). \textbf{c},~The preferred action $\hat a_t$ sets the indicators $z_{t,k}$, and the margin $m_t$ selects the bin whose weight scales both $A^{\mathrm{aux}}_{t,k}$ and the value-channel loss. Trees and values are schematic.}
\label{fig:pipeline}
\end{figure}

\wmppo{} adds one learned component to PPO: a world model $\wm$ trained on real state transitions to predict the next observation after an action. The policy $\pol$ keeps PPO's clipped objective, and the critic $\V$ learns primarily from real returns. At each training state, the world model predicts the consequences of the policy's candidate actions, and the critic values the predicted successors. These transition values enter learning through two channels. Comparisons among the candidates supply auxiliary advantages for the policy (the policy channel), and the mean transition value supplies a one-step target for the critic (the value channel). Small rollout trees provide real branch outcomes, and a gate uses earlier branches to weight both channels by the reliability of their value margins. Section~\ref{sec:method-eval} defines the evaluation, Sections~\ref{sec:method-update} and~\ref{sec:method-value} the two channels, Section~\ref{sec:method-loop} the training loop, and Section~\ref{sec:method-gate} the gate. Figure~\ref{fig:pipeline} gives an overview, and Algorithm~\ref{alg:wmppo} summarizes one iteration (Algorithm~\ref{alg:wmppo-full} in Appendix~\ref{app:alg} gives the full procedure).

\subsection{Evaluating predicted state transitions}
\label{sec:method-eval}

The state $s_t$ is what the agent sees in its prompt at step $t$: the task, the step counter, the five most recent observations and actions, the current observation, and the admissible actions (ALFWorld) or the objects that can be referred to (ScienceWorld). It is not the simulator's internal state, which no model sees. At a state $s_t$ reached during training, the policy samples $K$ responses $y_t^{(1)},\dots,y_t^{(K)}$ at temperature 1, and the actions $a_t^{(1)},\dots,a_t^{(K)}$ are parsed from them. The first action is executed, $a_t=a_t^{(1)}$ with response $y_t=y_t^{(1)}$, so collecting the candidates leaves the behavior policy unchanged. The success probability after an action and the value of a state (Equation~\eqref{eq:value-prob}) are related by
\begin{equation}
Q^\pi(s_t,a)=\mathbb E\bigl[V^\pi(s_{t+1})\bigm| s_t,a_t=a\bigr],\qquad
V^\pi(s_t)=\mathbb E_{a\sim\pi(\cdot\mid s_t)}\bigl[Q^\pi(s_t,a)\bigr],
\label{eq:bellman}
\end{equation}
where $V^\pi(s_{t+1})=R$ if the step ends the episode. When at least two candidates differ, the world model predicts the next observation of each distinct candidate $a$ (Figure~\ref{fig:pipeline}a), a successor state is assembled from the prediction, and the candidate is valued:
\begin{equation}
\begin{gathered}
\hat o_{t+1}(a) = \operatorname{greedy}\, \wm\bigl(\cdot \mid x_t, a\bigr),\qquad
\hat s_{t+1}(a) = C\bigl(s_t, a, \hat o_{t+1}(a)\bigr),\\
\qhat_t(a)=\begin{cases}1 & \text{if $\hat o_{t+1}(a)$ reports that the task is completed}\\
0 & \text{if $\hat o_{t+1}(a)$ reports that the episode has ended without completion}\\
\V\bigl(\hat s_{t+1}(a)\bigr) & \text{otherwise.}\end{cases}
\end{gathered}
\label{eq:score}
\end{equation}
We call $\qhat_t(a)$ the \emph{transition value} of $a$: the estimated value of the successor predicted for that action. It approximates $Q^\pi(s_t,a)$ in Equation~\eqref{eq:bellman} using one predicted next observation. The change in state value across a transition also subtracts the current-state value. A predicted ending gives its outcome directly, without a value estimate; otherwise the critic values the predicted successor. Only ScienceWorld's observations report the end of an episode, so in ALFWorld every transition value is a critic value.

The world model's context $x_t$ uses the same interaction history as $s_t$ with a different window: all previous observations and actions in ALFWorld, and the last eight actions without their observations in ScienceWorld. It also contains the task, the current observation and the admissible actions or objects. The prediction is one greedy decode of the next observation as the environment would print it. In ScienceWorld, this observation includes the room, the inventory, the game's progress score and, when the simulator ends the episode, a completion or failure marker.

The deterministic prompt constructor $C$ appends the candidate and the predicted observation to the history, advances the step counter, and updates the window of recent steps. The list of admissible actions or objects is carried over from before the step; only the observation is predicted. A real next state instead contains a refreshed list. Valuing an action through the state it leads to is the afterstate evaluation used when the effect of a move is known, as in backgammon~\citep{tdgammon,suttonbarto}; here the effect is predicted. Appendix~\ref{app:io} gives complete model inputs and outputs for one evaluated state (Figure~\ref{fig:io}).

Equation~\eqref{eq:bellman} underlies both channels of \wmppo{}. By the first relation, differences between the candidates' transition values estimate differences between their action values; the policy channel uses them to rank the candidates (Section~\ref{sec:method-update}). By the second, the mean transition value estimates the value of the state and serves as a critic target (Section~\ref{sec:method-value}). For two candidates at the same state, the difference between their advantages in Equation~\eqref{eq:adv-transition} equals the difference between the success probabilities after them,
\begin{equation}
A^\pi\bigl(s_t,a_t^{(k)}\bigr)-A^\pi\bigl(s_t,a_t^{(k')}\bigr)=Q^\pi\bigl(s_t,a_t^{(k)}\bigr)-Q^\pi\bigl(s_t,a_t^{(k')}\bigr)\approx \qhat_t\bigl(a_t^{(k)}\bigr)-\qhat_t\bigl(a_t^{(k')}\bigr).
\label{eq:delta}
\end{equation}
The common term $V^\pi(s_t)$ cancels, so the comparison does not depend on how well the critic values the current state. The approximation uses a single predicted observation in place of an expectation, a learned critic in place of $V^\pi$, and truncated prompts with the previous action list in place of full states. Section~\ref{sec:method-gate} checks the resulting comparisons against real outcomes. A preferred candidate need not increase the value in absolute terms; it only has to do better than the other candidates the policy proposed. The critic outputs an unconstrained scalar and is trained toward the probability of success; its estimates may fall outside $[0,1]$.

\subsection{From value comparisons to policy updates}
\label{sec:method-update}

Among the distinct sampled actions, the preferred candidate is $\hat a_t=\arg\max_a\qhat_t(a)$, whose predicted successor has the highest estimated probability of success. The margin $m_t=\qhat_t(\hat a_t)-\max_{a\neq\hat a_t}\qhat_t(a)$ measures how clearly it is preferred; a tie at the highest transition value ($m_t=0$) gives no preference. Each of the $K$ sampled responses receives an auxiliary advantage, which is the advantage $\tilde A_{t,k}$ of the diagnostic in Section~\ref{sec:diag-learn} (Equation~\eqref{eq:offline-adv}) scaled by a reliability weight:
\begin{equation}
A^{\mathrm{aux}}_{t,k}=\nu\, w_{\ell(m_t)}\bigl(z_{t,k}-\bar z_t\bigr),\qquad
z_{t,k}=\mathbf 1\bigl[a_t^{(k)}=\hat a_t\bigr],\qquad \bar z_t=\frac1K\sum_{k'=1}^{K}z_{t,k'},
\label{eq:aux}
\end{equation}
where $\ell(m_t)$ is the margin bin into which $m_t$ falls, $w_\ell\in[0,1]$ is that bin's reliability weight (Section~\ref{sec:method-gate}), and $\nu$ sets the strength of the auxiliary term relative to the PPO term.

Duplicate actions share one prediction, but advantages are assigned to all $K$ responses, duplicates included. The samples of the preferred candidate receive the positive advantage $\nu\,w_{\ell(m_t)}(1-\bar z_t)$, and the other samples the negative advantage $-\nu\,w_{\ell(m_t)}\bar z_t$; the weight scales both. Because the advantages sum to zero within the state (Figure~\ref{fig:pipeline}c), the update moves probability from the other sampled responses toward those of the preferred candidate. At the start of an update, when every probability ratio is 1 and no clipping applies, the auxiliary advantages give the gradient direction
\begin{equation}
\sum_{k=1}^{K}\bigl(z_{t,k}-\bar z_t\bigr)\nabla_\theta\log\pol\bigl(y_t^{(k)}\mid s_t\bigr)=\frac1K\sum_{k:\,z_{t,k}=1}\ \sum_{k':\,z_{t,k'}=0}\nabla_\theta\log\frac{\pol\bigl(y_t^{(k)}\mid s_t\bigr)}{\pol\bigl(y_t^{(k')}\mid s_t\bigr)},
\label{eq:aux-grad}
\end{equation}
scaled by $\nu\,w_{\ell(m_t)}$ and the token normalization of Equation~\eqref{eq:loss}. Each term raises the log-probability of a response with the preferred action relative to another sampled response at the same state, a pairwise preference. The size of the margin enters only through the bin's weight: the policy learns which candidate is preferred, not by how much. Equation~\eqref{eq:aux} is the group-relative advantage of GRPO~\citep{grpo} without division by the standard deviation, computed among the samples of one state with a label of 1 for the preferred candidate. Free-text actions cannot be enumerated, and PPO learns a state-value critic. We therefore evaluate the $K$ samples through their predicted successors and retain only which candidate is preferred (related methods in Section~\ref{sec:related}).

The policy is trained on two kinds of examples, each a (state, response, advantage) triple. Ordinary PPO examples $(s_t,y_t,\hat A_t)$ come from the executed steps. We compute their advantages by tree GAE on real returns (Section~\ref{sec:method-loop}), which gives raw estimates $A^{\mathrm{GAE}}_t$, and whiten these over the executed steps to obtain $\hat A_t$. Auxiliary examples $(s_t,y_t^{(k)},A^{\mathrm{aux}}_{t,k})$ come from the evaluated candidates; their advantages are given by Equation~\eqref{eq:aux} and are not whitened. A minibatch $\mathcal B$ containing both kinds of example enters one clipped objective:
\begin{equation}
\mathcal L_{\mathrm{PPO}}(\theta)=-\frac1{N}\sum_{(s,y,A)\in\mathcal B}\ \sum_{i=1}^{|y|}\min\!\Bigl(\rho_i A,\ \operatorname{clip}(\rho_i,1-\epsilon_{\mathrm{lo}},1+\epsilon_{\mathrm{hi}})A\Bigr),
\label{eq:loss}
\end{equation}
where $\rho_i=\pol(y_i\mid s,y_{<i})/\pi_{\mathrm{old}}(y_i\mid s,y_{<i})$ is the probability ratio of the $i$-th response token, $\pi_{\mathrm{old}}$ is the sampling policy, $N=\sum_{(s,y,A)\in\mathcal B}|y|$ is the number of response tokens in the minibatch, and $\epsilon_{\mathrm{lo}},\epsilon_{\mathrm{hi}}$ are the clipping bounds~\citep{dapo}. For a positive advantage, the clipping removes the incentive to raise a token's probability ratio above $1+\epsilon_{\mathrm{hi}}$; for a negative advantage, it removes the incentive to lower the ratio below $1-\epsilon_{\mathrm{lo}}$. An executed response can appear both as an ordinary example and as an auxiliary example. States whose bin has weight 0, and tied states, contribute no auxiliary examples. We use $K=4$, $\nu=1$ and $(\epsilon_{\mathrm{lo}},\epsilon_{\mathrm{hi}})=(0.2,0.28)$. The world model's text predictions and the critic's values are fixed before the update. Gradients reach the policy through its log-probabilities of the sampled responses.

\subsection{From predicted transitions to value targets}
\label{sec:method-value}

The value channel uses predicted consequences to estimate how likely the current state is to lead to success. By the second part of Equation~\eqref{eq:bellman}, the value of a state is the policy's average action value. The $K$ samples are draws from $\pol(\cdot\mid s_t)$ and $\qhat_t(a)$ estimates $Q^\pi(s_t,a)$, so their mean
\begin{equation}
\bar q_t=\frac1K\sum_{k=1}^{K}\qhat_t\bigl(a_t^{(k)}\bigr)
\label{eq:qbar}
\end{equation}
is a one-step model-based estimate of $V^\pi(s_t)$: the critic's own values one step ahead, averaged over the actions the policy proposes, with predicted endings valued at 1 or 0. This estimate supplies an auxiliary critic target, weighted by reliability,
\begin{equation}
\mathcal L_{\mathrm{cons}}(\psi)=\beta\sum_{s_t\in\mathcal D}w_{\ell(m_t)}\bigl(\V(s_t)-\operatorname{sg}[\bar q_t]\bigr)^2,
\label{eq:cons}
\end{equation}
where $\mathcal D$ is the set of evaluated chain states with $m_t>0$, $\operatorname{sg}$ stops the gradient, and $w_{\ell(m_t)}$ is the gate weight that also scales the state's auxiliary advantages; we use $\beta=0.5$. Like the transition values, the target is computed with the critic before the update. The term is added to the critic's regression on returns (Equation~\eqref{eq:critic}), which remains the critic's main supervision. The two targets average over different things. The return target $G(s_t)$ averages the final rewards of the few episodes that pass through $s_t$, all of which follow the executed actions, whereas $\bar q_t$ averages over the $K$ actions the policy proposed, each valued at its predicted successor. Where the world model predicts an ending, the target includes the predicted outcome directly, also for candidates that were not executed.

Both channels reuse the same predictions, critic and gate. Write the transition value of each sample as $\qhat_t(a_t^{(k)})=\bar q_t+\bigl(\qhat_t(a_t^{(k)})-\bar q_t\bigr)$. The value channel trains the critic on the mean, an estimate of how good the state is under the current policy. The policy channel uses the ordering of the deviations to form the gated preferences in Equation~\eqref{eq:aux}; both these deviations and the auxiliary advantages sum to zero over the $K$ samples. The value channel also affects the policy through the critic. PPO's advantages are built from critic-value differences along the executed path (Section~\ref{sec:method-loop}), so a critic that values states more accurately can give PPO's own update better advantages. This effect starts from the next iteration, because each iteration's advantages are computed before its critic update. We call the method with both channels \wmppo{}, and the variant without the term of Equation~\eqref{eq:cons} \wmppo{} (policy channel only).

\subsection{Learning through real interaction}
\label{sec:method-loop}

Each iteration uses one batch of branched real rollouts. Actions and returns train the policy and critic, observed transitions train the world model, and paired outcomes calibrate the gate (Figure~\ref{fig:pipeline}; Table~\ref{tab:roles} in Appendix~\ref{app:impl}).

\paragraph{Episodes and branches.} Each iteration draws 64 training tasks. For each task, the policy plays one main episode, the \emph{chain}, in which every step also records the $K$ candidates. The policy then plays three \emph{branches} from the chain. For each branch, we draw a step of the chain uniformly at random. Because the environments are deterministic, replaying the chain's actions up to that step restores its state. We resample the policy's first response at this state up to eight times, stopping when its action differs from the chain's, and then continue the branch to the end. If all eight samples repeat the chain's action, the branch continues with that action but yields no comparison. The policy proposed a different action within eight samples in 62--67\% of the branches on ScienceWorld and in 40--61\% on ALFWorld. Each iteration therefore plays 256 episodes. Branching from intermediate states to obtain real outcomes for different actions is common in tree-based RL for language models~\citep{treegrpo,treerl}. Here the branches add real interaction to the PPO update and test the candidate comparisons against real outcomes (Figure~\ref{fig:pipeline}b; Section~\ref{sec:method-gate}). The resampled action enters PPO as an ordinary sample with its unconditioned log-probabilities (Section~\ref{sec:discussion}).

\begin{figure}[t]
\centering
\includegraphics[width=\linewidth]{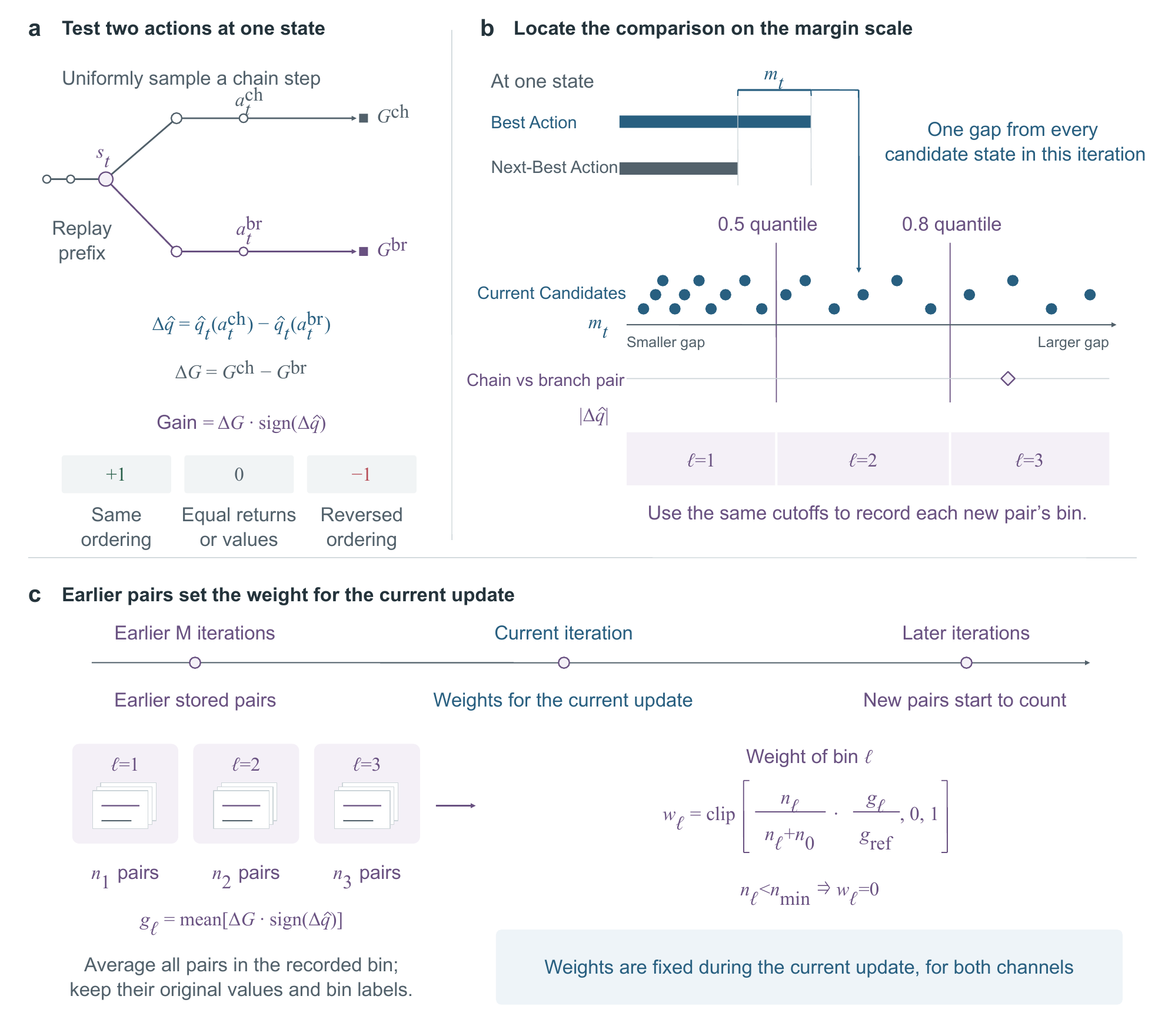}
\caption{\textbf{Calibrating transition-value comparisons with real outcomes.} \textbf{a},~A chain action and a different branch action are played from the same state; the pair's gain is $\Delta G\,\operatorname{sign}(\Delta\qhat)$ (Equation~\eqref{eq:gain}). \textbf{b},~The nonzero margins $m_t$ (best minus runner-up transition value) at the current iteration's evaluated states define two quantile cutoffs and thus three bins. Each new pair is assigned to a bin by $|\Delta\qhat|$. \textbf{c},~Pairs stored in the preceding $M$ iterations, with their recorded bins, determine $g_\ell$ and $w_\ell$ for the current iteration. Each current state's margin selects one weight for both its auxiliary advantages and its value-channel loss. New pairs are kept for later iterations. Dots and trajectories are schematic.}
\label{fig:gate}
\end{figure}

\paragraph{Critic.} The episodes of a task form a tree whose nodes are states and whose edges are executed steps; an edge shared by several episodes, such as the chain's prefix before a branch point, appears once. Let $\mathcal S$ be the set of nodes in the iteration's trees and $G(s)$ the mean final reward of the episodes that pass through node $s$. The critic minimizes
\begin{equation}
\mathcal L_V(\psi)=\sum_{s\in\mathcal S}\bigl(\V(s)-G(s)\bigr)^2+\mathcal L_{\mathrm{cons}}(\psi),
\label{eq:critic}
\end{equation}
that is, a regression on returns plus the value channel of Equation~\eqref{eq:cons}, which the policy-channel-only variant omits (Appendix~\ref{app:gae} gives implementation details of the regression). PPO's advantages are computed by GAE~\citep{gae} on the tree, with $\gamma=1$ and parameter $\lambda$. An edge $e$ from $s_t$ receives
\begin{equation}
A^{\mathrm{GAE}}(e)=\begin{cases}R-\V(s_t) & \text{if $e$ ends the episode,}\\[2pt]
\V(s_{t+1})-\V(s_t)+\lambda\,\overline A^{\mathrm{GAE}}(s_{t+1}) & \text{if $e$ leads to $s_{t+1}$,}\end{cases}
\label{eq:tree-gae}
\end{equation}
where $\overline A^{\mathrm{GAE}}(s)=|\mathcal E(s)|^{-1}\sum_{e'\in\mathcal E(s)}A^{\mathrm{GAE}}(e')$ is the mean raw advantage of the edges $\mathcal E(s)$ leaving $s$. On a path without branches, this is the usual GAE recursion. Its term $\V(s_{t+1})-\V(s_t)$ is the critic's estimate of the change in state value across the executed transition (Equation~\eqref{eq:adv-transition}). The executed step $t$ of an episode receives the raw advantage $A^{\mathrm{GAE}}_t$ of its edge. An episode cut at the step limit ends with $R=0$, without bootstrapping. We use $\lambda=0.95$.

\paragraph{World model.} The world model is trained on the state transitions that the policy executes in training tasks, with the cross-entropy of the next observation given the world model's context and the action,
\begin{equation}
\mathcal L_{\mathrm{WM}}(\phi)=-\frac1{N_o}\sum_{(x_t,a_t,o_{t+1})\in\mathcal B_{\mathrm{WM}}}\ \sum_{i=1}^{|o_{t+1}|}\log\wm\bigl(o_{t+1,i}\mid x_t,a_t,o_{t+1,<i}\bigr),
\label{eq:wm}
\end{equation}
where $\mathcal B_{\mathrm{WM}}$ is a minibatch of executed transitions, $o_{t+1,i}$ the $i$-th token of the observation the environment printed, and $N_o$ the number of observation tokens in the minibatch. The initial world model is trained at the end of iteration 12 on the executed transitions of the common PPO warm-up (Section~\ref{sec:setup}). Afterward, each iteration's executed transitions from chains and branches update it for one epoch, in parallel with the policy update; replayed prefix steps are not added again. Every recorded observation supervises what an action changed, whether or not the episode succeeded. This training reuses the interaction collected for RL and follows the states and actions of the current policy.

\subsection{Reliability calibration and the complete training loop}
\label{sec:method-gate}

Section~\ref{sec:diagnostics} shows that value margins carry information about comparison reliability. Both the policy-dependent values $V^\pi$ and the models estimating them change during training, so the gate recalibrates the margins in every iteration using recent real outcomes. The first step of every branch that obtained a different action provides a test (Figure~\ref{fig:gate}a). At the same state $s_t$, the chain took action $a_t^{\mathrm{ch}}$ and the branch took $a_t^{\mathrm{br}}\neq a_t^{\mathrm{ch}}$, and both continuations were played to the end, with returns $G^{\mathrm{ch}}$ and $G^{\mathrm{br}}$. Before the iteration's update, the world model and the critic value both actions as in Equation~\eqref{eq:score}. We write $\Delta\qhat=\qhat_t(a_t^{\mathrm{ch}})-\qhat_t(a_t^{\mathrm{br}})$ and $\Delta G=G^{\mathrm{ch}}-G^{\mathrm{br}}$. The pair's margin $|\Delta\qhat|$ assigns it to one of three bins $\ell=1,2,3$, ordered from the smallest to the largest margins, whose boundaries are the 0.5 and 0.8 quantiles of that iteration's nonzero candidate margins $m_t$ (Figure~\ref{fig:gate}b).

For bin $\ell$, the realized gain of following the preference is
\begin{equation}
g_\ell=\frac1{n_\ell}\sum_{\text{pairs in bin }\ell}\Delta G\,\operatorname{sign}(\Delta\qhat),
\label{eq:gain}
\end{equation}
computed over the $n_\ell$ pairs that the previous $M$ iterations stored in the bin, each with the value difference and bin recorded in its own iteration. The gain is positive when the preferred actions led to better real outcomes. Because final rewards lie in $\{0,1\}$, $\Delta G\in\{-1,0,1\}$, and a pair contributes only when its two outcomes differ. A bin thus earns weight when its comparisons often decide the task and its preferences are then usually right (Section~\ref{sec:valuing}).

\begin{algorithm}[t]
\caption{One iteration of \wmppo{} (simplified; Algorithm~\ref{alg:wmppo-full} in Appendix~\ref{app:alg} gives the full procedure). Predictions and transition values are computed without gradients.}
\label{alg:wmppo}
\small
\begin{algorithmic}[1]
\REQUIRE policy $\pol$, critic $\V$, world model $\wm$; branch pairs $\mathcal P$ of the last $M$ iterations
\STATE play 64 chains with $\pol$ (at every state $s_t$, sample $K$ responses and execute the first) and 3 real branches from each chain
\FOR{each chain state $s_t$ with at least two distinct candidate actions}
  \STATE value each distinct $a$ at its successor predicted by $\wm$: $\qhat_t(a)\leftarrow\V\bigl(\hat s_{t+1}(a)\bigr)$, or 1 / 0 at a predicted ending \hfill Eq.~\eqref{eq:score}
  \STATE compute the preferred candidate $\hat a_t$, margin $m_t$ and mean transition value $\bar q_t$ \hfill Eq.~\eqref{eq:qbar}
\ENDFOR
\STATE compute bin weights $w_\ell$ from the real gains $\Delta G\,\operatorname{sign}(\Delta\qhat)$ of the pairs in $\mathcal P$ \hfill Eqs.~\eqref{eq:gain}, \eqref{eq:gate}
\STATE policy: one PPO-clip objective, with $\hat A_t$ (whitened tree GAE on real returns) on executed responses and $A^{\mathrm{aux}}_{t,k}=\nu\,w_{\ell(m_t)}(z_{t,k}-\bar z_t)$ on all $K$ samples \hfill Eqs.~\eqref{eq:aux}, \eqref{eq:loss}
\STATE critic: regression on real returns plus $\beta\,w_{\ell(m_t)}\bigl(\V(s_t)-\operatorname{sg}[\bar q_t]\bigr)^2$ \hfill Eqs.~\eqref{eq:cons}, \eqref{eq:critic}
\STATE world model: next-observation loss on this iteration's executed transitions \hfill Eq.~\eqref{eq:wm}
\STATE store the new branch pairs with their pre-update values in $\mathcal P$; use the updated models next iteration
\end{algorithmic}
\end{algorithm}

The gate shrinks and rescales this gain to a weight capped at 1 (Figure~\ref{fig:gate}c),
\begin{equation}
w_\ell=\begin{cases}\operatorname{clip}\!\left(\dfrac{n_\ell}{n_\ell+n_0}\cdot\dfrac{g_\ell}{g_{\mathrm{ref}}},\,0,\,1\right), & n_\ell\ge n_{\min},\\[6pt] 0, & n_\ell<n_{\min}.\end{cases}
\label{eq:gate}
\end{equation}
Here $n_0$ sets how strongly a bin with few pairs is shrunk toward zero, $g_{\mathrm{ref}}$ is the shrunken gain at which a bin receives full weight, and $n_{\min}$ is the number of pairs below which a bin stays closed. A bin whose comparisons do not pay off in real outcomes thus contributes nothing. We use $M=4$, $n_0=40$, $g_{\mathrm{ref}}=0.05$ and $n_{\min}=20$; these values were set once from the first branch pairs of a preliminary ALFWorld run and kept unchanged in both environments. Pairs valued by the current pre-update models are stored for later iterations. The current weights depend on earlier branch outcomes.

All three models are separate LoRA adapters on one frozen base model and share no parameters beyond it. Only the policy is used at evaluation.

\section{Experimental Evaluation}
\label{sec:experiments}

\begin{table}[!t]
\centering\small
\caption{\textbf{Comparison protocol.} Columns show feedback beyond PPO's executed-step advantage, episodes per iteration (64 $\times$ (1 + 3): 64 tasks, each with one chain and three branches), and training-task coverage. Coverage is the share of distinct tasks seen in iterations 13--36 on ALFWorld (of 3,553 tasks) and 13--60 on ScienceWorld (of 1,228 variations). With 40 tasks $\times$ 4, only GRPO was run on ALFWorld. All methods start from the same warm-up checkpoint with the same task order.}
\label{tab:protocol}
\setlength{\tabcolsep}{3.4pt}
\begin{tabularx}{\linewidth}{@{}l l >{\raggedright\arraybackslash}X r r@{}}
\toprule
 & & & \multicolumn{2}{c}{Tasks seen} \\
\cmidrule(l){4-5}
Method & Episodes & Feedback beyond PPO's advantage & ALFWorld & ScienceWorld \\
\midrule
\wmppo{} & 64 $\times$ (1 + 3) & candidates valued by the critic at predicted successors; comparisons as auxiliary advantages, mean as critic target; both weighted by real branches & 43\% & 100\% \\
\midrule
PPO-160, PPO-256 & 160 / 256 & -- & 100\% & 100\% \\
GRPO, GiGPO & 20 $\times$ 8 & group-relative advantages instead of a critic & 14\% & 51\% \\
GRPO, GiGPO & 40 $\times$ 4 & as above, more tasks, fewer episodes each & 27\% & 91\% \\
CAPO, \sopo{}, \steppo{} & 160 & action- or turn-level ratios and clipping & 100\% & 100\% \\
\midrule
\wmppo{} (policy channel only) & 64 $\times$ (1 + 3) & as \wmppo{}, without the critic target & 43\% & 100\% \\
\qppo{} & 64 $\times$ (1 + 3) & as policy channel only; candidates valued by a learned $Q_\omega(s_t,y)$ & 43\% & 100\% \\
Wide-Tree-PPO & 64 $\times$ (1 + 3) & none; same trees and critic as policy channel only & 43\% & 100\% \\
Fixed WM & 64 $\times$ (1 + 3) & as policy channel only; world model never updated & 43\% & 100\% \\
\midrule
PaW & 160 & auxiliary next-observation loss on the policy & 100\% & 100\% \\
Majority-vote labels & 160 & plurality label among 4 samples per step & 100\% & 100\% \\
OPD (32B) & 160 & teacher's token log-probabilities in the student's context & 100\% & 100\% \\
OCSD (self, 32B) & 160 & advantage scaled by the teacher's response to the realized next observations & 100\% & 100\% \\
\bottomrule
\end{tabularx}
\end{table}

We test whether learning to value predicted consequences raises policy-only success for a comparable amount of real interaction. All methods share a 12-iteration PPO warm-up checkpoint and task order, and are evaluated as policies alone at fixed iterations. \wmppo{} and PPO-160 use approximately the same number of real steps. \wmppo{}, its policy-channel-only variant, PPO-160, PPO-256, \qppo{} and Wide-Tree-PPO have three training runs each, as does self-teacher OCSD on ALFWorld. \wmppo{}'s hyperparameters are set once on ALFWorld; baselines are re-tuned by rules fixed in advance when their settings fail (Section~\ref{sec:setup}).

\subsection{Experimental Setup}
\label{sec:setup}

\paragraph{Environments.} \emph{ALFWorld}~\citep{alfworld} is the simpler environment. Its six household task types mostly follow the same routine: find an object, heat, cool or clean it when required, and place it at the requested location. The environment lists admissible actions at every step, leaving search for hidden objects as the main difficulty. We train on 3,553 games and pool evaluation over 140 seen-room and 134 unseen-room games. \emph{ScienceWorld}~\citep{scienceworld} poses harder elementary science tasks. The 18 task types, fixed before training, all have oracle solutions within 40 steps. Eleven are procedural: the agent must first create or find an intermediate state, such as a mixture, a heated substance, a connected circuit or an organism in another room. The other seven require choosing the right object among candidates. The agent must identify the relevant conditions in long room descriptions and express its actions as free text using 25 templates. Unparseable commands consume a step, and focusing on the wrong object ends the task in failure. These dependencies make the value of a state transition less apparent from the next observation alone: the agent must judge whether a change establishes a condition needed for later success. Appendix~\ref{app:impl} describes ScienceWorld's fixed training and evaluation sets. In both environments, the reward is 1 for completion within 40 actions and 0 otherwise, with no intermediate rewards. ScienceWorld's reported progress is part of the observation and is also predicted by the world model.

\paragraph{Model and common start.} All methods train LoRA adapters (rank 64) on a frozen Qwen3-4B-Instruct-2507~\citep{qwen3} in non-thinking mode, with responses of at most 256 tokens. The shared PPO warm-up reaches 27.0\% success on ALFWorld and 20.8\% on ScienceWorld. Its real transitions also train the initial world model. Training then runs for 24 iterations on ALFWorld (13--36) and 48 on ScienceWorld (13--60).

\paragraph{Comparisons.} Table~\ref{tab:protocol} lists the comparisons in three groups. The first group contains established ways of training the same policy: PPO at the two budgets, GRPO and GiGPO~\citep{grpo,gigpo}, and three refinements of PPO for multi-turn agents, CAPO~\citep{capo}, \sopo{}~\citep{stppo} and \steppo{}, a turn-level form of GSPO~\citep{gspo}.

\begin{table}[!t]
\centering\small
\caption{\textbf{Policy-only success (\%) of the main comparisons.} Bold marks the highest mean in each column. Results are means over training runs; $\pm$ gives the standard deviation for methods with at least two runs. Runs: ALFWorld / ScienceWorld. Real steps from iteration 13 (ALFWorld / ScienceWorld): \wmppo{} 102k / 128k, \wmppo{} (policy channel only) 97k / 130k, PPO-160 105k / 113k, PPO-256 150k / 156k; Figure~\ref{fig:frontier} and Appendix~\ref{app:full} give all methods. $^\dagger$: \wmppo{} is significantly higher ($p<0.05$, per-game paired test over the evaluated checkpoints, uncorrected). $^{a}$ScienceWorld: the best of three settings, clipping ranges for \sopo{} and loss coefficients for PaW (Appendix~\ref{app:baselines}). $^{b}$ScienceWorld: all three clipping settings of CAPO diverged; the entry uses clip 0.1 at iteration 18, its highest evaluation before divergence, without a test.}
\label{tab:main}
\setlength{\tabcolsep}{4pt}
\begin{tabular}{@{}l c c c c c@{}}
\toprule
 & & \multicolumn{2}{c}{ALFWorld, iteration 36} & \multicolumn{2}{c}{ScienceWorld, iteration 60} \\
\cmidrule(lr){3-4}\cmidrule(lr){5-6}
Method & Runs & Greedy & pass@1 & Greedy & pass@1 \\
\midrule
\textbf{\wmppo{} (ours)} & 3 / 3 & 87.5\,{\scriptsize$\pm$0.9} & 87.8\,{\scriptsize$\pm$1.7} & \textbf{52.9\,{\scriptsize$\pm$2.3}} & \textbf{51.6\,{\scriptsize$\pm$4.8}} \\
\wmppo{} (policy channel only) & 3 / 3 & \textbf{88.4\,{\scriptsize$\pm$2.3}} & \textbf{89.4\,{\scriptsize$\pm$1.6}} & 47.2\sigup{}\,{\scriptsize$\pm$2.8} & 47.2\sigup{}\,{\scriptsize$\pm$1.7} \\
\midrule
\multicolumn{6}{l}{\emph{Standard training and refinements}} \\
PPO-160 & 3 / 3 & 63.5\sigup{}\,{\scriptsize$\pm$9.8} & 71.0\sigup{}\,{\scriptsize$\pm$6.0} & 39.3\sigup{}\,{\scriptsize$\pm$1.5} & 42.6\sigup{}\,{\scriptsize$\pm$2.5} \\
PPO-256 & 3 / 3 & 83.1\sigup{}\,{\scriptsize$\pm$3.1} & 84.8\sigup{}\,{\scriptsize$\pm$1.2} & 43.8\sigup{}\,{\scriptsize$\pm$8.7} & 43.5\sigup{}\,{\scriptsize$\pm$6.6} \\
GRPO & 1 / 1 & 59.5\rlap{\sigup{}} & 67.8\rlap{\sigup{}} & 35.2\rlap{\sigup{}} & 35.7\rlap{\sigup{}} \\
GiGPO & 1 / 1 & 78.1\rlap{\sigup{}} & 81.0\rlap{\sigup{}} & 38.4\rlap{\sigup{}} & 36.3\rlap{\sigup{}} \\
CAPO & 1 / 1 & 87.2 & 86.7 & 21.1\rlap{$^{b}$} & 20.0\rlap{$^{b}$} \\
\sopo{}$^{a}$ & 1 / 1 & 86.5 & 83.1\rlap{\sigup{}} & 46.5\rlap{\sigup{}} & 49.4 \\
\steppo{} & 1 / 1 & 85.8 & 85.0 & 25.0\rlap{\sigup{}} & 26.5\rlap{\sigup{}} \\
\midrule
\multicolumn{6}{l}{\emph{Structural controls (Section~\ref{sec:understanding})}} \\
\qppo{} (learned $Q$ values) & 3 / 3 & 84.4\sigup{}\,{\scriptsize$\pm$9.0} & 88.6\,{\scriptsize$\pm$3.3} & 46.1\sigup{}\,{\scriptsize$\pm$5.2} & 46.2\sigup{}\,{\scriptsize$\pm$2.5} \\
Wide-Tree-PPO (no aux.\ update) & 3 / 3 & 78.5\sigup{}\,{\scriptsize$\pm$9.4} & 82.3\sigup{}\,{\scriptsize$\pm$5.2} & 40.8\sigup{}\,{\scriptsize$\pm$3.9} & 40.0\sigup{}\,{\scriptsize$\pm$3.3} \\
Fixed WM & 1 / 1 & 76.3\rlap{\sigup{}} & 79.4\rlap{\sigup{}} & 38.7\rlap{\sigup{}} & 39.0\rlap{\sigup{}} \\
\midrule
\multicolumn{6}{l}{\emph{Auxiliary feedback added to PPO-160}} \\
PaW$^{a}$ & 1 / 1 & 85.4 & 87.3 & 35.6\rlap{\sigup{}} & 38.8\rlap{\sigup{}} \\
Majority-vote labels & 1 / 1 & 17.2\rlap{\sigup{}} & 17.6\rlap{\sigup{}} & 25.0\rlap{\sigup{}} & 25.0\rlap{\sigup{}} \\
OCSD (self teacher) & 3 / 1 & 83.0\sigup{}\,{\scriptsize$\pm$1.8} & 82.7\sigup{}\,{\scriptsize$\pm$3.1} & 34.2\rlap{\sigup{}} & 43.7\rlap{\sigup{}} \\
OCSD (32B teacher) & 1 / 1 & 83.2 & 70.1\rlap{\sigup{}} & 43.0\rlap{\sigup{}} & 41.5\rlap{\sigup{}} \\
OPD (32B teacher) & 1 / 1 & 41.2\rlap{\sigup{}} & 51.0\rlap{\sigup{}} & 36.3\rlap{\sigup{}} & 33.9\rlap{\sigup{}} \\
\bottomrule
\end{tabular}

\end{table}

The second group contains structural controls. \wmppo{} (policy channel only) removes the critic target of Section~\ref{sec:method-value}; each other control changes one element of that variant. Wide-Tree-PPO keeps the trees and critic training but removes the auxiliary update. \qppo{} values candidates directly with $Q_\omega(s_t,y)$, learned from real returns. Fixed WM keeps the initial world model throughout training (Appendix~\ref{app:baselines}).

The third group adds other auxiliary supervision to PPO-160. PaW~\citep{paw} uses the published auxiliary next-observation loss to train the policy on dynamics, without evaluating predicted consequences for task success. Majority-vote labels use ungated self-training on the most common of four sampled actions~\citep{ttrl,srt}. Standard on-policy distillation (OPD)~\citep{gkd} uses a frozen Qwen3-32B teacher (Appendix~\ref{app:teacher}). OCSD~\citep{ocsd} uses the policy itself or the 32B model as a teacher with access to the observations that followed the action.

\paragraph{Hyperparameters across environments.} \wmppo{} uses its ALFWorld hyperparameters unchanged on ScienceWorld, adapting the prompts and the world model's context and output lengths (Appendix~\ref{app:impl}). Baselines also start from their ALFWorld settings and follow pre-set re-tuning rules if those settings fail. CAPO, \sopo{} and PaW required re-tuning (Appendix~\ref{app:baselines}).

\paragraph{Interaction budgets.} The main comparison approximately matches the number of real steps. A real step is an environment step whose action the policy generated and executed, in chains and branches alike, counted from iteration 13; the shared warm-up and the prefix replays are not counted. Because branch points are drawn uniformly along the chain, the 64 chains and 192 branches of a \wmppo{} iteration generate about as many real steps as $64\times(1+3\times\tfrac12)=160$ episodes, the budget of PPO-160 (Table~\ref{tab:protocol}). In practice, \wmppo{} executed 3\% fewer real steps than PPO-160 on ALFWorld and 13\% more on ScienceWorld, where its 128k steps lie between those of PPO-160 (113k) and PPO-256 (156k). We report gains relative to PPO-160. Restoring a branch point by replaying the chain's prefix brings the total environment calls to about those of the $64\times4=256$ episodes of PPO-256, which we include as a supplementary reference. Within this budget, \wmppo{} covers 43\% of the ALFWorld training tasks, against 100\% for PPO.

\paragraph{Evaluation protocol.} Checkpoints are evaluated with the policy alone. Greedy success is the success rate of one greedy episode per evaluation game; pass@1 is the success probability of a single temperature-1 episode, estimated from three episodes per game (pass@3 in Appendix~\ref{app:full}). Unless stated otherwise, success refers to greedy success. The main comparison reports iteration 36 on ALFWorld and iteration 60 on ScienceWorld, with the checkpoints and hyperparameters of \wmppo{} and PPO fixed before evaluation. On ScienceWorld, \sopo{} and PaW each use the best of three settings in Table~\ref{tab:main}. All three CAPO settings diverged, so its entry uses the checkpoint with the highest greedy success during training, without a significance test (Appendix~\ref{app:baselines}). Each seed is represented by its first training run. Differences are tested per evaluation game with two-sided $z$-tests on the mean paired difference, uncorrected for multiple comparisons; Appendix~\ref{app:seeds} reports each run.

\subsection{Policy-only Task Performance}
\label{sec:main-results}

\paragraph{Higher success at comparable real interaction.} On ALFWorld at iteration 36, \wmppo{} succeeds on 87.5\% of the evaluation games under greedy decoding and on 87.8\% under pass@1 evaluation (Table~\ref{tab:main}); \wmppo{} (policy channel only) reaches 88.4\% and 89.4\%. PPO-160, whose budget of real steps is matched by design, reaches 63.5\% and 71.0\%, and every run of either variant is above every PPO-160 run. The supplementary PPO-256, with about 50\% more real steps, reaches 83.1\% and 84.8\%, also significantly lower in the per-game comparison.

On ScienceWorld at iteration 60, using the hyperparameters from ALFWorld, \wmppo{} reaches 52.9\% greedy and 51.6\% pass@1 success, against 39.3\% and 42.6\% for PPO-160 ($p<0.001$). Every \wmppo{} run (50.4--54.9\% greedy) is above every PPO-160 run (37.7--40.5\%). PPO-256, with 22\% more real steps, is 9.2 points lower under greedy decoding and 8.0 points lower for pass@1 ($p<0.001$). GRPO and GiGPO remain well below \wmppo{} in both environments. They stay below when drawing 40 tasks of 4 episodes each, which covers 91\% of the ScienceWorld variations (GiGPO on ScienceWorld only; Appendix~\ref{app:full}).

\paragraph{ScienceWorld separates the strongest methods.} On the simpler ALFWorld, CAPO, \sopo{}, \steppo{}, PaW and the two \wmppo{} variants cluster at 85.4--88.4\% greedy success, approaching the environment's ceiling. This leaves limited room for success rates to distinguish the methods. The two \wmppo{} variants have the highest greedy success, and adding the value channel gives no further gain at this level (87.5\% against 88.4\%, not significant). Under pass@1, \wmppo{} (policy channel only) is ahead of CAPO by 2.7 points, \sopo{} by 6.3 ($p<0.001$) and \steppo{} by 4.4 ($p=0.002$), and the full method is 4.7 points ahead of \sopo{} ($p=0.01$).

ScienceWorld, with its greater demands on intermediate conditions, gives a clearer separation. \wmppo{} reaches 52.9\% greedy success against 46.5\% for the best of \sopo{}'s three settings, the strongest baseline: a gain of 6.5 points ($p=0.004$). The lead under pass@1 is 2.2 points. Here, the value channel adds 5.8 greedy and 4.4 pass@1 points to the policy channel alone ($p<0.001$; Section~\ref{sec:learning}). The PPO refinements change how executed actions are credited; \wmppo{} adds comparisons with unexecuted actions and a critic target built from them.

\begin{figure}[t]
\centering
\includegraphics[width=\linewidth]{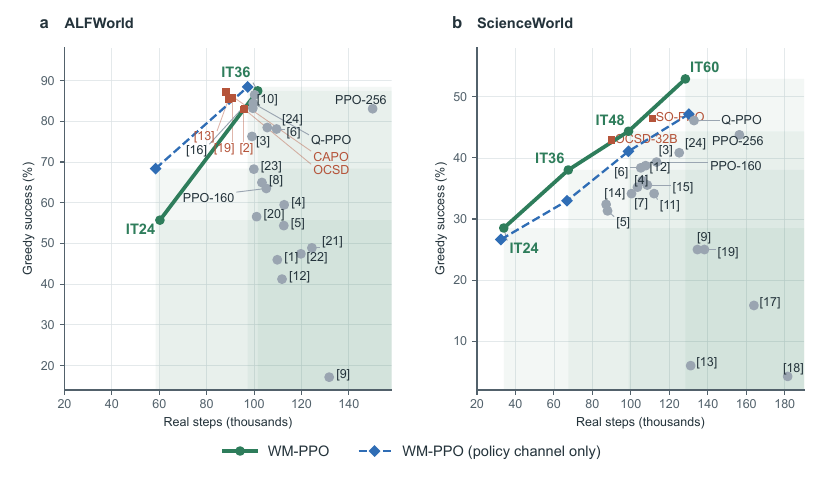}%
\caption{\textbf{Success against real interaction, for every method.} Each gray or red point shows one method at the end of the common protocol (ALFWorld iteration 36, ScienceWorld iteration 60): mean greedy success in the official evaluation against mean real steps from iteration 13, averaged over training runs. Green: the evaluated checkpoints of \wmppo{} (IT: iteration); dashed: \wmppo{} (policy channel only). The shaded region lies below and to the right of at least one evaluated checkpoint (of either variant on ALFWorld, of \wmppo{} on ScienceWorld). A method inside it used more real steps and achieved lower success; darker shading marks regions dominated by more checkpoints. Red squares: methods outside the region. Bracketed numbers identify the other methods, named in Figure~\ref{fig:frontier-full} in Appendix~\ref{app:full}. Same-seed reruns and the diverged ScienceWorld CAPO runs are omitted.}
\label{fig:frontier}
\end{figure}

PaW reaches 85.4\% success on ALFWorld, 2.1 points below \wmppo{}. On ScienceWorld, the single run with its best of three loss weights reaches 35.6\%, broadly at PPO-160's level and slightly below its 39.3\%. \wmppo{} is 17.4 points higher ($p<0.001$). \wmppo{} uses the predicted observations to evaluate which consequences matter for task success.

\paragraph{Higher success than the tested teacher-based methods.} OPD and OCSD use teachers to supply feedback finer than a final 0/1 reward. Self-teacher OCSD improves PPO-160 to 83.0\% greedy and 82.7\% pass@1 success on ALFWorld over three runs; \wmppo{} remains 4.5 and 5.1 points higher ($p=0.003$ and $p<0.001$). On ScienceWorld, self-teacher OCSD is 5.2 points below PPO-160 under greedy decoding ($p=0.02$), level with it for pass@1, and 18.8 and 7.9 points below \wmppo{}. The teacher's judgment separates valid from invalid actions on ALFWorld's short observations (AUC 0.80), but is inverted on ScienceWorld's long descriptions (AUC 0.25 and 0.21; Appendix~\ref{app:teacher}). With the 32B teacher, OCSD reaches 70.1\% pass@1 on ALFWorld and is level with PPO-160 on ScienceWorld. Standard OPD ends below PPO-160 in both environments (Appendix~\ref{app:teacher}). Ungated majority-vote labels reinforce the policy's repeated choices without information about consequences; success falls to 17.2\% on ALFWorld and 25.0\% on ScienceWorld. \wmppo{} outperforms these methods using a world model and critic trained on the agent's own interaction.

\paragraph{Performance across runs.} The runs of \wmppo{} vary little (86.5--88.3\% success on ALFWorld, 50.4--54.9\% on ScienceWorld), whereas several baselines vary much more, for example PPO-160 over 52.2--70.1\% on ALFWorld and PPO-256 over 33.8--49.3\% on ScienceWorld (Appendix~\ref{app:seeds}).

\subsection{Interaction Efficiency}
\label{sec:efficiency}

\paragraph{No other method achieves higher success with fewer real steps.} Figure~\ref{fig:frontier} compares each method's success and real interaction. On ALFWorld, 18 of the 23 other methods lie in the region below and to the right of the evaluated checkpoints of the two \wmppo{} variants; on ScienceWorld, 18 of 20 lie below and to the right of \wmppo{}'s checkpoints. The methods outside the region used fewer real steps but reached lower success: on ALFWorld, CAPO (with two clipping ranges), \steppo{}, OCSD and PaW; on ScienceWorld, the tuned \sopo{} and OCSD with the 32B teacher. PPO-256 uses 150k real steps against \wmppo{}'s 102k on ALFWorld and 156k against 128k on ScienceWorld, and it ends below \wmppo{} in both environments.

\paragraph{The same success takes fewer real steps than with PPO.} Figure~\ref{fig:curves} in Appendix~\ref{app:full} plots success against real interaction; Table~\ref{tab:reach} reports each run's first evaluated checkpoint above a reference success. By iteration 36 on ALFWorld, the three \wmppo{} runs have passed 80\% success after 69k--97k real steps and the three runs with the policy channel only after 67k--84k, against 129k--141k for PPO-256. No PPO-160 run has reached 80\% by this iteration. The PPO refinements rise fastest early: CAPO and \steppo{} pass 80\% after 63k and 64k real steps. \wmppo{} reaches their level by iterations 30--36, and the two variants end with the highest success at iteration 36. On ScienceWorld, \wmppo{} separates from PPO-160 after about 30k real steps, and the gap widens to about 6 points by 100k. Its runs pass 45\% after 94k--120k real steps and the tuned \sopo{} after 97k, whereas no PPO-160 run passes 45\% within its 108k--118k steps. The gap grows over training, in step with the reliability of the comparisons (Section~\ref{sec:valuing}).

\paragraph{Computation.} Serving and training the world model takes additional GPUs during training. At comparable real steps, an iteration therefore takes 1.3--1.6 times as long as for PPO-160 in wall-clock time and uses 2.7--3.8 times as many GPU-hours (Appendix~\ref{app:cost}). This cost stays in training, where the world model and the critic extract the information implicit in environment dynamics; at inference the policy acts alone, without the extra models and GPUs of agents that keep a world model~\citep{wma,itp,comap}. The additional computation lets \wmppo{} learn more from each real state transition, a tradeoff that is useful when interaction is slow or costly.

\section{Understanding the Gains}
\label{sec:understanding}

A transition value estimates the current policy's chance of success after a candidate action (Equation~\eqref{eq:score}). This section examines whether these values identify choices that determine task success, how the policy and critic learn from them, and how the trained policy behaves as a result.

\begin{figure}[t]
\centering
\includegraphics[width=\linewidth]{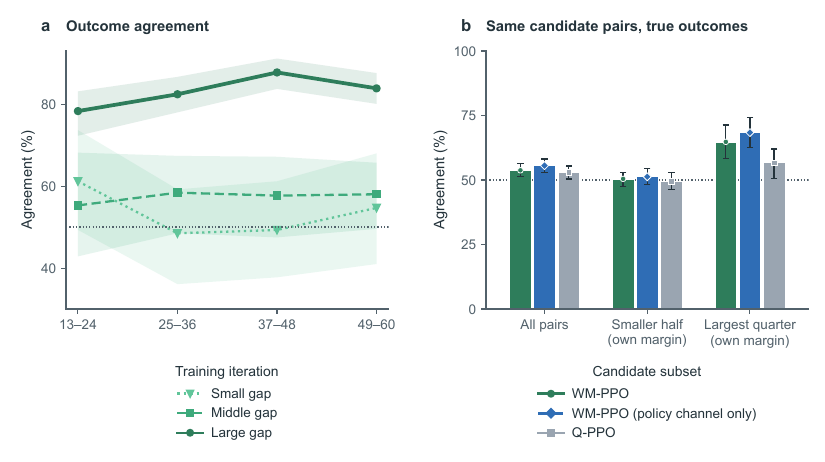}
\caption{\textbf{Agreement of preferences with real outcomes} (ScienceWorld). \textbf{(a)}~Training branch pairs from three \wmppo{} runs with different real outcomes and untied transition values, evaluated before each update, grouped by phase and margin bin (0.5 and 0.8 quantiles of nonzero margins). \textbf{(b)}~Final evaluators score a shared pool of 775 candidate pairs from evaluation states whose outcome values, estimated from five real continuations per candidate, differ. The smaller half and largest quarter are selected by each evaluator's own margins. Dotted line: 50\%; 95\% intervals: bands in (a), error bars in (b), from resampling (run, iteration) clusters and decision points, respectively. Training protocol and full results: Appendix~\ref{app:branch}.}
\label{fig:branch}
\end{figure}

\subsection{Valuing Predicted State Transitions}
\label{sec:valuing}

Section~\ref{sec:diag-real} showed that large value gaps identify comparisons the critic orders more reliably. This pattern persists during online training with predicted successors. Late in ScienceWorld training, the preferred action has the better outcome in 84\% of branch pairs with different real outcomes in the largest margin bin, while the smaller bins are close to chance (Figure~\ref{fig:branch}a; Table~\ref{tab:branch-split}).

\paragraph{What a large margin means.} A large margin indicates a large estimated difference in the current policy's chances of success after the candidate actions. Such a difference can arise when an action commits the agent to a choice that prevents later completion. For example, a ScienceWorld task requires focusing on a non-living thing and moving it into a box. Focusing on air leaves the episode running, but air cannot be moved into the box. Across the three runs, the critics value the predicted successor of focusing on air at 0.03--0.34, compared with 0.54--0.80 for going to the hallway. None of the 24 real continuations after focusing on air succeeds, whereas 19 of 24 succeed after going to the hallway.

\begin{samepage}
Such decisive choices make up a growing part of the largest bin. Its share of pairs with different real outcomes grows from 14\% early in training to 33\% later, and its realized gain per pair from +0.08 to +0.22 (Table~\ref{tab:branch}), which keeps its gate weight high (Section~\ref{sec:learning}).

\end{samepage}

\paragraph{Predicting successors improves candidate comparisons.} Since a transition value approximates an action value, directly learning a Q network is a natural alternative. The relation from Equation~\eqref{eq:bellman},
\[
\qhat_t(a)\approx Q^\pi(s_t,a)=\mathbb E\bigl[V^\pi(s_{t+1})\bigm| s_t,a_t=a\bigr],
\]
allows either a direct estimate of $Q^\pi$ or an estimate through the action's successor. \qppo{} tests the first route by adding a network $Q_\omega(s_t,y)$ trained on executed actions' returns. It replaces the world model and critic's candidate evaluation while retaining the candidates, margins, gate and auxiliary update of \wmppo{} (policy channel only). The world model can also learn from the next observation at every executed step, including those in failed episodes. Predicting successors therefore lets the evaluation use supervision about what actions change as well as the returns used to learn value. In ScienceWorld's largest margin bin, \wmppo{} (policy channel only), which differs from \qppo{} only in how candidates are valued, favors the successful action in 85\% of training branch pairs with different real outcomes, against 76\% for \qppo{} (Table~\ref{tab:branch-split}). On a shared pool of candidates, the comparisons using the world model are also more accurate at large margins (Figure~\ref{fig:branch}b).

\begin{figure}[t]
\centering
\includegraphics[width=\linewidth]{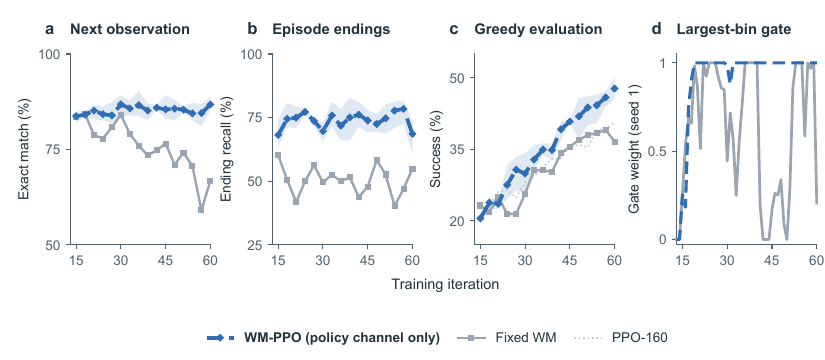}%
\caption{\textbf{Online updates maintain prediction quality as the policy learns} (ScienceWorld). \wmppo{} (policy channel only) and Fixed WM start from the same world model. \textbf{(a)}~Exact next-observation prediction on executed steps. \textbf{(b)}~Recall of episode endings. \textbf{(c)}~Greedy success during training on dev and test games. \textbf{(d)}~Gate weight of the largest margin bin. Each model is evaluated on its policy's visited states, in windows of three iterations plotted at the last iteration. Band: range of three online runs; Fixed WM: one run; dotted gray line: mean of three PPO-160 runs.}
\label{fig:keepup}
\end{figure}

\paragraph{Online updates preserve prediction quality on ScienceWorld.} As the policy learns, it visits new states and proposes different actions, so the world model must keep learning from the resulting transitions. With a fixed world model, prediction accuracy declines as the policy changes (Figure~\ref{fig:keepup}a), and so does candidate evaluation: late in training, its largest-margin agreement is 53--64\% against 88--93\% online (same seed; Appendix~\ref{app:branch}), and the gate keeps closing that bin (Figure~\ref{fig:keepup}d).

\FloatBarrier
\subsection{Learning from Transition Values}
\label{sec:learning}
\label{sec:value-channel}

The comparisons supply information that additional branches alone do not. Wide-Tree-PPO keeps the trees and critic of \wmppo{} (policy channel only) but drops the auxiliary update; across three runs per method, the policy-channel-only variant exceeds it in success by 10.0 percentage points on ALFWorld and 6.3 on ScienceWorld (Table~\ref{tab:main}). A single offline update on these comparisons alone also improves the policy (Section~\ref{sec:diag-learn}).

The value channel adds feedback for the critic. Training the critic toward the mean transition value $\bar q_t$ changes how it values the next round's predicted successors, and hence the preferences used in later policy updates. In ScienceWorld iterations 13--36, the largest-margin comparisons favor the successful action in 78--82\% of branch pairs with different outcomes, against 71--74\% with the policy channel only (Figure~\ref{fig:value-channel}b).

\begingroup
\setlength{\intextsep}{4pt}\setlength{\abovecaptionskip}{3pt}
\begin{figure}[H]
\centering
\includegraphics[width=\linewidth]{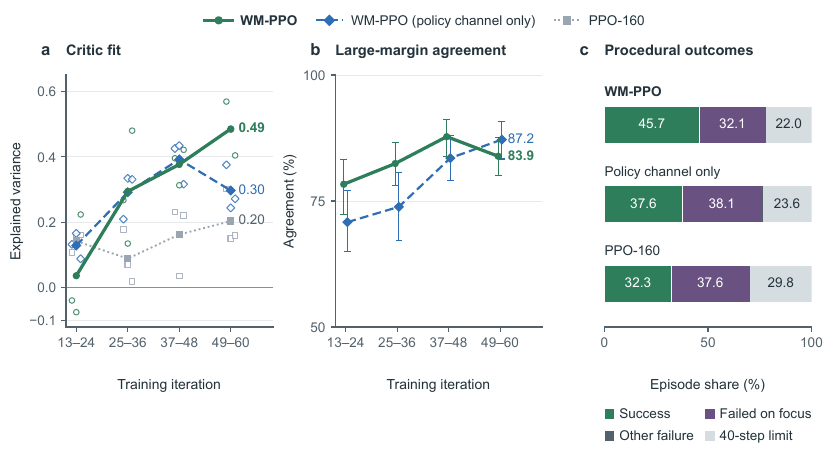}
\caption{\textbf{What the value channel adds} (ScienceWorld, three runs per method). \textbf{(a)}~Explained variance of real returns at tree nodes, using the critic for that iteration's advantages; lines: means over runs, markers: individual runs. \textbf{(b)}~Agreement in the largest margin bin among branch pairs whose outcomes and transition values differ; bars: 95\% intervals from resampling (run, iteration) clusters. \textbf{(c)}~Episode endings in greedy re-runs of the final checkpoints on 188 procedural games per run. Official procedural success is reported in Section~\ref{sec:behavior}.}
\label{fig:value-channel}
\end{figure}
\endgroup

\begingroup
\setlength{\intextsep}{4pt}\setlength{\abovecaptionskip}{3pt}
\begin{figure}[H]
\centering
\includegraphics[width=\linewidth]{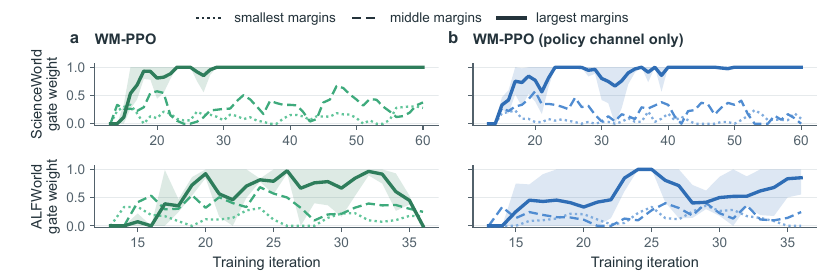}
\caption{\textbf{Gate weights during training}: \wmppo{} (a, green) and the policy-channel-only variant (b, blue). Top: ScienceWorld, iterations 13--60; bottom: ALFWorld, iterations 13--36. Dotted, dashed and solid lines: means over three runs for the smallest, middle and largest margin bins; band: range over runs for the largest bin. Weights use branch pairs from the preceding four iterations (Section~\ref{sec:method-gate}).}
\label{fig:gate-weights}
\end{figure}
\endgroup

A single gate decides when, and how much, these comparisons shape both networks. The same weight $w_{\ell(m_t)}$ scales the policy's auxiliary advantages and the critic's auxiliary loss (Equations~\eqref{eq:aux} and~\eqref{eq:cons}); while the gate is closed, the update is ordinary PPO on the trees. Each margin bin earns its weight from the realized gain of its earlier comparisons (Section~\ref{sec:method-gate}).

On ScienceWorld, the gate starts closed and opens as the critic and world model become reliable; with a fixed world model, its largest bin closes repeatedly (Figure~\ref{fig:keepup}d). The largest bin reaches full weight early in training and keeps it as its comparisons agree more often with real outcomes (Figure~\ref{fig:gate-weights}a; Table~\ref{tab:branch}); its weight still dips when its recent comparisons are less reliable, as around iteration 27. The smaller bins, whose comparisons are close to chance but carry a slightly positive gain, stay at low and fluctuating weights, so their weak signal enters both channels without dominating them.

The value channel feeds this calibration. With more reliable comparisons, the largest bin of \wmppo{} opens earlier and stays open; with the policy channel only, the weights are less stable and the largest bin dips in mid-training, so less feedback reaches the two networks (Figure~\ref{fig:gate-weights}a,b). With the fuller feedback, \wmppo{} pulls ahead of the policy-channel-only variant from iteration 36 onward (Tables~\ref{tab:full-sw} and~\ref{tab:full-sw-b}), its critic fits real returns better late in training (explained variance 0.49 against 0.30; Figure~\ref{fig:value-channel}a), and fewer procedural episodes end in failure on a wrong focus (Figure~\ref{fig:value-channel}c).

On ALFWorld, the same calibration withdraws feedback as success approaches the ceiling. Few branch pairs then have different outcomes, and equal outcomes contribute no realized gain. The weights fluctuate, and \wmppo{}'s largest-margin bin closes at iteration 36 (Figure~\ref{fig:gate-weights}a, lower). With less branch evidence and little room to raise success, the value channel gives no further gain on ALFWorld (Section~\ref{sec:main-results}).

\begin{figure}[H]
\centering
\includegraphics[width=4.8in]{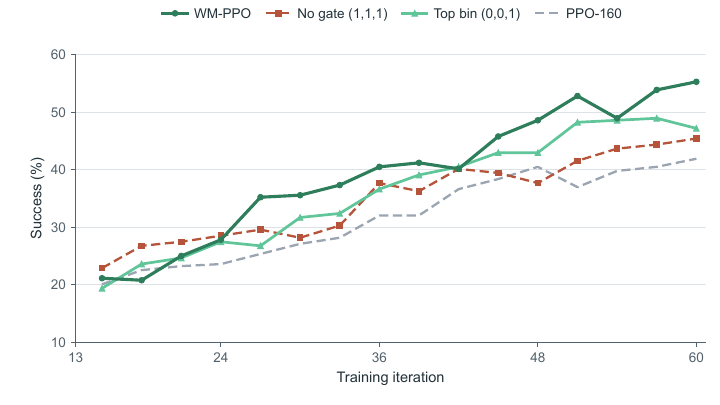}
\caption{\textbf{Gate ablation on ScienceWorld} (one seed: the first run of each method). Success during training on the dev and test games. The two copies of \wmppo{} fix the bin weights from iteration 13 at $(1,1,1)$ (no gate) or $(0,0,1)$ (top bin, the largest-margin bin only) in both channels; all other settings are unchanged.}
\label{fig:gate-ablation}
\end{figure}

Fixing the weights shows what this calibration contributes (Figure~\ref{fig:gate-ablation}). Full weight rises fastest at first, because it admits the comparisons that the gate still holds back. Later, the near-chance preferences of the smaller bins keep entering at the same weight as the reliable ones, and this copy's success grows no faster than PPO-160's; on its own branch records, the gate would have admitted the small bin in only about a quarter of the iterations. Keeping only the largest bin is stable but stays below \wmppo{}: it never receives the weak signal that the gate would have drawn from the smaller bins in more than half of its iterations, and it trusts the largest bin from the first iteration, before its comparisons have proven reliable (Appendix~\ref{app:gate-ablation}). Past the early phase, \wmppo{} pulls clearly ahead, and both copies end significantly below it, ordered by the reliability of the margin bins (Section~\ref{sec:valuing}).

\begin{figure}[t]
\centering
\includegraphics[width=\linewidth]{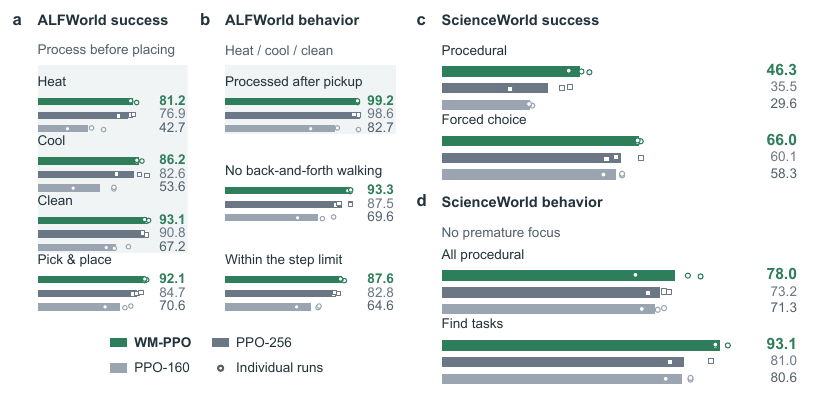}
\caption{\textbf{How the trained policies advance the task}: \wmppo{}, PPO-256 and PPO-160, three runs each. Bars: pooled percentages; markers: individual runs. \textbf{(a, c)}~Official greedy success by task type; shaded ALFWorld tasks require processing before placement. \textbf{(b)}~ALFWorld behavior in greedy re-runs. Processing is measured among episodes in which the object was picked up; other rates cover all episodes. Back-and-forth walking: at least two A--B--A--B patterns, possibly overlapping, in the \emph{go to} destinations; within the step limit: ending before the 40th action. \textbf{(d)}~Procedural ScienceWorld episodes without premature focus in greedy re-runs: all procedural games and the find subset. A premature focus commits to an answer before its prerequisite state exists. Further behavior analyses: Appendix~\ref{app:behavior}.}
\label{fig:endings}
\end{figure}

\FloatBarrier

\subsection{How the Policy Advances the Task}
\label{sec:behavior}

The trained \wmppo{} policy more often completes prerequisites before taking the action that depends on them.

\begingroup
\setlength{\intextsep}{4pt}
\begin{figure}[t]
\centering
\includegraphics[width=\linewidth]{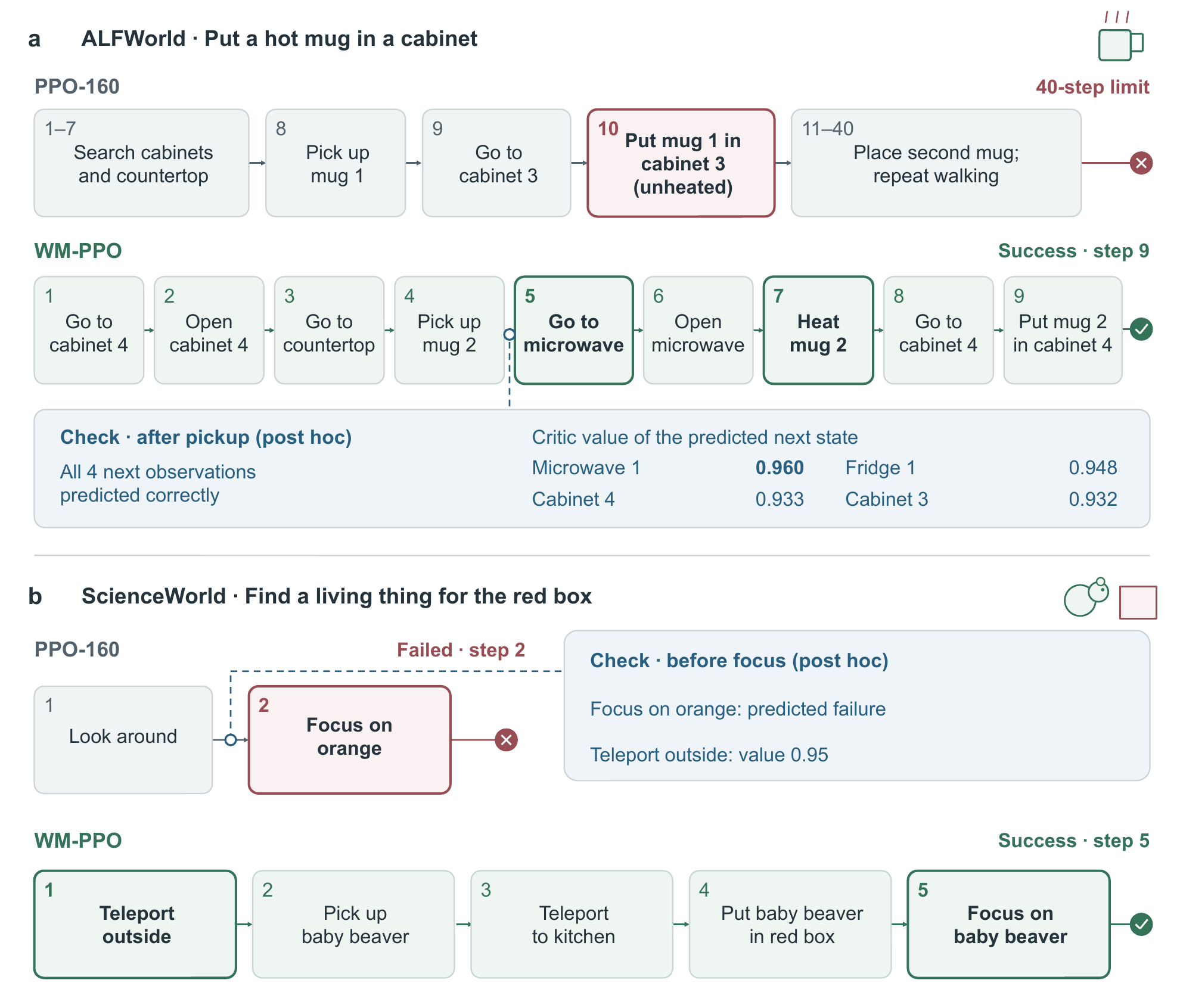}
\caption{\textbf{Two evaluation games} played greedily by the final policies; highlighted: decisive steps. \emph{Check}: post hoc evaluation by the world model and critic of the same \wmppo{} run; displayed critic values are raw outputs. \textbf{(a)}~ALFWorld, the second PPO-160 run and first \wmppo{} run. \textbf{(b)}~ScienceWorld, the first run of each method.}
\label{fig:case}
\end{figure}
\endgroup

\paragraph{ALFWorld.} \wmppo{} more often processes an object before placing it. Its gain over PPO-160 is largest on tasks that require heating, cooling or cleaning before placement: 31.5 points, against 21.5 on plain placing (Figure~\ref{fig:endings}a). The action order helps explain this difference. On processing tasks, \wmppo{} places the object before processing it in 17\% of episodes in which the object was picked up, against 41\% for PPO-160 (Appendix~\ref{app:behavior}). In the mug-heating game, both policies find the mug; \wmppo{} takes it to the microwave, whereas PPO-160 puts it into the cabinet unheated and fails (Figure~\ref{fig:case}a). PPO-256 also learns the processing order with more interaction; \wmppo{} learns it within PPO-160's budget (Section~\ref{sec:efficiency}).

\paragraph{ScienceWorld.} \wmppo{} is more successful on tasks that require intermediate steps and less often commits to an answer before meeting its prerequisites. On the procedural games, it reaches 46.3\% success, against 29.6\% for PPO-160 (Figure~\ref{fig:endings}c). An accepted focus command commits the agent to an answer, so its timing reveals whether the necessary conditions have been met. On these games, premature focus occurs in 22.0\% of \wmppo{}'s episodes, against 28.7\% for PPO-160 (Figure~\ref{fig:endings}d). The living-thing task illustrates the difference: \wmppo{} goes outside and brings a baby beaver to the box, while PPO-160 focuses on a fruit in the starting room and fails (Figure~\ref{fig:case}b).

\section{Related Work}
\label{sec:related}

Language agents can draw on information beyond the terminal reward in several ways. Table~\ref{tab:related} compares the main approaches by what is predicted, how it is evaluated, where the resulting signal enters training, and what the deployed agent needs.

\begin{table}[!ht]%
\centering\footnotesize
\caption{\textbf{How information beyond the terminal reward enters agent training.} ``--'': not part of the method. References are given in the text.}
\label{tab:related}
\setlength{\tabcolsep}{3pt}
\begin{tabularx}{\linewidth}{@{}>{\raggedright\arraybackslash}p{0.17\linewidth} >{\raggedright\arraybackslash}X >{\raggedright\arraybackslash}X >{\raggedright\arraybackslash}X >{\raggedright\arraybackslash}p{0.15\linewidth}@{}}
\toprule
Method & What is predicted & How it is evaluated & Where the signal enters & Deployed agent \\
\midrule
ECHO, PaW, DGS-WM & next observation, by the policy & -- & auxiliary loss on the policy & policy \\\addlinespace[1pt]
Early Experience, SPA & future or true states & -- & supervised stage before RL & policy \\\addlinespace[1pt]
RWML & next state & similarity to the true state & RL stage before task RL & policy \\\addlinespace[1pt]
WMA, WebDreamer & change caused by each candidate & value model or prompted judge & action selection at test time & policy, model, evaluator \\\addlinespace[1pt]
ITP & multi-step imagined rollout & the policy's reflection (training-free variant) & context of the policy; SFT, then actor-critic RL & policy and model \\\addlinespace[1pt]
CoMAP & imagined next state & the policy's own reflection & supervised refinement & policy and model \\\addlinespace[1pt]
Internalizing the Future & imagined plan (prospective rollout) & verbalized confidence, calibrated against outcomes & auxiliary step-level advantage in GRPO & policy with explicit foresight \\\addlinespace[1pt]
ProAct (MC-GRPO / MC-PPO) & -- (simulator transitions) & immediate reward and random-policy Monte-Carlo state values & group advantages (MC-GRPO); TD(0) advantages (MC-PPO) & policy \\\addlinespace[1pt]
OPD, OCSD & -- & token likelihood under a teacher or a hindsight self-teacher & distillation loss or rescaled token advantage & policy \\\addlinespace[1pt]
Dreamer, MBPO, DreamGym & imagined rollouts & critic or returns on imagined data & training on imagined rollouts & policy \\
\midrule
\wmppo{} & next observation of each candidate, by a separate model & PPO's critic on the predicted successor, weighted by real branches & auxiliary advantage in PPO-clip; one-step target for the critic & policy \\
\bottomrule
\end{tabularx}
\end{table}

\paragraph{Observation prediction and dynamics learning.} ECHO, PaW and DGS-WM use real observations as auxiliary prediction targets during RL, so the policy's own parameters are also trained to predict what its actions produce~\citep{echo,paw,dgswm}. DGS-WM keeps the observation fragments that best support the policy's real next response, which favors decision-relevant content, but this selection score does not estimate eventual success. Other methods learn the dynamics before task-level policy training. Early Experience offers next-state prediction as a warm-up before imitation, and alternatively mixes self-reflection rationales into imitation~\citep{earlyexp}; SPA writes true states into the policy's reasoning before RL~\citep{spaiwm}; and RWML rewards predictions that are close to the true next state~\citep{rwml}. In all of these methods, a prediction is judged by how well it matches the real observation. \wmppo{} keeps the world model separate and values each predicted change by its effect on task success. The two uses of a world model could be combined.

\paragraph{Predicted outcomes at decision time.} WMA already pairs candidate actions with predicted consequences and a value judgment. With the policy frozen, a world model predicts the next observation of each candidate, a value model scores it, and the best candidate is executed at test time~\citep{wma}. The predictions guide which action is taken, while the policy itself is not updated. The value model is fine-tuned on progress labels $t/\mathrm{len}(\tau)$ from Mind2Web trajectories, or replaced by a prompted LLM, whereas the critic of \wmppo{} is trained on the returns of the current policy. WMA also reports that scoring the predicted next state works better than scoring the action alone; \qppo{} makes the corresponding comparison inside training (Section~\ref{sec:valuing}). WebDreamer plans in the same way with an LLM simulator and a prompted judge~\citep{webdreamer}, and QWM selects robot actions by world-model search with a learned Q-function~\citep{qlwm}. ITP lets the agent reflect on imagined multi-step rollouts before acting, either without training or after an expert warm-up and actor-critic RL that learns the imagination length together with the action; in both variants the world model stays in the deployed agent~\citep{itp}.

CoMAP, which is also evaluated on ALFWorld and ScienceWorld, shares several design choices with \wmppo{}~\citep{comap}. Its world model is a separately parameterized copy of the same backbone, retrained on the agent's own transitions; its labels come from real two-branch rollouts; and a gate decides when to refine. The policy drafts an action, reads the predicted next state and revises the draft after reflection. Training is supervised by expert or corrected actions, and the world model also learns from an EMA teacher that sees the real next state. The prediction thus enters the decision itself, and the world model remains part of the deployed agent. \wmppo{} instead turns predicted consequences into training signals through PPO's critic, and only the policy acts at deployment.

\paragraph{Learned foresight.} Internalizing the Future combines future simulation and success estimation within a single policy~\citep{itf}. Its mid-training and supervised fine-tuning use foresight traces that DeepSeek V3.1 writes from recorded trajectories, including their future continuations. During RL, the policy generates an imagined plan and a confidence score before reasoning about the next action. A calibration reward trains this score against the terminal outcome, and the scores of corresponding steps, normalized within the group, form an auxiliary step-level advantage in GRPO. The feedback therefore concerns the generated plan rather than a comparison of alternative actions from a shared state. \wmppo{} makes that comparison explicitly: the world model predicts the consequences of sampled actions, and PPO's critic evaluates the resulting states. Both learn from the agent's own interactions, and their estimates provide feedback for policy and critic updates. At inference, the policy acts without an explicit foresight stage.

\paragraph{Teacher and hindsight feedback.} Distillation supplies token-level feedback from a stronger or better-informed model. In standard on-policy distillation, the student generates its own sequences and a teacher supervises its token distribution on these prefixes~\citep{gkd}. OCSD, in its main configuration, uses the agent's own earlier policy instead: it scores already generated tokens in matched replay contexts with and without the real later observations, and the score difference rescales the token update while keeping the sign of the original advantage~\citep{ocsd}. Its hindsight also comes from the agent's own interaction; the difference is that OCSD re-scores executed responses, whereas \wmppo{} predicts the consequences of candidates that were not executed. VASPO gives a return-trained critic and a self-teacher access to reference solutions during training and weights tokens by their judgments, on mathematical and logical reasoning~\citep{vaspo}. Like \wmppo{}, it deploys the policy alone, but it predicts no state transitions. Section~\ref{sec:main-results} compares \wmppo{} with OPD and OCSD.

\paragraph{Model-based reinforcement learning.} \wmppo{} learns dynamics from the agent's own interactions and values predicted consequences to guide policy and critic learning. Dyna adds simulated transitions to learning~\citep{dyna}; World Models and Dreamer learn behavior inside a latent model~\citep{worldmodels,dreamer,dreamerv3}; MBPO branches short model rollouts from real states~\citep{mbpo}; MVE and STEVE build the critic's targets from model rollouts~\citep{mve,steve}; and DreamGym replaces the environment of text agents with an LLM simulator~\citep{dreamgym}. Value Prediction Networks and MuZero learn models through values~\citep{vpn,muzero}, and value equivalence judges a model by the values it induces~\citep{valueequiv}. In \wmppo{}, real rollouts provide every trajectory, the returns that are the critic's main supervision, and PPO's advantages. The world model predicts one step for the policy's candidates, and the critic's values of these predictions enter learning as candidate comparisons and an auxiliary critic target. This target, the mean transition value, is related to value expansion~\citep{mve,steve} and supplements regression on real returns. The gate checks the comparisons on real outcomes, in the spirit of value equivalence; the world model itself is trained by likelihood.

\paragraph{Policy improvement over sampled actions.} Valuing actions that were not executed, through a learned critic, is the idea of all-action policy gradients~\citep{mac,epg}. MPO reweights actions sampled from the policy by their estimated values, and AWR reweights the actions in its replay buffer by their estimated advantages~\citep{mpo,awr}; Muesli regularizes policy-gradient learning toward sampled actions valued with a learned model~\citep{muesli}; Sampled and Gumbel MuZero improve the policy over a sampled subset of actions with search~\citep{sampledmuzero,gumbelmuzero}; and the MC-GRPO variant of ProAct executes candidates sampled at one state in the simulator and values each by its immediate reward plus Monte-Carlo continuations from a random policy~\citep{proact}. The auxiliary term of \wmppo{} belongs to this family: it moves probability toward the best of the policy's own $K$ samples. Its choices concern how to value these samples and how much to trust their comparisons. \wmppo{} values candidates through their predicted successor states and PPO's critic, then uses outcomes from real branches to calibrate the reliability weights. Expert Iteration, AlphaZero, Dyna-Mind and the first stage of ProAct instead train the policy toward the output of a search~\citep{exit,alphazero,dynamind,proact}, whereas \wmppo{} supplies scalar advantages for the policy's own samples rather than a target distribution.

ProAct's MC-PPO variant estimates state values through random-policy rollouts in the simulator, with 1,000 rollouts per state in our Sokoban comparison~\citep{proact}. With comparable numbers of real environment steps, \wmppo{} reaches 87.0\% success versus 77.3\% for MC-PPO, a significant gain of 9.7 percentage points (Appendix~\ref{app:sokoban}).

\paragraph{Value and credit signals for language agents.} QLASS guides action selection with a Q-network trained on exploration trees~\citep{qlass}; SWEET-RL turns a privileged turn-level critic's ranking of sampled actions into preference pairs~\citep{sweetrl}; SPA-RL redistributes the final reward into per-step progress rewards~\citep{spa}; and process reward models score steps from step-level labels~\citep{prm800k}. Agent Q trains on sibling pairs whose value gap exceeds a fixed threshold~\citep{agentq}, whereas the gate of \wmppo{} measures which margins are reliable. VinePPO finds that PPO's value network ranks candidate next steps near chance and replaces it by Monte-Carlo estimates~\citep{vineppo}, consistent with the small-margin comparisons of Sections~\ref{sec:diag-real} and~\ref{sec:valuing}. Group-based methods compare episodes~\citep{grpo,treegrpo} or actions taken at the same state or history~\citep{gigpo,hgpo}; PPO refinements change how advantages and ratios apply within a turn~\citep{capo,stppo,vapo,dapo}; and majority-vote self-training rewards agreement with the policy's most frequent answer~\citep{ttrl,srt}. \wmppo{} complements these forms of credit assignment with learned comparisons of unexecuted alternatives.

\section{Discussion and Limitations}
\label{sec:discussion}

The value of a predicted consequence depends on what the current policy can still accomplish from the resulting state. As the policy improves, the same successor can acquire a different value. The world model learns which changes an action produces, and the critic relates those changes to task success using supervision from real returns. Evaluating alternatives in this way lets the policy learn preferences that a few realized outcomes leave unresolved, including between candidates whose observed continuations end with the same reward (Appendix~\ref{app:gae}).

Large margins increasingly identify choices that change what remains achievable (Section~\ref{sec:valuing}). A premature commitment can close off success, while a step that establishes a prerequisite can make it attainable. These are the decisions on which the comparisons are most reliable. The behavioral analyses show the corresponding change in the policy: it more often completes prerequisites before placing an object or committing to an answer. Gains are largest on procedural ScienceWorld tasks and ALFWorld tasks that require heating, cooling or cleaning before placement. They are smaller on forced-choice tasks, where correctness becomes apparent only at the end.

What the world model predicts determines which part of a comparison the critic must supply. Predicted endings supply outcomes directly, and comparisons between continuing successors rely on the critic. For the latter comparisons on ScienceWorld, the value channel raises agreement with real outcomes from 55\% to 62\% (Appendix~\ref{app:branch}). On ALFWorld, no predicted ending settles a comparison, so every comparison rests on the critic. The gate assigns only partial weights there (Figure~\ref{fig:gate-weights}). The value channel improves early candidate comparisons and the critic's fit to real returns later in training (Figure~\ref{fig:value-channel}; Section~\ref{sec:learning}).

\wmppo{} trades additional prediction and evaluation during training for more learning from each real interaction. The benefit is higher success at comparable interaction budgets and fewer steps to reach PPO's success levels. The extra computation is confined to training; the deployed policy acts alone.

\paragraph{Limitations.} Our main evaluation covers one 4B base model and two text environments, with episodes limited to 40 steps. Branching and gate calibration require a reproducible prefix, as in our deterministic environments, or a saved state. A branch's first action is sampled conditional on differing from the chain's action. The update does not correct for this conditioning.

A single greedy prediction of the next observation does not represent uncertainty over possible outcomes. Longer-range effects enter only through the critic. Prediction errors can bias the auxiliary policy advantages and critic targets. Comparisons cover only the policy's candidates, and many small-margin comparisons receive little or no weight. Extending the benefit to these decision points remains an open problem.

\section{Conclusion}
\label{sec:conclusion}

\wmppo{} uses learned dynamics to evaluate candidate actions through their predicted consequences. A world model trained on real transitions predicts each candidate's next observation, and PPO's critic, trained on real returns, values the predicted successor according to the current policy's chance of success. In the policy channel, comparisons between transition values become auxiliary advantages for the policy's own samples. In the value channel, the mean transition value becomes a one-step target for the critic at the current state. Real branch outcomes calibrate the weights of both channels. On ALFWorld and ScienceWorld, the resulting policy achieves higher success than PPO with comparable real interaction. It matches the strongest PPO refinements we tested on ALFWorld and exceeds the best of them by 6.5 points on ScienceWorld. Unlike distillation, \wmppo{} needs neither a teacher model nor privileged information, yet it also surpasses the teacher-based methods we tested. When deployed alone, the policy more often completes prerequisite steps before committing to an answer or placing an object.

\section*{Acknowledgements}
We thank the developers of ALFWorld and ScienceWorld for making their environments available, the Qwen team for releasing the base model, and the authors of the baseline methods for releasing their code.

\section*{Code and Data Availability}
The code, training configurations, evaluation scripts and per-game evaluation results are available at \url{https://github.com/nuskonghou/wm-ppo}, and the model checkpoints at \url{https://huggingface.co/nuskonghou/wm-ppo}. Both environments and the base model are publicly available, and Appendix~\ref{app:impl} describes the training and evaluation task sets.

\setlength{\bibsep}{3pt plus 1pt}   %
\bibliography{refs/references}
\bibliographystyle{icml2026}

\newpage
\appendix

\renewcommand{\floatpagefraction}{0.8}\renewcommand{\topfraction}{0.9}\renewcommand{\bottomfraction}{0.7}\renewcommand{\textfraction}{0.1}
\setcounter{topnumber}{3}\setcounter{bottomnumber}{2}\setcounter{totalnumber}{5}
\appendixfloatsafter   %
\makeatletter\setlength{\@fptop}{0pt}\setlength{\@fpbot}{0pt plus 1fil}\makeatother   %

\section{Prompts}
\label{app:prompts}

The policy and world model use chat prompts, each consisting of a system message followed by one user message. Fields in braces are filled at every step. The policy sees the five most recent observation--action pairs; the action is the last \texttt{<action>} tag of its response (ALFWorld: one of the admissible actions; ScienceWorld: a free-text command, judged by the simulator).

\begin{promptbox}{cpol}{Policy \textperiodcentered{} ALFWorld}
|System:| You are an expert agent operating in the ALFRED embodied household environment.
|User:|
Your task is to: {task}
Prior to this step, you have already taken {t} step(s). Below are the most recent {k} observations and the corresponding actions you took: [Observation {i}: '{o_i}', Action {i}: '{a_i}'] ...
You are now at step {t+1} and your current observation is: {obs}
Your admissible actions of the current situation are: [{admissible actions}].

Now it's your turn to take an action. First reason briefly (at most two sentences) about the current situation inside <think> </think> tags. Then choose one admissible action for the current step and present it inside <action> </action> tags.
\end{promptbox}
At the first step the history line is omitted and the observation is the initial room description.

\begin{promptbox}{cpol}{Policy \textperiodcentered{} ScienceWorld}
|System:| You are an expert agent operating in the ScienceWorld environment, a text-based simulator of elementary science experiments.
|User:|
Your task is to: {task}
Rules: 'focus on OBJ' declares which object the task is about and cannot be undone; focusing on a wrong object ends the task as a failure, so explore first and focus only on the object the task asks about. OBJ must be an object you can refer to now (listed below); LOC is a room name (teleport moves you there directly). Answering questions such as 'which one ...' is done by focusing on the answer object.
Prior to this step, you have already taken {t} step(s). Below are the most recent {k} observations and the corresponding actions you took:
[Observation {i}: '{o_i}', Action {i}: '{a_i}'] ...
You are now at step {t+1} and your current observation is:
{obs}
Action templates: [look around, look at OBJ, look in OBJ, inventory, task, teleport to LOC, go to LOC, open OBJ, close OBJ, activate OBJ, deactivate OBJ, pick up OBJ, put down OBJ, move OBJ to OBJ, pour OBJ in OBJ, dunk OBJ in OBJ, mix OBJ, use OBJ on OBJ, connect OBJ to OBJ, disconnect OBJ, eat OBJ, flush OBJ, read OBJ, focus on OBJ, wait, wait1].
Objects you can refer to: [{objects}].

Now it's your turn to take an action. First reason step by step (at most two sentences) inside <think> </think>, then give exactly one action inside <action> </action>.
\end{promptbox}

\begin{promptbox}{cwm}{World model \textperiodcentered{} ALFWorld}
|System:| You are a world model of the ALFRED text household environment.
|User:|
Task: {task}
Initial observation: {room}
History of steps so far: [Step {i}: observation '{o_i}', action '{a_i}'] ...
Current observation: {obs}
Admissible actions: [{admissible actions}]
Action taken now: {action}

Predict the environment's next observation exactly as the environment would print it, inside <obs> </obs>.
\end{promptbox}

\begin{promptbox}{cwm}{World model \textperiodcentered{} ScienceWorld}
|System:| You are a world model of the ScienceWorld text environment.
|User:|
Task: {task}
Recent actions: [Step {i}: action '{a_i}'] ...   (the last 8 actions)
Current observation:
{obs}
Objects the agent can refer to: [{objects}]
Action taken now: {action}

Predict the environment's next observation exactly as the environment would print it (action feedback, [Current room], [Inventory] and [score] lines), inside <obs> </obs>.
\end{promptbox}

The world model's target is the true next observation inside \texttt{<obs>} tags. In ScienceWorld, our environment interface appends \texttt{[task completed]} or \texttt{[task failed]} at an episode ending, which the world model can then predict. The critic receives the predicted successor in the policy's prompt format, with the prediction as its current observation.

\section{Implementation and method details}
\label{app:impl}

\FloatBarrier
\subsection{One training iteration}
\label{app:alg}

Algorithm~\ref{alg:wmppo-full} follows one iteration from real interaction through candidate evaluation to the updates of all three models.

\begin{algorithm}[H]
\caption{Full iteration of \wmppo{} (summarized in Algorithm~\ref{alg:wmppo}). Predictions and transition values are computed without gradients; each model is updated only by its own loss. Equation numbers refer to Section~\ref{sec:method}.}
\label{alg:wmppo-full}
\small
\newcommand{\PHASE}[1]{\item[]\textcolor{black!60}{$\triangleright$~\textit{#1}}}
\begin{algorithmic}[1]
\REQUIRE policy $\pol$, critic $\V$ and world model $\wm$ from the previous iteration; branch pairs $\mathcal P$ stored in the last $M$ iterations
\PHASE{Real interaction (Section~\ref{sec:method-loop})}
\FOR{each of 64 training tasks}
  \STATE play a chain with $\pol$; at each $s_t$, sample $K$ responses $y_t^{(k)}$ with actions $a_t^{(k)}$ and execute $a_t^{(1)}$
  \STATE play 3 branches: replay the chain to a uniformly drawn step, resample until the action differs from the chain's (at most eight times; otherwise no calibration pair), play to the end
\ENDFOR
\PHASE{Evaluating predicted transitions (Section~\ref{sec:method-eval})}
\FOR{each chain state $s_t$ with at least two distinct candidates}
  \STATE for each distinct $a$: \ $\hat o_{t+1}(a)\leftarrow\operatorname{greedy}\wm(\cdot\mid x_t,a)$, \ $\qhat_t(a)\leftarrow\V\bigl(C(s_t,a,\hat o_{t+1}(a))\bigr)$ \hfill Eq.~\eqref{eq:score}
  \STATE \quad ScienceWorld: if $\hat o_{t+1}(a)$ ends the episode, $\qhat_t(a)\leftarrow 1$ (completed) or $0$ (failed)
  \STATE $\hat a_t\leftarrow\arg\max_a\qhat_t(a)$, \quad margin $m_t\leftarrow\qhat_t(\hat a_t)-\max_{a\neq\hat a_t}\qhat_t(a)$; skip $s_t$ if $m_t=0$
  \STATE compute $\bar q_t$ over all $K$ samples, including duplicates \hfill Eq.~\eqref{eq:qbar}
\ENDFOR
\PHASE{Reliability gate (Section~\ref{sec:method-gate})}
\STATE margin bins $\ell=1,2,3$: split at the 0.5 and 0.8 quantiles of this iteration's nonzero margins $m_t$
\STATE value this iteration's branch pairs $(a_t^{\mathrm{ch}},a_t^{\mathrm{br}})$ in the same way and bin them; they only enter later weights
\STATE for each bin with $n_\ell\ge n_{\min}$ stored pairs in $\mathcal P$: \ $g_\ell\leftarrow\operatorname{mean}\bigl[\Delta G\,\operatorname{sign}(\Delta\qhat)\bigr]$ \hfill Eq.~\eqref{eq:gain}
\STATE \quad $w_\ell\leftarrow\operatorname{clip}\bigl(\tfrac{n_\ell}{n_\ell+n_0}\cdot\tfrac{g_\ell}{g_{\mathrm{ref}}},\,0,\,1\bigr)$; other bins: $w_\ell\leftarrow 0$; iterations 13--14: start-up rule of Table~\ref{tab:hparams} \hfill Eq.~\eqref{eq:gate}
\PHASE{Updates (Sections~\ref{sec:method-update} and~\ref{sec:method-loop})}
\STATE each sample $y_t^{(k)}$ at an evaluated $s_t$ with $w_{\ell(m_t)}>0$: \ $A^{\mathrm{aux}}_{t,k}\leftarrow \nu\,w_{\ell(m_t)}(z_{t,k}-\bar z_t)$ \hfill Eq.~\eqref{eq:aux}
\STATE each executed response: $A^{\mathrm{GAE}}_t$ by tree GAE on real returns, whitened to $\hat A_t$ \hfill Eq.~\eqref{eq:tree-gae}
\STATE policy $\pol$: one PPO-clip loss over real rows $(s_t,y_t,\hat A_t)$ and auxiliary rows $(s_t,y_t^{(k)},A^{\mathrm{aux}}_{t,k})$ \hfill Eq.~\eqref{eq:loss}
\STATE critic $\V$: regression on returns plus $\beta\,w_{\ell(m_t)}\bigl(\V(s_t)-\operatorname{sg}[\bar q_t]\bigr)^2$ at evaluated states \hfill Eqs.~\eqref{eq:cons}, \eqref{eq:critic}
\STATE world model $\wm$: next-observation loss on this iteration's executed transitions (run in parallel) \hfill Eq.~\eqref{eq:wm}
\STATE add the new branch pairs to $\mathcal P$; the updated models are used from the next iteration
\end{algorithmic}
\end{algorithm}

\paragraph{Order within an iteration.} All data collection and candidate evaluation in an iteration use the models from the preceding iteration. The world-model update on the iteration's executed transitions runs in parallel with the policy and critic updates and replaces the served model when it finishes. Within the trainer, the critic first values every real node and every predicted successor, without gradients. The gate weights, auxiliary rows and auxiliary critic targets $\bar q_t$ are then computed, the policy is updated on ordinary and auxiliary rows, and the critic is updated last. The branch pairs of an iteration are valued by the same pre-update critic, binned with that iteration's thresholds, and used from the next iteration on.

\FloatBarrier
\subsection{Advantage estimation on the trees}
\label{app:gae}

\paragraph{The critic's regression.} Besides the regression of Equation~\eqref{eq:critic}, the implementation adds a sibling-centered term at the nodes $s$ with two or more executed children $s^{(c)}$, $c=1,\dots,C_s$:
\begin{equation*}
\sum_{c=1}^{C_s}\Bigl(\bigl(\V(s^{(c)})-\overline V(s)\bigr)-\bigl(G(s^{(c)})-\overline G(s)\bigr)\Bigr)^2,
\end{equation*}
where $\overline V(s)$ and $\overline G(s)$ average over the children of $s$, and a child that ends the episode enters with its final reward in place of a value. The term, with weight 1, asks that the centered values of the children match their centered returns, which is the within-state comparison that Equation~\eqref{eq:delta} relies on. Wide-Tree-PPO and \qppo{} use the same term.

\paragraph{The tree estimator.} The raw estimates $A^{\mathrm{GAE}}$ follow Equation~\eqref{eq:tree-gae} on the tree of a task's episodes, whose edges are executed (state, response) pairs; the ordinary PPO rows use their whitened values $\hat A_t$. $R$ is the episode's final reward, 0 also for an episode cut at the 40-step limit, without bootstrapping. If $s_{t+1}$ has a single outgoing edge, $\overline A^{\mathrm{GAE}}(s_{t+1})=A^{\mathrm{GAE}}_{t+1}$ and Equation~\eqref{eq:tree-gae} is the usual GAE recursion $A^{\mathrm{GAE}}_t=\delta_t+\lambda A^{\mathrm{GAE}}_{t+1}$ with $\delta_t=\V(s_{t+1})-\V(s_t)$.

\paragraph{Two executed actions at the same state.} For two executed actions with non-terminal successors $s_{t+1}$ and $s'_{t+1}$, Equation~\eqref{eq:tree-gae} gives the difference between their raw advantages,
\begin{equation}
\bigl[\V(s_{t+1})-\V(s'_{t+1})\bigr]+\lambda\bigl[\overline A^{\mathrm{GAE}}(s_{t+1})-\overline A^{\mathrm{GAE}}(s'_{t+1})\bigr].
\label{eq:gae-pair}
\end{equation}
The first term compares the successor values, and the second carries feedback from their continuations.

\paragraph{A terminated path.} Let the actions at $s_t,\dots,s_{t+L-1}$ lead to termination after the last of them, with reward $R$ at that step and zero before, and write $V_j=\V(s_{t+j})$. Then
\begin{equation}
\begin{aligned}
A^{\mathrm{GAE}}_t&=(V_1-V_0)+\lambda(V_2-V_1)+\dots+\lambda^{L-2}(V_{L-1}-V_{L-2})+\lambda^{L-1}(R-V_{L-1})\\
&=-V_0+(1-\lambda)\sum_{j=1}^{L-1}\lambda^{j-1}V_j+\lambda^{L-1}R .
\end{aligned}
\label{eq:gae-expanded}
\end{equation}
The executed continuation, summarized by a $\lambda$-weighted average of the critic's values along it and the final reward, is compared with the value of the state in which the action was taken. For $L=1$ the sum is empty and $A^{\mathrm{GAE}}_t=R-V_0$; for $\lambda=1$ the intermediate values cancel and $A^{\mathrm{GAE}}_t=R-V_0$, the Monte-Carlo return minus the baseline. The coefficients of the intermediate values sum to $1-\lambda^{L-1}$, so with $\lambda=0.95$ and $L=36$ the final reward has weight 0.17 and the values along the path 0.83 together, although each single value has a small weight. This follows directly from the definition of GAE~\citep{gae}.

\paragraph{GAE and candidate comparisons.} With the true policy value function and an on-policy continuation, the conditional expectation of $A^{\mathrm{GAE}}_t$ is the executed action's true advantage. How informative a sampled estimate is depends on the variances and covariances of the critic's errors and of the outcome, rather than on the coefficients alone. PPO already provides step-level feedback through its critic and GAE, using state values and returns along the executed paths. The single-path expression in Equation~\eqref{eq:gae-expanded} compares the executed continuation with $V_0=\V(s_t)$. \wmppo{} retains these advantages and extends evaluation to other candidates at the same state through their predicted successors. Their transition values supply direct preferences for the auxiliary policy update. In the policy-channel-only variant, these preferences are the additional policy feedback; critic training and ordinary PPO advantages remain standard.

\paragraph{Equal terminal returns.} In Wide-Tree-PPO on ScienceWorld, all executed candidates have the same terminal return at 82\% of its branching states. Those observed outcomes provide no preference among the candidates, although their immediate consequences can differ in whether they advance the task. Appendix~\ref{app:io} gives a training example in which both continuations succeed but the transition values favor completing the task's prerequisite first. \wmppo{} (policy channel only) outperforms this branch-only control in both environments (Section~\ref{sec:learning}).

\paragraph{Value and advantage scales.} On candidate pairs whose estimated true values differ by at least 0.4, the critics' value differences average 0.02--0.10 in absolute value (Appendix~\ref{app:truth}), and the raw advantages of PPO-160 have a median standard deviation of 0.22--0.23 per iteration on ScienceWorld and 0.27--0.30 on ALFWorld before whitening. These numbers indicate the scales involved; they are sampled and normalized differently.

\FloatBarrier
\subsection{Settings and components}

Table~\ref{tab:hparams} lists the hyperparameters. Table~\ref{tab:roles} identifies each component's data, what it learns or computes, and how its output is used.

\begin{table}[!ht]
\centering\small
\caption{\textbf{The components of \wmppo{} and their data.} The three models are LoRA adapters on one frozen base model; the gate has no learned parameters. Only real transitions train the world model and supply returns; predictions enter learning only through transition values, and only the policy's own loss sends gradients to the policy.}
\label{tab:roles}
\begin{tabularx}{\linewidth}{@{}l >{\raggedright\arraybackslash}X >{\raggedright\arraybackslash}X >{\raggedright\arraybackslash}X@{}}
\toprule
Component & Learns or computes & From & Used for \\
\midrule
Policy $\pol$ & action distribution given the state & ordinary PPO rows of real executed steps (tree GAE) and auxiliary rows of evaluated candidates & acting; alone at test time \\
World model $\wm$ & next observation after an action & cross-entropy on real executed transitions of chains and branches, every iteration & predicting the consequences of the $K$ candidates \\
Critic $\V$ & expected task return from a state & Monte-Carlo returns of the trees, the sibling-centered term, and, at each evaluated state $s_t$, the mean transition value $\bar q_t$ of its candidates (value channel) & GAE of real steps; transition values $\qhat_t(a)$ of predicted successors \\
Gate & weight $w_\ell$ per margin bin & real branch pairs of the previous $M$ iterations & scaling the auxiliary advantages and the value-channel loss \\
\bottomrule
\end{tabularx}
\end{table}

\paragraph{Task sets.} ScienceWorld training draws up to 120 official training variations per task type (1,228 of 2,106) in a fixed shuffled order with equal weight per task type. The evaluation set is fixed. Within each of the development and test splits, it contains all variations of the seven task types that have few variations and nine evenly spaced variations of every other type: 140 of the 1,053 development and 144 of the 1,060 test variations. Three settings of \wmppo{} are environment-specific (Table~\ref{tab:hparams}): the prompts, the world model's context and output length, and the rule for predicted endings, which applies only where observations report the end of an episode. Among the baselines' settings, only those of CAPO, \sopo{} and PaW differ between the environments, because their ALFWorld settings diverged or lost the action format on ScienceWorld (Appendix~\ref{app:baselines}).

\begin{table}[!ht]
\centering\small
\caption{Hyperparameters of \wmppo{} (identical in both environments unless noted).}
\label{tab:hparams}
\begin{tabular}{l p{0.7\linewidth}}
\toprule
Base model & Qwen3-4B-Instruct-2507, non-thinking mode, frozen \\
Adapters & policy, critic, world model: separate LoRA, rank 64, $\alpha=128$, all linear layers \\
Critic head & separate LoRA with a linear value head on the last prompt token, zero-initialized; inputs left-truncated to 8,192 tokens \\
Rollouts & temperature 1, at most 256 response tokens, at most 40 actions per episode \\
Tasks per iteration & 64; one chain and three branches each \\
Candidates & $K=4$ per chain step, first one executed \\
Branches & branch step uniform over the chain; up to 8 samples to obtain a different action \\
Policy optimizer & AdamW, lr $10^{-5}$, betas $(0.9,0.99)$, gradient clipping 1.0, 64 rows per minibatch, one epoch \\
PPO clip & $\epsilon_{\mathrm{lo}}=0.2$, $\epsilon_{\mathrm{hi}}=0.28$; token-level normalization per minibatch \\
Advantages & tree GAE (Appendix~\ref{app:gae}), $\gamma=1$, $\lambda=0.95$, no bootstrap at the step limit, whitened over executed steps \\
Critic optimizer & AdamW, lr $5\times10^{-5}$; target: mean final reward below the node; sibling term weight 1; value-channel weight $\beta=0.5$ (Equation~\eqref{eq:cons}) \\
Auxiliary rows & Eq.~\eqref{eq:aux}, $\nu=1$; decision points with $\ge2$ distinct candidates; exact ties carry no preference (checked in ScienceWorld; none occurred in ALFWorld) \\
Margin bins & 0.5 and 0.8 quantiles of the iteration's nonzero margins $m_t$ \\
Reliability gate & pairs of the previous $M=4$ iterations; $n_{\min}=20$ pairs per bin; $n_0=40$, $g_{\mathrm{ref}}=0.05$ (a shrinkage estimate of the bin's gain, Equation~\eqref{eq:gate}, scaled so that a shrunken gain of 0.05 per decision earns full weight; set once from the first branch pairs of a preliminary ALFWorld run and used unchanged in both environments) \\
Gate, iterations 13--14 & iteration 13: no stored pairs, all $w_\ell=0$; iteration 14: $w_\ell=\max(g_\ell-1.64\,\mathrm{se}_\ell,0)/\max_{\ell'}\max(g_{\ell'}-1.64\,\mathrm{se}_{\ell'},0)$; bins with fewer than $n_{\min}$ pairs count as 0, and if no bound is positive all weights are 0 (in 9 of the 12 runs of the two \wmppo{} variants all iteration-14 weights were 0; in the other three one or two bins opened), the rule of the preliminary run, kept for comparability \\
World-model input & goal, history (ALFWorld: all previous observations and actions; ScienceWorld: the last 8 actions, without observations), \\
 & current observation, admissible actions or referable objects, the action \\
World-model output & the next observation as the environment prints it; greedy decoding \\
 & (at most 300 tokens in ALFWorld, 1,024 in ScienceWorld) \\
Initial world model & executed transitions of warm-up iterations 1--12 (ALFWorld 29,030, ScienceWorld 10,291), trained at the end of iteration 12; lr $10^{-4}$, one epoch, 32 transitions per step. \\
Online world model & this iteration's executed transitions; lr $5\times10^{-5}$, one epoch, continues from the current model \\
Predicted endings & ScienceWorld: predicted completion or failure replaces the critic's value by 1 or 0 \\
Evaluation & policy only; one dedicated inference server per checkpoint; greedy and 3 runs at temperature 1 \\
\bottomrule
\end{tabular}
\end{table}

\section{Diagnostic experiments}
\label{app:diag}

The diagnostics in Section~\ref{sec:diagnostics} test whether the critic can compare actions by valuing their real successors, whether predictions preserve these comparisons, and whether the policy can learn from them. All use the policy and critic of the PPO run continued from the common warm-up to iteration 24 on ALFWorld (Section~\ref{sec:diagnostics}), the 274 evaluation games and per-game paired tests.

\paragraph{Selection with true next states.} At every step, the policy samples four candidates at temperature 1. Each distinct candidate is executed for one step in a copy of the game. The critic values the history extended by the resulting observation and admissible-action list, and the candidate with the highest value is executed in the real game. In the \emph{peek} control, a candidate that completes the task is taken if one exists; otherwise the majority vote decides.

\paragraph{Pairwise test.} In each evaluation game, one state of the policy's own episode is drawn at random; the draw is repeated with three random seeds. Every distinct candidate is executed and followed by three independent continuations of the same policy, and the fraction of successes estimates its value. For the 195 candidate pairs with different estimates, this order is compared with the order of the critic's values of their true next states.

\paragraph{World model for the diagnostic.} The policy played 1,000 training games at temperature 1, executing the first of four samples; every distinct candidate was also executed for one step in a copy of the game, which gives 48k transitions, 44\% of them for unexecuted candidates. The world model is a LoRA adapter on the base model, trained for one epoch with the next-observation loss. Accuracy is measured on 1,500 random transitions of the evaluation games, by category of the true next observation: observations determined by information already seen (71\% of the transitions; 99.7\% predicted exactly), observations revealing objects not seen before (20\%; 26\% predicted exactly, object F1 0.41), and ``Nothing happens'' (9\%). Choosing with the true observation from which newly revealed objects are removed gives 59.1\%, so hidden content affects selection. Where it matters, the world model still provides a prior: it estimates whether the target object is present with an AUC of 0.81 (untrained model: 0.62).

\paragraph{Offline update.} The policy played 500 training games that are not in the world model's data. At the 8,042 decision points with at least two distinct candidates, the label is the candidate whose predicted successor the critic values highest. One epoch of PPO-clip (lr $10^{-5}$, about 32k samples) uses the advantages of Equation~\eqref{eq:offline-adv} and no task reward. The KL divergence of the updated policy from the original is 0.017 with world-model labels and 0.033 with majority-vote labels. The evaluation samples at temperature 1, twice per game, without the world model.

\paragraph{Inference-time lookahead on trained policies.}\label{app:lookahead} Table~\ref{tab:lookahead} applies the selection test of Section~\ref{sec:diag-real} to trained ALFWorld policies on the 274 evaluation games. At every step, the policy samples four responses at temperature 1. \emph{pass@1} executes the first, which is the standard evaluation (three runs). \emph{Majority of 4} executes the most frequent action. \emph{True next state} executes each distinct candidate for one step in a copy of the game, values each extended history with the policy's own critic of the same iteration, and executes the candidate with the highest value. \emph{Predicted next state} does the same with the next observation predicted greedily by the world model of the diagnostic in Section~\ref{sec:diag-pred}, which is not updated during RL. Each lookahead rule has two runs, and differences are paired per game against pass@1.

\begin{table}[!ht]
\centering\scriptsize
\caption{Inference-time lookahead on trained ALFWorld policies (success in \%; pass@1: the policy's own first sample; in parentheses, the change against pass@1 in percentage points; $^\dagger$: $p<0.05$). One training run per policy (the first run of each method).}
\label{tab:lookahead}
\setlength{\tabcolsep}{3.5pt}
\small\begin{tabular}{l l c c c c}
\toprule
It. & Policy & pass@1 & Majority of 4 & True next state & Predicted next state \\
\midrule
24 & PPO-160 & 68.9 & 68.8 (-0.1) & 74.6 (+5.8$^\dagger$) & 73.9 (+5.0$^\dagger$) \\
 & PPO-256 & 72.6 & 69.7 (-2.9) & 81.0 (+8.4$^\dagger$) & 77.2 (+4.6$^\dagger$) \\
 & Wide-Tree-PPO & 64.0 & 59.5 (-4.5$^\dagger$) & 79.0 (+15.0$^\dagger$) & 76.1 (+12.1$^\dagger$) \\
\midrule
36 & PPO-160 & 77.4 & 76.6 (-0.7) & 84.9 (+7.5$^\dagger$) & 82.3 (+4.9$^\dagger$) \\
 & PPO-256 & 86.1 & 85.4 (-0.7) & 90.1 (+4.0$^\dagger$) & 89.2 (+3.1$^\dagger$) \\
 & Wide-Tree-PPO & 83.8 & 83.0 (-0.8) & 91.1 (+7.2$^\dagger$) & 86.9 (+3.0$^\dagger$) \\
\midrule
55 & PPO-160 & 89.8 & 88.7 (-1.1) & 94.0 (+4.2$^\dagger$) & 92.5 (+2.7$^\dagger$) \\
 & PPO-256 & 91.0 & 90.7 (-0.3) & 94.9 (+3.9$^\dagger$) & 93.2 (+2.3$^\dagger$) \\
 & Wide-Tree-PPO & 89.2 & 89.2 (+0.1) & 94.2 (+5.0$^\dagger$) & 93.6 (+4.4$^\dagger$) \\
\bottomrule
\end{tabular}

\end{table}

\section{Baselines}
\label{app:baselines}

\FloatBarrier
\subsection{Budgets, methods and tuning}

\paragraph{Budgets.} PPO-160: 160 tasks $\times$ 1 episode per iteration; PPO-256: 256 tasks $\times$ 1; Wide-Tree-PPO, \qppo{} and Fixed WM: \wmppo{}'s trees (64 tasks $\times$ (1 chain + 3 branches)); group methods: 20 tasks $\times$ 8 episodes, also 40 tasks $\times$ 4 for GRPO and, on ScienceWorld, GiGPO (Tree-GRPO: \wmppo{}'s trees or 32 trees of 8 leaves, Table~\ref{tab:tuning}); PPO refinements and the auxiliary-feedback methods: PPO-160's sampling. All start from the warm-up checkpoint with the same task order.

\paragraph{Methods.} \emph{Wide-Tree-PPO}: \wmppo{} (policy channel only) without the auxiliary term, that is, the same trees, tree GAE, critic targets and sibling term, and the same ordinary PPO rows of the executed steps. \emph{\qppo{}}: a third LoRA adapter with a linear head reads the history followed by a candidate response and is regressed on the return below each executed step, plus the sibling term; it is initialized from the warm-up critic and replaces $\qhat$ in Equation~\eqref{eq:score}; candidates, margins, gate and auxiliary rows are unchanged. \emph{Fixed WM}: \wmppo{} (policy channel only) with a world model that is never updated after its initialization; the policy and the critic are trained, and the gate and the auxiliary rows computed, as in that variant. In both environments it starts from the same initial world model as \wmppo{}, trained on the warm-up's executed transitions (on ScienceWorld, predictions at iteration 13 are token-identical for 78 of 81 shared state--candidate pairs). \emph{CAPO}~\citep{capo}: action-level critic and importance ratio. \emph{\sopo{}}~\citep{stppo}: one geometric-mean importance ratio per turn with clipping-triggered normalization. \emph{\steppo{}}: the GSPO-style turn-level ratio~\citep{gspo} with a narrow clip, as implemented in the CAPO code base~\citep{capo}. \emph{VAPO-lite}~\citep{vapo}: PPO-160 plus a language-model loss (weight 0.1) on responses of successful episodes. \emph{GiGPO}~\citep{gigpo} and \emph{HGPO}~\citep{hgpo}: step-level grouping by identical environment states and by history, ported from their released code. \emph{Tree-GRPO}~\citep{treegrpo}: tree-based group advantages on \wmppo{}'s trees. \emph{Majority-vote labels}: PPO-160 plus four samples per step and auxiliary rows as in Equation~\eqref{eq:aux} with the unique plurality action as the preferred one, weight 1, no gate; computed on PPO-160's episodes, their total auxiliary weight is about two to three times that of \wmppo{}. \emph{Standard OPD}~\citep{gkd}: a frozen Qwen3-32B in non-thinking mode, which shares the student's vocabulary, computes the log-probabilities of the executed response given exactly the student's prompt; every token's advantage is the whitened GAE advantage plus $\operatorname{clip}(\log\pi_T-\log\pi_{\mathrm{old}},\pm5)$, the reverse-KL reward of on-policy distillation, with coefficient 1. \emph{OCSD}~\citep{ocsd}: the teacher, the policy of the previous iteration (self) or the same Qwen3-32B, computes the log-probability of the executed response with and without the next one or two observations of the episode, but not the current action; the PPO advantage is multiplied by $1+0.5\,\operatorname{sign}(A)\tanh(\Delta/2)$, where $\Delta$ is the difference of the two log-probabilities, on the 20\% least likely steps per episode, following the released implementation. All OPD and OCSD variants are built on PPO-160 and use its interaction. \emph{PaW}~\citep{paw}: PPO-160 plus PaW's auxiliary next-observation loss on the policy with the published settings: the clipped loss with $\rho=0.2$ on the tokens of the true next observation, applied to the 75\% of transitions with the highest action uncertainty, the reward-adaptive weight following the paper's formula (with the batch's mean episode return under PPO), and the paper's 20 prediction prompts. Unlike the original full fine-tuning at learning rate $10^{-6}$, it trains PPO-160's LoRA at $10^{-5}$.

\paragraph{Published results.} Several of these methods report higher ALFWorld success in their own papers. Those results come from different base models, longer training and other budgets. Here, every method shares one base model, one warm-up checkpoint, one task order and a comparable interaction budget, and is reported at fixed iterations.

\paragraph{Tuning rules fixed in advance.} With the published clipping ranges, CAPO (0.001) clipped about 97\% of tokens and \steppo{} ($3\times10^{-4}$) about 85\% on ALFWorld, so both barely learned (Table~\ref{tab:tuning}). For each of the two, three larger clipping values were tried for one update on the iteration-13 data. The rule chose the smallest value whose clipped-token fraction was at most 20\%, or the largest value if none qualified: 0.3 for CAPO and 0.1 for \steppo{}. On ScienceWorld every method first used its ALFWorld setting unchanged, and only a setting that diverged or lost the action format was re-tuned (CAPO, \sopo{} and PaW); \wmppo{} keeps its ALFWorld hyperparameters unchanged. Table~\ref{tab:tuning-sw} lists the ScienceWorld settings.

\paragraph{\sopo{} on ScienceWorld.} On ScienceWorld, \sopo{} with its published clip of 0.2 had a per-iteration KL of 0.07--0.14 from the first update (PPO-160: 0.01--0.02) and collapsed after iteration 36 (15.8\% at iteration 60). The rule fixed beforehand was to take the largest of 0.1, 0.05 and 0.02 whose KL after one update on the iteration-13 data was at most 0.02, or the smallest if none qualified. All three exceeded the threshold slightly (0.054, 0.035 and 0.022), so 0.02 was used. For \sopo{} we also ran the clip of 0.1 chosen by the clipped-token rule of CAPO and \steppo{}. This run did not diverge (per-update KL at most 0.08), but its success peaked at 22.9\% among the evaluated checkpoints (iteration 36) and fell to 4.2\% at iteration 60, when 92\% of its evaluation episodes ended at the step limit. The main table reports the best of the three settings, clip 0.02. The same \sopo{} with clip 0.2 did not collapse on ALFWorld and is reported with the published value there.

\paragraph{CAPO on ScienceWorld.} CAPO with its ALFWorld clip of 0.3 diverged on ScienceWorld at iteration 14, and with 0.1 at iteration 28 (per-update KL spikes of 0.15--0.33, zero success); the last value of the pre-set grid, 0.03, diverged at iteration 18. No setting of the grid remains, so the main table reports, for CAPO on ScienceWorld, the official evaluation of clip 0.1 at iteration 18, the checkpoint with the highest greedy success in the evaluations during training (21.5\%; 20.1\% at iteration 15, 16.9\% at 21 and 9.5\% at 24), without a test against \wmppo{}. In the official evaluation it reaches 21.1\% greedy and 20.0\% pass@1 success, after 12.8k real steps; the neighboring checkpoints reach 19.0\% (iteration 15) and 19.4\% (iteration 21) greedy, so the choice does not depend on which evaluation is used.

\paragraph{\steppo{} and PaW on ScienceWorld.} \steppo{} runs on ScienceWorld with its ALFWorld clip of 0.1. It did not diverge (per-update KL of 0.02--0.06, three to five times that of PPO-160), but its success stayed near 25\% from iteration 21 on, below PPO-160; because only diverged settings were re-tuned, it keeps this setting. PaW with its published coefficient of 1.0 kept a per-update KL close to that of PPO-160 on ScienceWorld but gradually stopped writing the action tag: the share of training responses without it rose from 0.8\% at iteration 16 to 78\% at iteration 20, and 67\% of its greedy evaluation episodes at iteration 60 end at the step limit; with the coefficients 0.5 and 0.25, the share stays at about 1\%, as for PPO-160. Before any result of the two runs, we fixed the coefficients 0.5 and 0.25 and the rule that the main table reports the best of the three settings by greedy success at iteration 60, which is 0.5. On ALFWorld, PaW uses the published coefficient.

\begin{table}[!ht]
\centering\small
\caption{Baseline settings tried on ALFWorld (iteration 36, policy-only, greedy and sampling success in \%). The main table uses the setting chosen by the rule.}
\label{tab:tuning}
\begin{tabular}{l l c c}
\toprule
Baseline & Setting & Greedy & pass@1 \\
\midrule
CAPO & action-ratio clip 0.001 (original) & 46.0 & 63.9 \\
 & clip 0.1 & 82.8 & 84.1 \\
 & clip 0.3 (selected by the pre-set rule; main table) & 87.2 & 86.7 \\
\steppo{} & clip $3\times10^{-4}$ (released code) & 56.6 & 74.6 \\
 & clip 0.1 (selected) & 85.8 & 85.0 \\
GRPO & 40 tasks $\times$ 4 & 54.4 & 60.0 \\
 & 20 tasks $\times$ 8 (main table) & 59.5 & 67.8 \\
Tree-GRPO & 32 trees $\times$ 8 leaves & 47.4 & 60.2 \\
 & 64 trees $\times$ 4 leaves & 48.9 & 59.6 \\
\bottomrule
\end{tabular}

\end{table}

\begin{table}[!ht]
\centering\small
\caption{Baseline settings on ScienceWorld (policy-only success in \%; iteration 60 unless stated). Every method first used its ALFWorld setting; only settings that diverged or lost the action format were re-tuned by the rules fixed in advance.}
\label{tab:tuning-sw}
\setlength{\tabcolsep}{3.5pt}
\begin{tabular}{l l l c c}
\toprule
Baseline & Setting & Outcome & Greedy & pass@1 \\
\midrule
CAPO & clip 0.3 (ALFWorld setting) & diverged, it.~14 & -- & -- \\
 & clip 0.1 & diverged, it.~28; it.~18 in Table~\ref{tab:main} & 21.1 & 20.0 \\
 & clip 0.03 & diverged, it.~18 & -- & -- \\
\sopo{} & clip 0.2 (published) & collapsed after it.~36 & 15.8 & 16.2 \\
 & clip 0.02 (KL rule; Table~\ref{tab:main}) & it.~60 & 46.5 & 49.4 \\
 & clip 0.1 (clipped-token rule) & no divergence; it.~60 (best 22.9\%, it.~36) & 4.2 & 10.8 \\
\steppo{} & clip 0.1 (ALFWorld setting) & it.~60 & 25.0 & 26.5 \\
PaW & coefficient 1.0 (published) & action tags lost from it.~20 & 6.0 & 30.8 \\
 & coefficient 0.5 (Table~\ref{tab:main}) & it.~60 & 35.6 & 38.8 \\
 & coefficient 0.25 & it.~60 & 32.4 & 32.9 \\
\bottomrule
\end{tabular}

\end{table}

\FloatBarrier
\subsection{Teacher-based feedback}
\label{app:teacher}

\paragraph{How strong is the teacher?} Table~\ref{tab:teacher} evaluates candidate teachers as agents in the student's prompt and action format on the evaluation games. Under greedy decoding and strict parsing of the \texttt{<action>} tag, Qwen3-32B solves 13.5\% of the ALFWorld games and 26.4\% of the ScienceWorld games. On ALFWorld it often writes its action in its own tool-call format; counting those as actions raises its success to 48.5\%. On ScienceWorld, in non-thinking mode, it tends to act without reasoning and to focus early on a wrong object. Qwen3-30B-A3B-Instruct, a second candidate, performs at the same level. The teacher was fixed before these evaluations, which serve only to report its level; no available model of this family was clearly stronger than the trained student, so we did not add a second external teacher.

\begin{table}[!ht]
\centering\small
\caption{Candidate teachers as agents (success in \%, evaluation games, student's prompt format).}
\label{tab:teacher}
\begin{tabular}{l c c}
\toprule
Model & ALFWorld & ScienceWorld \\
\midrule
Student at the common start (iteration 12) & 27.0 & 20.8 \\
\wmppo{} after training (iteration 36 / 60, greedy) & 87.5 & 52.9 \\
Qwen3-32B, non-thinking, greedy (the teacher used) & 13.5 & 26.4 \\
\quad sampled at temperature 1, three runs & 35.0 / 34.0 / 33.2 & -- \\
\quad tool-call format also accepted as an action & 48.5 & -- \\
Qwen3-30B-A3B-Instruct-2507, greedy & 28.8 & 25.4 \\
\bottomrule
\end{tabular}
\end{table}

\paragraph{What OCSD's judgment tracks.} For every step that OCSD selected during the first self-teacher run, we averaged $\tanh(\Delta/2)$ over the response tokens, which is OCSD's judgment of the step, and compared it with labels taken from the true next observation. On ALFWorld (10,019 steps), this judgment separates valid actions from those answered by \emph{Nothing happens} with an AUC of 0.80, and the adjustment concentrates on the action tokens. On ScienceWorld (8,055 steps), the same judgment is inverted: its AUC is 0.25 for valid against rejected commands and 0.21 for steps that raised the game's score against steps that did not, and the strongest negative adjustments fall on the reasoning tokens of steps that made progress, which are followed by observations with much new content. The adjustment is bounded and keeps the sign of the advantage; on ScienceWorld, the resulting policy is level with PPO-160 for pass@1.

\paragraph{Standard OPD in training.} On ALFWorld, the student's responses changed abruptly at the start. The KL divergence from the previous policy was 1.33 in iteration 13; in iteration 14, only 0.5\% of the responses contained a parsable action; and the format recovered in iteration 15. The teacher prefers its own tool-call format and suppressed the student's action tags. On ScienceWorld, the teacher penalized the student's reasoning heavily (about 63 nats per response early on), and the median response length fell from 77 tokens to 12 by iteration 16. The student led PPO-160 at iteration 36 and then stayed at 31--34\% pass@1; the teacher itself reaches 26.4\% as a greedy agent (Table~\ref{tab:teacher}).

\section{Complete results}
\label{app:full}

This appendix reports results for every method at every evaluated iteration (Appendix~\ref{app:full-iter}), the learning curves and the real steps to reference success levels (Appendix~\ref{app:curves}), the individual runs, the ScienceWorld task types and the costs per iteration. Marks are as in Table~\ref{tab:main}; a number in parentheses after a method is its number of runs.

\FloatBarrier
\subsection{Results at every evaluated iteration}
\label{app:full-iter}

Tables~\ref{tab:full-alf} and~\ref{tab:full-sw} list every method at every evaluated iteration, with the real steps up to that iteration and pass@3. On ALFWorld, the methods trained to iteration 55 approach the ceiling of the benchmark. There, \wmppo{} succeeds on 88.7\% (greedy) and 90.8\% (pass@1) and \wmppo{} (policy channel only) on 91.4\% and 92.9\%; PPO-160 reaches 85.0\% and 87.3\%, PPO-256, with about 1.5 times the real steps, 90.8\% and 91.1\%, Wide-Tree-PPO (three runs) 87.3\% and 90.6\%, Fixed WM 93.4\% and 92.2\%, and PaW 93.1\% and 91.8\%. With this much training, the differences on ALFWorld lie in how quickly each method gets there (Figure~\ref{fig:curves}, Table~\ref{tab:reach}) rather than in the final level. GRPO with 40 tasks of 4 episodes per iteration, which sees about twice as many tasks as with 20 tasks of 8 episodes, also stays below \wmppo{} in both environments, and so does GiGPO on ScienceWorld. On ScienceWorld at iteration 60, \sopo{} has a slightly higher pass@3 than \wmppo{} (60.9\% against 59.6\%).

\begin{table}[!ht]
\centering\footnotesize
\caption{ALFWorld, all methods, iterations 24 and 36 (policy-only success in \%; iteration 55 in Table~\ref{tab:full-alf-b}). $^\dagger$/$^\ddagger$: significantly lower/higher than \wmppo{} at that iteration (per-game paired test, $p<0.05$, uncorrected). $^{s}$Second run of an existing seed, launched later with the same configuration; every mean and test in the paper uses the first run of each seed.}
\label{tab:full-alf}
\setlength{\tabcolsep}{2pt}
\begin{tabular}{l r c c c r c c c }
\toprule
 & \multicolumn{4}{c}{iteration 24} & \multicolumn{4}{c}{iteration 36} \\
\cmidrule(lr){2-5}\cmidrule(lr){6-9}
Method & Steps & Greedy & pass@1 & pass@3 & Steps & Greedy & pass@1 & pass@3 \\
\midrule
\wmppo{} (ours) (3) & 60k & 55.7 & 60.8 & 73.5 & 102k & 87.5 & 87.8 & 93.4 \\
\wmppo{} (policy channel only) (3) & 58k & 68.4\sigdn{} & 70.9\sigdn{} & 82.5 & 97k & 88.4 & 89.4 & 94.6 \\
PPO-160 (3) & 57k & 47.1\sigup{} & 54.3\sigup{} & 67.9 & 105k & 63.5\sigup{} & 71.0\sigup{} & 81.1 \\
PPO-256 (3) & 88k & 63.6\sigdn{} & 69.3\sigdn{} & 81.9 & 150k & 83.1\sigup{} & 84.8\sigup{} & 92.9 \\
GRPO & 58k & 53.3 & 57.2\sigup{} & 73.0 & 113k & 59.5\sigup{} & 67.8\sigup{} & 83.6 \\
GRPO, 40 tasks $\times$ 4 & 60k & 36.5\sigup{} & 42.6\sigup{} & 56.6 & 113k & 54.4\sigup{} & 60.0\sigup{} & 71.2 \\
GiGPO & 60k & 44.9\sigup{} & 54.4\sigup{} & 65.7 & 109k & 78.1\sigup{} & 81.0\sigup{} & 90.9 \\
HGPO & 56k & 52.6 & 59.4 & 68.2 & 103k & 65.0\sigup{} & 70.8\sigup{} & 81.0 \\
Tree-GRPO & 66k & 43.4\sigup{} & 47.8\sigup{} & 69.0 & 124k & 48.9\sigup{} & 59.6\sigup{} & 74.8 \\
CAPO & 54k & 78.5\sigdn{} & 80.5\sigdn{} & 87.6 & 88k & 87.2 & 86.7 & 92.0 \\
\sopo{} & 55k & 74.8\sigdn{} & 74.0\sigdn{} & 89.8 & 100k & 86.5 & 83.1\sigup{} & 89.8 \\
\steppo{} & 54k & 71.5\sigdn{} & 74.6\sigdn{} & 88.3 & 91k & 85.8 & 85.0 & 92.7 \\
VAPO-lite & 56k & 65.0\sigdn{} & 66.8\sigdn{} & 80.7 & 100k & 68.2\sigup{} & 80.2\sigup{} & 90.9 \\
\qppo{} (3) & 60k & 63.7\sigdn{} & 68.3\sigdn{} & 81.1 & 100k & 84.4\sigup{} & 88.6 & 94.3 \\
Wide-Tree-PPO (3) & 61k & 54.0 & 63.3\sigdn{} & 76.5 & 106k & 78.5\sigup{} & 82.3\sigup{} & 90.6 \\
Fixed WM & 59k & 67.2\sigdn{} & 67.4\sigdn{} & 77.0 & 99k & 76.3\sigup{} & 79.4\sigup{} & 87.2 \\
PaW (coefficient 1.0) & 52k & 68.6\sigdn{} & 70.6\sigdn{} & 81.8 & 89k & 85.4 & 87.3 & 92.7 \\
Majority-vote labels & 65k & 22.6\sigup{} & 24.1\sigup{} & 28.8 & 132k & 17.2\sigup{} & 17.6\sigup{} & 20.8 \\
OCSD (self teacher) (3) & 54k & 60.6\sigdn{} & 64.0\sigdn{} & 81.0 & 96k & 83.0\sigup{} & 82.7\sigup{} & 93.6 \\
OCSD (32B teacher) & 56k & 66.1\sigdn{} & 67.2\sigdn{} & 82.1 & 99k & 83.2 & 70.1\sigup{} & 88.7 \\
OPD (32B teacher) & 58k & 43.4\sigup{} & 51.6\sigup{} & 64.6 & 112k & 41.2\sigup{} & 51.0\sigup{} & 64.2 \\
\wmppo{}, second run of seed 2$^{s}$ & 59k & 55.8 & 61.1 & 74.8 & 102k & 85.8 & 86.7 & 92.0 \\
\bottomrule
\end{tabular}

\vspace{10pt}
\centering\footnotesize
\caption{ALFWorld, the methods trained to iteration 55 (continued from Table~\ref{tab:full-alf}; same marks).}
\label{tab:full-alf-b}
\setlength{\tabcolsep}{2pt}
\begin{tabular}{l r c c c }
\toprule
 & \multicolumn{4}{c}{iteration 55} \\
\cmidrule(lr){2-5}
Method & Steps & Greedy & pass@1 & pass@3 \\
\midrule
\wmppo{} (ours) (3) & 153k & 88.7 & 90.8 & 95.4 \\
\wmppo{} (policy channel only) (3) & 147k & 91.4\sigdn{} & 92.9\sigdn{} & 96.5 \\
PPO-160 (3) & 161k & 85.0\sigup{} & 87.3\sigup{} & 93.4 \\
PPO-256 (3) & 226k & 90.8\sigdn{} & 91.1 & 95.4 \\
Wide-Tree-PPO (3) & 165k & 87.3 & 90.6 & 96.1 \\
Fixed WM & 153k & 93.4\sigdn{} & 92.2 & 96.0 \\
PaW (coefficient 1.0) & 136k & 93.1\sigdn{} & 91.8 & 95.6 \\
\wmppo{}, second run of seed 2$^{s}$ & 152k & 90.1 & 91.2 & 93.8 \\
\bottomrule
\end{tabular}

\end{table}

\begin{table}[!ht]
\centering\footnotesize
\caption{ScienceWorld, all methods, iterations 24 and 36 (policy-only success in \%; iterations 48 and 60 in Table~\ref{tab:full-sw-b}). $^{s}$Second runs of an existing seed, launched later with the same configuration; every mean and test in the paper uses the first run of each seed, and these rows are listed for completeness, without tests. $^\dagger$/$^\ddagger$: significantly lower/higher than \wmppo{} at that iteration (per-game paired test, $p<0.05$, uncorrected).}
\label{tab:full-sw}
\setlength{\tabcolsep}{2pt}
\begin{tabular}{l r c c c r c c c }
\toprule
 & \multicolumn{4}{c}{iteration 24} & \multicolumn{4}{c}{iteration 36} \\
\cmidrule(lr){2-5}\cmidrule(lr){6-9}
Method & Steps & Greedy & pass@1 & pass@3 & Steps & Greedy & pass@1 & pass@3 \\
\midrule
\wmppo{} (ours) (3) & 34k & 28.5 & 27.7 & 35.9 & 67k & 38.0 & 38.0 & 45.8 \\
\wmppo{} (policy channel only) (3) & 32k & 26.6 & 27.2 & 35.2 & 67k & 33.0\sigup{} & 32.9\sigup{} & 42.3 \\
\wmppo{}, terminal-only value target & 31k & 29.9 & 31.0\sigdn{} & 37.7 & 64k & 34.5\sigup{} & 38.0 & 45.8 \\
PPO-160 (3) & 25k & 26.5 & 26.2\sigup{} & 33.2 & 54k & 32.0\sigup{} & 31.5\sigup{} & 39.1 \\
PPO-256 (3) & 38k & 26.8 & 26.5 & 34.7 & 75k & 33.7\sigup{} & 32.3\sigup{} & 40.1 \\
GRPO & 23k & 20.8\sigup{} & 22.1\sigup{} & 29.6 & 50k & 24.3\sigup{} & 24.4\sigup{} & 33.8 \\
GRPO, 40 tasks $\times$ 4 & 23k & 24.3\sigup{} & 24.6\sigup{} & 32.0 & 46k & 28.5\sigup{} & 25.4\sigup{} & 33.5 \\
GiGPO & 26k & 25.7 & 23.9\sigup{} & 31.0 & 53k & 27.8\sigup{} & 25.9\sigup{} & 33.5 \\
GiGPO, 40 tasks $\times$ 4 & 25k & 22.2\sigup{} & 23.0\sigup{} & 29.9 & 51k & 27.8\sigup{} & 27.5\sigup{} & 35.6 \\
\sopo{} (tuned clip) & 26k & 26.1 & 23.0\sigup{} & 28.9 & 53k & 39.8 & 33.5\sigup{} & 47.2 \\
\sopo{} (original clip) & 34k & 19.4\sigup{} & 27.9 & 38.0 & 75k & 33.5 & 31.8\sigup{} & 41.9 \\
\steppo{} & 26k & 19.7\sigup{} & 19.8\sigup{} & 25.0 & 68k & 25.0\sigup{} & 23.5\sigup{} & 33.5 \\
\qppo{} (3) & 31k & 23.8\sigup{} & 24.4\sigup{} & 31.2 & 65k & 33.9\sigup{} & 33.8\sigup{} & 41.4 \\
Wide-Tree-PPO (3) & 32k & 27.1 & 26.4 & 33.8 & 65k & 32.7\sigup{} & 31.2\sigup{} & 38.4 \\
Fixed WM & 30k & 22.9\sigup{} & 21.4\sigup{} & 25.4 & 57k & 31.7\sigup{} & 29.8\sigup{} & 34.9 \\
PaW (coefficient 0.5) & 27k & 21.5\sigup{} & 22.7\sigup{} & 28.5 & 53k & 26.8\sigup{} & 25.4\sigup{} & 35.6 \\
PaW (coefficient 1.0) & 30k & 18.7\sigup{} & 18.2\sigup{} & 26.1 & 67k & 10.9\sigup{} & 23.8\sigup{} & 34.2 \\
PaW (coefficient 0.25) & 24k & 20.8\sigup{} & 21.0\sigup{} & 25.7 & 44k & 26.1\sigup{} & 26.4\sigup{} & 35.2 \\
Majority-vote labels & 25k & 23.2\sigup{} & 23.9\sigup{} & 29.9 & 57k & 26.8\sigup{} & 26.3\sigup{} & 30.3 \\
OCSD (self teacher) & 26k & 29.2 & 25.7 & 32.7 & 53k & 29.2\sigup{} & 26.9\sigup{} & 34.9 \\
OCSD (32B teacher) & 25k & 25.7 & 25.4\sigup{} & 32.0 & 48k & 27.1\sigup{} & 27.0\sigup{} & 34.9 \\
OPD (32B teacher) & 27k & 27.5 & 27.6 & 39.4 & 54k & 32.4\sigup{} & 34.0\sigup{} & 47.2 \\
PPO-256, second run of seed 1$^{s}$ & 40k & 25.4 & 24.1 & 33.1 & 77k & 35.6 & 34.6 & 40.5 \\
\wmppo{} (policy channel only), second run of seed 3$^{s}$ & 32k & 29.2 & 26.9 & 34.5 & 65k & 39.4 & 38.1 & 46.1 \\
\bottomrule
\end{tabular}

\vspace{10pt}
\centering\footnotesize
\caption{ScienceWorld, all methods, iterations 48 and 60 (continued from Table~\ref{tab:full-sw}; same marks).}
\label{tab:full-sw-b}
\setlength{\tabcolsep}{2pt}
\begin{tabular}{l r c c c r c c c }
\toprule
 & \multicolumn{4}{c}{iteration 48} & \multicolumn{4}{c}{iteration 60} \\
\cmidrule(lr){2-5}\cmidrule(lr){6-9}
Method & Steps & Greedy & pass@1 & pass@3 & Steps & Greedy & pass@1 & pass@3 \\
\midrule
\wmppo{} (ours) (3) & 99k & 44.4 & 43.2 & 51.5 & 128k & 52.9 & 51.6 & 59.6 \\
\wmppo{} (policy channel only) (3) & 99k & 41.1\sigup{} & 41.2\sigup{} & 50.1 & 130k & 47.2\sigup{} & 47.2\sigup{} & 56.0 \\
\wmppo{}, terminal-only value target & 93k & 43.3 & 42.4 & 50.0 & 125k & 48.2\sigup{} & 49.3 & 59.9 \\
PPO-160 (3) & 84k & 36.3\sigup{} & 37.1\sigup{} & 45.9 & 113k & 39.3\sigup{} & 42.6\sigup{} & 52.2 \\
PPO-256 (3) & 114k & 38.7\sigup{} & 38.3\sigup{} & 47.4 & 156k & 43.8\sigup{} & 43.5\sigup{} & 51.9 \\
GRPO & 79k & 31.0\sigup{} & 28.1\sigup{} & 38.4 & 103k & 35.2\sigup{} & 35.7\sigup{} & 43.3 \\
GRPO, 40 tasks $\times$ 4 & 69k & 29.9\sigup{} & 29.1\sigup{} & 35.9 & 88k & 31.3\sigup{} & 31.5\sigup{} & 39.8 \\
GiGPO & 79k & 28.9\sigup{} & 28.3\sigup{} & 35.9 & 105k & 38.4\sigup{} & 36.3\sigup{} & 43.7 \\
GiGPO, 40 tasks $\times$ 4 & 78k & 29.6\sigup{} & 31.9\sigup{} & 39.1 & 100k & 34.2\sigup{} & 33.0\sigup{} & 41.2 \\
\sopo{} (tuned clip) & 83k & 40.5 & 35.7\sigup{} & 48.6 & 111k & 46.5\sigup{} & 49.4 & 60.9 \\
\sopo{} (original clip) & 128k & 16.9\sigup{} & 16.4\sigup{} & 17.6 & 164k & 15.8\sigup{} & 16.2\sigup{} & 16.9 \\
\steppo{} & 104k & 25.4\sigup{} & 23.8\sigup{} & 29.9 & 138k & 25.0\sigup{} & 26.5\sigup{} & 38.4 \\
\qppo{} (3) & 99k & 40.6\sigup{} & 38.9\sigup{} & 46.7 & 133k & 46.1\sigup{} & 46.2\sigup{} & 54.8 \\
Wide-Tree-PPO (3) & 95k & 36.0\sigup{} & 35.8\sigup{} & 43.2 & 125k & 40.8\sigup{} & 40.0\sigup{} & 47.8 \\
Fixed WM & 82k & 35.2\sigup{} & 34.9\sigup{} & 40.1 & 108k & 38.7\sigup{} & 39.0\sigup{} & 47.5 \\
PaW (coefficient 0.5) & 80k & 29.2\sigup{} & 28.9\sigup{} & 38.7 & 109k & 35.6\sigup{} & 38.8\sigup{} & 48.9 \\
PaW (coefficient 1.0) & 97k & 6.0\sigup{} & 29.3\sigup{} & 39.1 & 131k & 6.0\sigup{} & 30.8\sigup{} & 37.0 \\
PaW (coefficient 0.25) & 65k & 29.2\sigup{} & 29.5\sigup{} & 37.7 & 87k & 32.4\sigup{} & 32.9\sigup{} & 39.4 \\
Majority-vote labels & 94k & 24.6\sigup{} & 24.6\sigup{} & 29.6 & 135k & 25.0\sigup{} & 25.0\sigup{} & 27.8 \\
OCSD (self teacher) & 81k & 33.5\sigup{} & 36.6\sigup{} & 43.0 & 112k & 34.2\sigup{} & 43.7\sigup{} & 53.2 \\
OCSD (32B teacher) & 70k & 33.8\sigup{} & 32.7\sigup{} & 41.9 & 90k & 43.0\sigup{} & 41.5\sigup{} & 49.6 \\
OPD (32B teacher) & 80k & 28.2\sigup{} & 31.3\sigup{} & 44.4 & 106k & 36.3\sigup{} & 33.9\sigup{} & 46.8 \\
PPO-256, second run of seed 1$^{s}$ & 116k & 39.4 & 39.4 & 45.4 & 160k & 40.8 & 44.6 & 55.3 \\
\wmppo{} (policy channel only), second run of seed 3$^{s}$ & 95k & 43.7 & 43.5 & 50.7 & 122k & 44.4 & 45.7 & 53.2 \\
\bottomrule
\end{tabular}

\end{table}

\FloatBarrier
\subsection{Learning curves and steps to a reference success}
\label{app:curves}

\begin{figure}[!ht]
\centering
\includegraphics[width=\linewidth]{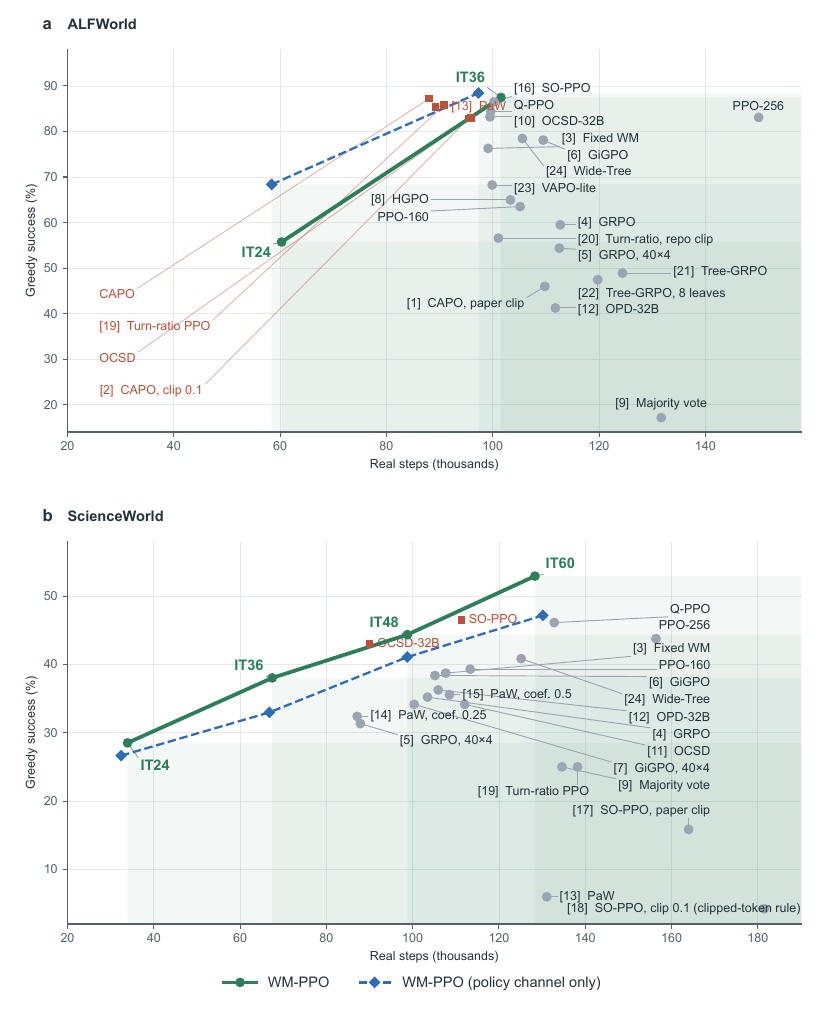}
\caption{\textbf{Success against real interaction, with every method named.} The data and conventions of Figure~\ref{fig:frontier}; the bracketed numbers there refer to the names given here.}
\label{fig:frontier-full}
\end{figure}

\begin{figure}[!htp]
\centering
\includegraphics[width=\linewidth]{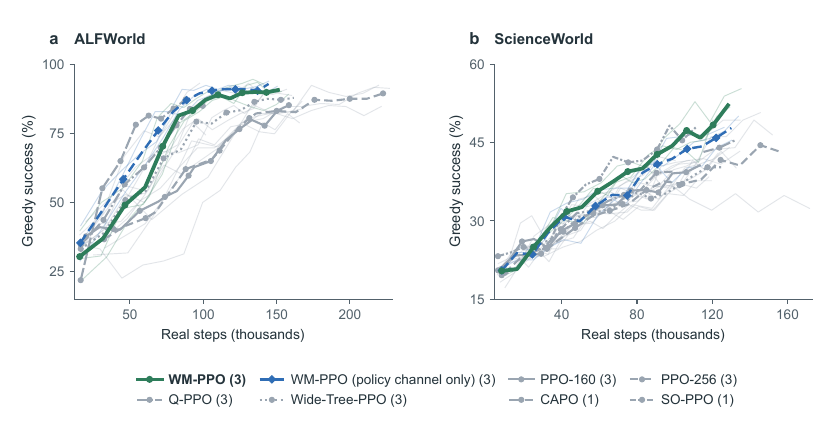}
\caption{\textbf{Learning curves against real interaction.} Greedy success on all evaluation games during training (every third iteration; one greedy episode per game) against the real steps executed since the common warm-up checkpoint (warm-up and prefix replays for branches are not counted). Thin gray lines: individual training runs; colored lines: their mean, with the number of runs in parentheses.}
\label{fig:curves}

\vspace{4pt}
\centering\small
\captionof{table}{\textbf{Real steps to reach a reference success.} Real steps (thousands, from iteration 13) at which each training run first reached the given greedy success in the evaluations during training (every third iteration, one greedy episode per game, so the resolution is coarse); one entry per run, methods as in Table~\ref{tab:main}; $>$: not reached by the run's last evaluation, after the given number of real steps. On ALFWorld both \wmppo{} variants, PPO-160, PPO-256, Wide-Tree-PPO, Fixed WM and PaW were trained to iteration 55 and the other methods to iteration 36, so their last evaluations come at different step counts.}
\label{tab:reach}
{\setlength{\tabcolsep}{4pt}\begin{tabular}{@{}l c c c c@{}}
\toprule
 & \multicolumn{2}{c}{ALFWorld} & \multicolumn{2}{c}{ScienceWorld} \\
\cmidrule(lr){2-3}\cmidrule(lr){4-5}
Method & 80\% & 85\% & 40\% & 45\% \\
\midrule
\textbf{\wmppo{} (ours)} & \textbf{69, 94, 97} & \textbf{78, 103, 106} & \textbf{69, 86, 93} & \textbf{94, 109, 120} \\
\wmppo{} (policy channel only) & 84, 67, 78 & 93, 86, 96 & 89, 86, 90 & 96, 110, 130 \\
\midrule
\multicolumn{5}{@{}l}{\emph{Standard training and refinements}} \\
PPO-160 & 106, 157, 158 & 121, $>$165, 167 & 85, $>$118, 93 & $>$114, $>$118, $>$108 \\
PPO-256 & 129, 141, 133 & 185, 180, 133 & $>$172, 82, 111 & $>$172, 142, 119 \\
GRPO & $>$113 & $>$113 & $>$103 & $>$103 \\
GiGPO & $>$109 & $>$109 & $>$105 & $>$105 \\
CAPO & 63 & 88 & -- & -- \\
\sopo{} & 100 & 100 & 67 & 97 \\
\steppo{} & 64 & 73 & -- & -- \\
\midrule
\multicolumn{5}{@{}l}{\emph{Structural controls}} \\
\qppo{} & 81, 81, $>$102 & 100, 89, $>$102 & 98, 92, 105 & $>$135, 118, $>$129 \\
Wide-Tree-PPO & 96, 93, 139 & 143, 103, $>$168 & $>$120, 110, 125 & $>$120, $>$123, $>$133 \\
Fixed WM & 80 & 108 & $>$108 & $>$108 \\
\midrule
\multicolumn{5}{@{}l}{\emph{Auxiliary feedback added to PPO-160}} \\
Majority-vote labels & $>$132 & $>$132 & $>$135 & $>$135 \\
OCSD (self teacher) & 87, 81, 101 & $>$97, $>$90, $>$101 & $>$112 & $>$112 \\
OCSD (32B teacher) & 89 & $>$99 & 90 & $>$90 \\
OPD (32B teacher) & $>$112 & $>$112 & $>$106 & $>$106 \\
PaW & 89 & 97 & $>$109 & $>$109 \\
\bottomrule
\end{tabular}
}
\end{figure}

Figure~\ref{fig:frontier-full} names every method of Figure~\ref{fig:frontier}. Figure~\ref{fig:curves} gives the learning curves against real interaction, and Table~\ref{tab:reach} gives the real steps at which each run first passed reference success levels.

\FloatBarrier
\subsection{Individual training runs}
\label{app:seeds}

Table~\ref{tab:seeds} lists every run of the methods with more than one run.

\begin{table}[!ht]
\centering\small
\caption{Results of individual training runs for the methods with more than one completed run (policy-only success in \%, greedy / pass@1). Each seed is represented by its first run. Same-seed reruns, in no mean or test: ALFWorld \wmppo{} seed 2, 85.8 / 86.7 at iteration 36 (first run 87.6 / 86.9); ScienceWorld PPO-256 seed 1, 40.8 / 44.6, and \wmppo{} (policy channel only) seed 3, 44.4 / 45.7 (first run 44.0 / 45.5).}
\label{tab:seeds}
\begin{tabular}{l l l c c c}
\toprule
Environment & It. & Method & Run 1 & Run 2 & Run 3 \\
\midrule
ALFWorld & 36 & \wmppo{} & 88.3 / 89.7 & 87.6 / 86.9 & 86.5 / 86.7 \\
 &  & \wmppo{} (policy channel only) & 89.4 / 90.8 & 85.8 / 87.7 & 90.1 / 89.8 \\
 &  & PPO-160 & 70.1 / 77.4 & 52.2 / 65.5 & 68.2 / 70.3 \\
 &  & PPO-256 & 83.2 / 86.1 & 86.1 / 83.8 & 79.9 / 84.4 \\
 &  & \qppo{} & 85.8 / 86.1 & 92.7 / 92.3 & 74.8 / 87.2 \\
 &  & Wide-Tree-PPO & 76.6 / 83.8 & 88.7 / 86.6 & 70.1 / 76.5 \\
 &  & OCSD (self teacher) & 84.7 / 83.7 & 81.0 / 85.2 & 83.2 / 79.2 \\
\midrule
ALFWorld & 55 & \wmppo{} & 91.2 / 92.3 & 88.3 / 89.7 & 86.5 / 90.5 \\
 &  & \wmppo{} (policy channel only) & 91.2 / 92.5 & 90.5 / 93.8 & 92.3 / 92.6 \\
 &  & PPO-160 & 89.8 / 89.8 & 80.3 / 86.6 & 85.0 / 85.4 \\
 &  & PPO-256 & 86.1 / 91.0 & 93.8 / 90.3 & 92.3 / 92.1 \\
 &  & Wide-Tree-PPO & 86.1 / 89.2 & 92.0 / 93.6 & 83.9 / 89.2 \\
\midrule
ScienceWorld & 60 & \wmppo{} & 54.9 / 56.5 & 53.5 / 51.3 & 50.4 / 46.9 \\
 &  & \wmppo{} (policy channel only) & 48.6 / 48.9 & 48.9 / 47.1 & 44.0 / 45.5 \\
 &  & PPO-160 & 40.5 / 43.4 & 37.7 / 39.8 & 39.8 / 44.5 \\
 &  & PPO-256 & 33.8 / 36.0 & 49.3 / 46.4 & 48.2 / 48.2 \\
 &  & \qppo{} & 45.1 / 44.6 & 51.8 / 49.2 & 41.5 / 45.0 \\
 &  & Wide-Tree-PPO & 37.3 / 36.3 & 45.1 / 42.4 & 40.1 / 41.4 \\
\bottomrule
\end{tabular}

\vspace{8pt}
\centering\footnotesize
\caption{ScienceWorld at iteration 60 by task type (policy-only success in \%, greedy / pass@1; methods averaged over their completed runs as in Table~\ref{tab:main}). Shortened task names: melting-point (known) = measure-melting-point-known-substance; paint-secondary, paint-tertiary = chemistry-mix-paint-secondary-color, -tertiary-color; power-(non)renewable = power-component-renewable-vs-nonrenewable-energy; conductivity (unknown) = test-conductivity-of-unknown-substances; lifespan-longest-then-shortest = lifespan-longest-lived-then-shortest-lived.}
\label{tab:swtasks}
\setlength{\tabcolsep}{3pt}
\begin{tabular}{l r c c c c c }
\toprule
Task type & Games & \wmppo{} & Policy ch.\ only & PPO-160 & \qppo{} & \sopo{} (tuned) \\
\midrule
\emph{Procedural (all)} & 188 & 46.3/45.1 & 39.5/38.6 & 29.6/34.4 & 39.0/39.4 & 39.4/42.2 \\
\quad use-thermometer & 18 & 0.0/0.0 & 0.0/0.0 & 0.0/0.0 & 0.0/0.0 & 0.0/0.0 \\
\quad melting-point (known) & 18 & 66.7/58.0 & 53.7/48.1 & 37.0/46.9 & 38.9/47.5 & 50.0/44.4 \\
\quad power-component & 10 & 73.3/76.7 & 70.0/70.0 & 43.3/48.9 & 43.3/36.7 & 60.0/76.7 \\
\quad test-conductivity & 18 & 25.9/25.3 & 35.2/34.6 & 22.2/21.0 & 20.4/24.7 & 27.8/31.5 \\
\quad find-living-thing & 18 & 70.4/77.8 & 40.7/34.6 & 31.5/32.1 & 53.7/45.7 & 100.0/88.9 \\
\quad find-non-living-thing & 18 & 83.3/75.9 & 66.7/71.0 & 63.0/66.7 & 85.2/83.3 & 33.3/44.4 \\
\quad find-plant & 18 & 90.7/97.5 & 96.3/98.8 & 88.9/95.7 & 92.6/99.4 & 77.8/88.9 \\
\quad find-animal & 18 & 75.9/67.3 & 51.9/48.8 & 25.9/51.2 & 64.8/67.3 & 77.8/88.9 \\
\quad chemistry-mix & 16 & 0.0/0.0 & 0.0/0.7 & 0.0/0.0 & 0.0/1.4 & 6.2/0.0 \\
\quad paint-secondary & 18 & 29.6/26.5 & 29.6/27.8 & 16.7/18.5 & 27.8/22.2 & 5.6/11.1 \\
\quad paint-tertiary & 18 & 0.0/0.0 & 0.0/0.0 & 0.0/0.0 & 0.0/0.0 & 0.0/0.0 \\
\midrule
\emph{Forced-choice (all)} & 96 & 66.0/64.2 & 62.2/63.9 & 58.3/58.6 & 60.1/59.6 & 60.4/63.5 \\
\quad power-(non)renewable & 10 & 30.0/24.4 & 16.7/25.6 & 13.3/11.1 & 16.7/6.7 & 40.0/36.7 \\
\quad conductivity (unknown) & 18 & 55.6/51.2 & 31.5/45.1 & 35.2/40.7 & 27.8/35.2 & 55.6/51.9 \\
\quad lifespan-longest-lived & 18 & 100.0/100.0 & 100.0/100.0 & 100.0/100.0 & 100.0/100.0 & 94.4/96.3 \\
\quad lifespan-shortest-lived & 18 & 94.4/90.1 & 94.4/93.5 & 92.6/92.6 & 96.3/93.2 & 83.3/88.9 \\
\quad lifespan-longest-then-shortest & 18 & 79.6/80.9 & 90.7/84.6 & 72.2/70.4 & 83.3/79.6 & 66.7/70.4 \\
\quad identify-life-stages-1 & 8 & 0.0/0.0 & 0.0/0.0 & 0.0/0.0 & 0.0/0.0 & 0.0/0.0 \\
\quad identify-life-stages-2 & 6 & 16.7/20.4 & 16.7/11.1 & 11.1/7.4 & 11.1/18.5 & 0.0/33.3 \\
\bottomrule
\end{tabular}

\end{table}

\subsection{ScienceWorld by task type}
\label{app:swtasks}Table~\ref{tab:swtasks} breaks down the iteration-60 results of Table~\ref{tab:main} by task type. With 6--18 games per type, differences on single types are noisy, so Section~\ref{sec:behavior} uses the procedural and forced-choice totals. Three types are solved by none of these methods, and the chemistry-mix tasks are almost never solved.

\FloatBarrier
\subsection{Interaction and computation}
\label{app:cost}

Table~\ref{tab:cost} separates the costs per iteration. \wmppo{} executes 3\% fewer real steps per iteration than PPO-160 on ALFWorld and 13\% more on ScienceWorld, where its branches run longer than the budget assumes. It generates about twice as many policy responses (the four candidates per chain step) and adds 1,500--1,950 world-model generations. An iteration takes 40 minutes on ALFWorld against 32 for PPO-160, and 55 minutes against 34 on ScienceWorld. Because \wmppo{} also occupies the world-model servers and trainers, its GPU-hours per iteration are about 2.7 and 3.8 times those of PPO-160; these are upper bounds, since shared servers are counted in full. The value channel adds no generation, only a term of the critic's loss, and its GPU-hours per iteration are similar to those of the policy-channel-only variant. Branching, which Wide-Tree-PPO shares, adds little (1.5 against 1.7 GPU-hours on ALFWorld, 1.9 against 1.7 on ScienceWorld); most of the difference comes from sampling the candidates and from the world model's predictions and training.

\begin{table}[!ht]
\centering\small
\caption{\textbf{Interaction and computation per iteration} (iterations 13--36 for ALFWorld and 13--60 for ScienceWorld). Real steps: steps the policy generated and executed, averaged over all runs as in Table~\ref{tab:main}. Policy samples: generated responses (four per chain step for \wmppo{} and \qppo{}; resampling at branch points is not logged, so these are lower bounds for branched methods). WM generations: one per distinct candidate at an evaluated decision point and two per branch pair (\qppo{} values the same candidates with its action-value model and generates no predictions). Wall: measured wall-clock time per iteration, including the in-training evaluation; GPU$\cdot$h: wall-clock time times all GPUs allocated to the run (7 for \wmppo{}, 3--5 for the other methods; shared servers counted in full, an upper bound on use). Both are means over the iterations of each method's three runs on A100-40GB nodes; two early segments on H100 nodes are excluded.}
\label{tab:cost}
\resizebox{\linewidth}{!}{\begin{tabular}{l l r r r r r}
\toprule
Environment & Method & Real steps & Policy samples & WM generations & Wall (min) & GPU$\cdot$h \\
\midrule
ALFWorld & \wmppo{} & 4,232 & 8,914 & 1,930 & 40 & 4.6 \\
 & PPO-160 & 4,381 & 4,381 & 0 & 32 & 1.7 \\
 & PPO-256 & 6,252 & 6,252 & 0 & 28 & 2.3 \\
 & Wide-Tree-PPO & 4,399 & 4,399 & 0 & 31 & 1.5 \\
 & \qppo{} & 4,152 & 8,795 & -- & 46 & 2.7 \\
\midrule
ScienceWorld & \wmppo{} & 2,674 & 5,205 & 1,483 & 55 & 6.4 \\
 & PPO-160 & 2,361 & 2,361 & 0 & 34 & 1.7 \\
 & PPO-256 & 3,258 & 3,258 & 0 & 32 & 2.4 \\
 & Wide-Tree-PPO & 2,607 & 2,607 & 0 & 37 & 1.9 \\
 & \qppo{} & 2,767 & 5,443 & -- & 70 & 3.5 \\
\bottomrule
\end{tabular}
}
\end{table}

\subsection{Sokoban comparison with ProAct's MC-PPO}
\label{app:sokoban}

We compare \wmppo{} with the MC-PPO configuration that ProAct releases for Sokoban~\citep{proact}. The environment is text Sokoban on a $6\times6$ grid with one box and at most 15 steps; solving a level gives a terminal reward of 1 and any other ending 0. Training draws from 18,908 levels, and evaluation uses 300 held-out levels with greedy decoding and three temperature-1 samples per level. All three methods start from a common 12-iteration PPO warm-up (52.0\% greedy success) and train through iteration 36 with the budgets and hyperparameters of the main experiments, except a response limit of 1,024 tokens (also in the warm-up), because the prompt asks for step-by-step reasoning: PPO-160 plays 160 levels per iteration, and \wmppo{} plays 64 levels with one chain and three branches, both channels and the gate as in Section~\ref{sec:method}; its initial world model is trained on the 7,962 executed transitions of the warm-up. MC-PPO uses PPO-160's budget and level schedule and estimates the value of each visited state from 1,000 uniformly random rollouts of at most 15 steps in a copy of the simulator, with TD(0) advantages, discount 0.95 and a Monte-Carlo value weight of 1. ProAct's GLAD stage, which fine-tunes on reasoning derived from environment search, is not part of this comparison. Over iterations 13--36, MC-PPO performs about 33 million random rollouts, which are not counted as real steps. ProAct's own Sokoban experiments use one to five boxes and shaped rewards. Each method has one run.

\begin{table}[!ht]
\centering\small
\caption{\textbf{Sokoban} (300 held-out levels; success in \%; one run per method). Steps: real environment steps from iteration 13. Marks as in Table~\ref{tab:main}, against \wmppo{}.}
\label{tab:sokoban}
\begin{tabular}{l r c c r c c}
\toprule
 & \multicolumn{3}{c}{iteration 24} & \multicolumn{3}{c}{iteration 36} \\
\cmidrule(lr){2-4}\cmidrule(lr){5-7}
Method & Steps & Greedy & pass@1 & Steps & Greedy & pass@1 \\
\midrule
\wmppo{} (ours) & 20k & 80.3 & 69.8 & 36k & 87.0 & 84.4 \\
ProAct (MC-PPO) & 18k & 77.3 & 70.9 & 33k & 77.3\sigup{} & 68.1\sigup{} \\
PPO-160 & 19k & 62.0\sigup{} & 53.4\sigup{} & 36k & 67.7\sigup{} & 63.2\sigup{} \\
\bottomrule
\end{tabular}

\end{table}

\section{Analyses behind Section~\ref{sec:understanding}}
\label{app:analyses}

The following analyses support Section~\ref{sec:understanding}. Real branch outcomes test the reliability of transition values (Section~\ref{sec:valuing}); measurements of prediction quality, advantages and critic values examine how transition values enter learning (Section~\ref{sec:learning}); and task trajectories show how the trained policy acts (Section~\ref{sec:behavior}).

\FloatBarrier
\subsection{Branch pairs and the reliability gate}
\label{app:branch}

\paragraph{Protocol.} In each iteration, every branch pair (the chain action and the branch action at the branch point, both played to the end) was valued by the iteration's pre-update models and assigned to a margin bin with that iteration's thresholds. A pair therefore never contributes to the weights that were applied to it. Pairs whose two returns differ and whose value difference is non-zero are \emph{decisive}, and a decisive pair is correct when the preferred action has the higher return. \wmppo{}'s transition value includes the ending rule: a predicted completion or failure is assigned 1 or 0, and any other successor is valued by the critic. \qppo{}'s value is $Q_\omega(s_t,y)$. Each method's pairs come from its own sampling, so the two methods are not evaluated on the same pairs. In ScienceWorld, a pair \emph{involves focus} if either action starts with \texttt{focus on}; this is a property of the action text, not a label of whether the outcome is decided at that step. Intervals: 2,000 bootstrap resamples of (run, iteration) clusters. Data: three runs of each method, iterations 13--60 on ScienceWorld and 13--36 (both \wmppo{} variants: to 55) on ALFWorld. Table~\ref{tab:branch} gives the agreement and the gain by phase and margin bin, and Table~\ref{tab:branch-split} splits the late ScienceWorld pairs. For the single Fixed WM run on ScienceWorld, agreement in the largest bin is 0.67, 0.75, 0.53 and 0.64 in the four phases, against 0.72, 0.79, 0.88 and 0.93 for the run of \wmppo{} (policy channel only) with the same seed and the same initial world model.

\paragraph{Gain and agreement.} The gain in Equation~\eqref{eq:gain} averages over all branch pairs in a margin bin. Among untied pairs in bin $\ell$, let $d_\ell$ be the share whose outcomes differ and $c_\ell$ the share of these in which the preferred action has the better outcome. Then
\begin{equation}
g_\ell=\Pr\bigl(\Delta\qhat\neq0\mid\ell\bigr)\,d_\ell\,\bigl(2c_\ell-1\bigr).
\label{eq:gain-factor}
\end{equation}
When the outcomes differ, a correct preference contributes $+1$ and an incorrect preference contributes $-1$; equal outcomes or tied transition values contribute zero. The gate converts the gain and the number of supporting pairs into a training weight through Equation~\eqref{eq:gate}.

\paragraph{Early comparisons with the value channel.} Among ScienceWorld branch pairs with different real outcomes, the largest-margin comparisons favor the successful action in 78--82\% of cases with the value channel early in training, against 71--74\% with the policy channel only (Figure~\ref{fig:value-channel}b). Later, the two variants reach similar ranking accuracy.

\begin{table}[!ht]
\centering\scriptsize
\caption{\textbf{Agreement of training-time preferences with real outcomes}, by phase and margin bin (decisive pairs only). Gain per pair: Equation~\eqref{eq:gain} averaged over all pairs of the bin, decisive or not (small+middle: mean of the two bins).}
\label{tab:branch}
\setlength{\tabcolsep}{3pt}\footnotesize
\begin{tabular}{l l l r c c c c c c}
\toprule
 & & & & & \multicolumn{3}{c}{Agreement by margin bin} & \multicolumn{2}{c}{Gain per pair} \\
\cmidrule(lr){6-8}\cmidrule(lr){9-10}
Environment & Valued by & Iterations & Pairs & Agreement & small & middle & large & small+middle & large \\
\midrule
ScienceWorld & \wmppo{} & 13--24 & 415 & 0.71 & 0.61 & 0.55 & 0.78 & +0.008 & +0.077 \\
 &  & 25--36 & 549 & 0.73 & 0.48 & 0.58 & 0.82 & +0.007 & +0.145 \\
 &  & 37--48 & 626 & 0.77 & 0.49 & 0.58 & 0.88 & +0.008 & +0.199 \\
 &  & 49--60 & 654 & 0.76 & 0.55 & 0.58 & 0.84 & +0.010 & +0.222 \\
 & Policy channel only & 13--24 & 397 & 0.67 & 0.49 & 0.65 & 0.71 & +0.009 & +0.060 \\
 &  & 25--36 & 543 & 0.65 & 0.49 & 0.55 & 0.74 & +0.005 & +0.092 \\
 &  & 37--48 & 527 & 0.73 & 0.41 & 0.57 & 0.84 & +0.003 & +0.164 \\
 &  & 49--60 & 579 & 0.73 & 0.54 & 0.45 & 0.87 & -0.003 & +0.200 \\
 & \qppo{} (learned $Q$) & 13--24 & 514 & 0.59 & 0.53 & 0.52 & 0.65 & +0.003 & +0.053 \\
 &  & 25--36 & 579 & 0.70 & 0.62 & 0.58 & 0.75 & +0.012 & +0.117 \\
 &  & 37--48 & 679 & 0.67 & 0.67 & 0.54 & 0.71 & +0.012 & +0.114 \\
 &  & 49--60 & 775 & 0.70 & 0.63 & 0.52 & 0.78 & +0.010 & +0.171 \\
\midrule
ALFWorld & \wmppo{} & 13--24 & 424 & 0.60 & 0.51 & 0.61 & 0.66 & +0.015 & +0.051 \\
 &  & 25--36 & 424 & 0.55 & 0.49 & 0.52 & 0.64 & +0.002 & +0.047 \\
 &  & 37--55 & 398 & 0.53 & 0.56 & 0.44 & 0.60 & -0.002 & +0.026 \\
 & Policy channel only & 13--24 & 525 & 0.55 & 0.52 & 0.48 & 0.64 & -0.000 & +0.050 \\
 &  & 25--36 & 334 & 0.58 & 0.50 & 0.59 & 0.68 & +0.009 & +0.051 \\
 &  & 37--55 & 361 & 0.58 & 0.58 & 0.50 & 0.65 & +0.004 & +0.037 \\
 & \qppo{} (learned $Q$) & 13--24 & 501 & 0.54 & 0.53 & 0.52 & 0.56 & +0.007 & +0.020 \\
 &  & 25--36 & 390 & 0.56 & 0.54 & 0.56 & 0.57 & +0.013 & +0.016 \\
\bottomrule
\end{tabular}

\end{table}

\begin{table}[!ht]
\centering\small
\caption{\textbf{ScienceWorld, iterations 45--60: which comparisons the agreement rests on} (agreement in \%, with the number of decisive pairs, i.e. pairs whose real outcomes differ; three runs of each method). Last column: \wmppo{} minus \qppo{} in percentage points, with cluster-bootstrap 95\% intervals.}
\label{tab:branch-split}
\setlength{\tabcolsep}{4pt}
\begin{tabular}{l c c c c}
\toprule
Decisive pairs & \wmppo{} & Policy ch.\ only & \qppo{} & \wmppo{} $-$ \qppo{} \\
\midrule
All & 76.2 (867) & 72.3 (755) & 69.9 (1017) & +6.3 [+1, +11] \\
Largest-margin bin & 84.4 (604) & 84.7 (483) & 76.3 (683) & +8.1 [+3, +14] \\
A \texttt{focus on} command involved & 85.5 (543) & 84.0 (451) & 77.7 (623) & +7.8 [+2, +13] \\
No \texttt{focus on} command & 60.8 (324) & 54.9 (304) & 57.6 (394) & +3.2 [$-5$, +10] \\
\midrule
One successor predicted to end (1 / 0) & 94.7 (378) & 94.5 (328) & -- & -- \\
Both continue, valued by the critic & 62.0 (489) & 55.3 (427) & -- & -- \\
\quad of these, in the largest-margin bin & 67.0 (230) & 62.2 (172) & -- & -- \\
\bottomrule
\end{tabular}

\end{table}

Table~\ref{tab:branch-split} splits the late ScienceWorld pairs by the source of the values and by the kind of action. Where one successor is predicted to end the episode, its transition value is the predicted outcome, 1 or 0, rather than a critic value (Equation~\eqref{eq:score}). These comparisons agree with the real outcome in 95\% of the pairs, with or without the value channel. Where both successors continue, the critic orders them, and the value channel raises agreement from 55\% to 62\% (from 62\% to 67\% in the largest-margin bin). The value channel thus strengthens the part of the valuation that rests on the critic. The run with a terminal-only value target, which values every candidate not predicted to end the episode at the critic's value of the current state, reaches 48.2\% greedy success at iteration 60, close to the policy-channel-only run with the same seed (48.6\%) and below the \wmppo{} run with that seed (54.9\%; Tables~\ref{tab:full-sw-b} and~\ref{tab:seeds}). Against \qppo{}, the difference lies in comparisons that involve a \texttt{focus} command, which commits the agent to an answer and whose consequence the next observation often reveals; without such a command, the two methods are close. \qppo{}'s gate gives more weight in total, a summed weight of about 263 per iteration against 183 for \wmppo{}, yet \wmppo{} is the more successful of the two (Section~\ref{sec:valuing}).

\label{app:gate}

Table~\ref{tab:gate} summarizes the gate per phase of training, averaged over iterations and over the three runs in each environment. \emph{Decision pts}: chain steps with at least two distinct candidates. \emph{Non-zero}, \emph{Full}: the number of these whose margin bin had a positive weight or a weight of 1. \emph{Mean $w_\ell$}: the weight applied to each margin bin in the iteration's update; over ScienceWorld iterations 45--60, the window of Table~\ref{tab:branch-split}, it is 0.15, 0.37 and 1.00. The 13--24 rows include iterations 13 and 14, which use the start-up rule of Table~\ref{tab:hparams}. \emph{Agreement}: fraction of the iteration's branch pairs with different outcomes in which the action with the higher value had the better outcome. \emph{Pairs}: number of such pairs per iteration. \emph{Gain}: the realized gain of Equation~\eqref{eq:gain} on the iteration's pairs, per margin bin.

\paragraph{Gate ablation.}\label{app:gate-ablation} Two copies of the first \wmppo{} run on ScienceWorld fix the three bin weights from iteration 13 onward; both channels, the world-model updates and all other settings are unchanged. Weights $(1,1,1)$ admit every comparison at full weight, and weights $(0,0,1)$ (top bin) keep only the largest-margin bin. Relative to the calibrated first run, the summed gate weight per iteration is 1.9--2.8 times as large without the gate and 0.4--0.7 times as large with the top bin (ranges over the four training phases). Applying the gate rule to each copy's own branch records shows what calibration would have admitted: in the top-bin copy, positive weight for the small and middle bins in 28 and 27 of 48 iterations; in the full-weight copy, positive weight for the small bin in only 11 of 48 iterations, where that bin's mean realized gain is $-0.005$. Table~\ref{tab:gate-ablation} gives the official evaluations.

\begin{table}[!ht]
\centering\small
\caption{\textbf{Gate ablation on ScienceWorld} (284 evaluation games; success in \%; first run of each method). $^\dagger$/$^\ddagger$: \wmppo{} / the other method is significantly higher in a per-game paired test ($p<0.05$).}
\label{tab:gate-ablation}
\setlength{\tabcolsep}{4pt}
\begin{tabular}{l l c c c c c c }
\toprule
 & & \multicolumn{2}{c}{iteration 36} & \multicolumn{2}{c}{iteration 48} & \multicolumn{2}{c}{iteration 60} \\
\cmidrule(lr){3-4}\cmidrule(lr){5-6}\cmidrule(lr){7-8}
Method & $(w_1, w_2, w_3)$ & Greedy & pass@1 & Greedy & pass@1 & Greedy & pass@1 \\
\midrule
\wmppo{} (ours) & calibrated & 40.8 & 41.7 & 45.8 & 46.5 & 54.9 & 56.5 \\
No gate & (1, 1, 1) & 35.6\sigup{} & 35.9\sigup{} & 39.1\sigup{} & 41.4\sigup{} & 44.0\sigup{} & 43.1\sigup{} \\
Top bin & (0, 0, 1) & 34.9\sigup{} & 34.0\sigup{} & 43.0 & 45.3 & 48.6\sigup{} & 48.6\sigup{} \\
PPO-160 & -- & 32.4\sigup{} & 29.8\sigup{} & 40.5\sigup{} & 40.4\sigup{} & 40.5\sigup{} & 43.4\sigup{} \\
\bottomrule
\end{tabular}

\end{table}

\paragraph{Gate coverage with the value channel.} On ScienceWorld, the value channel increases the number of evaluated states with nonzero weight early in training. In iterations 13--24 and 25--36, \wmppo{} admits 327 states per iteration in each phase, against 281 and 275 with the policy channel only, averaged over iterations and the three runs. The corresponding full-weight counts are 56 against 41 in the first phase and 98 against 75 in the second. The largest-margin bin of \wmppo{} remains at full weight from iteration 29 onward; the policy-channel-only variant's mean weight falls during iterations 28--39, reaching about 0.67 (Figure~\ref{fig:gate-weights}a,b).

\begin{table}[!ht]
\centering\scriptsize
\caption{Reliability gate of \wmppo{} during training (per iteration, averaged over each phase and over the runs).}
\label{tab:gate}
\setlength{\tabcolsep}{2.5pt}
\begin{tabular}{l l r r r c c c c c}
\toprule
Environment & Iterations & Decision pts & Non-zero & Full & Mean $w_\ell$: small / middle / large & Agreement & Pairs & \multicolumn{2}{c}{Gain: small / middle / large} \\
\midrule
ALFWorld & 13--24 & 846 & 470 & 45 & 0.17 / 0.36 / 0.42 & 0.61 & 12 & \multicolumn{2}{c}{+0.002 / +0.033 / +0.051} \\
 & 25--36 & 593 & 339 & 55 & 0.11 / 0.33 / 0.70 & 0.57 & 12 & \multicolumn{2}{c}{-0.003 / +0.005 / +0.046} \\
 & 37--48 & 448 & 201 & 5 & 0.18 / 0.03 / 0.38 & 0.56 & 7 & \multicolumn{2}{c}{+0.008 / -0.013 / +0.026} \\
 & 49--55 & 470 & 204 & 6 & 0.17 / 0.01 / 0.26 & 0.54 & 7 & \multicolumn{2}{c}{+0.001 / -0.004 / +0.029} \\
\midrule
ScienceWorld & 13--24 & 540 & 327 & 56 & 0.15 / 0.27 / 0.64 & 0.71 & 12 & \multicolumn{2}{c}{+0.009 / +0.010 / +0.078} \\
 & 25--36 & 542 & 327 & 98 & 0.11 / 0.25 / 0.98 & 0.73 & 15 & \multicolumn{2}{c}{-0.002 / +0.016 / +0.149} \\
 & 37--48 & 473 & 253 & 117 & 0.11 / 0.35 / 1.00 & 0.77 & 17 & \multicolumn{2}{c}{-0.001 / +0.016 / +0.199} \\
 & 49--60 & 471 & 302 & 97 & 0.15 / 0.32 / 1.00 & 0.75 & 18 & \multicolumn{2}{c}{+0.003 / +0.015 / +0.225} \\
\bottomrule
\end{tabular}

\end{table}

\FloatBarrier
\subsection{World-model predictions along training}
\label{app:wmq}Table~\ref{tab:wmq} gives the statistics behind Figure~\ref{fig:keepup}a,b in four phases of training. At each evaluated chain state, we compare that iteration's world model's prediction for the executed action with the observation the environment returned. \emph{Exact}: the two match after whitespace normalization. \emph{Rejections}: recall and precision of the prediction that the simulator rejects the command. \emph{Endings foreseen}: fraction of the steps that ended the episode for which the world model predicted an ending. On ScienceWorld, the online world model (the three policy-channel-only runs) is compared with the run whose world model is fixed, starting from the same initial model; each is measured on the states visited by its own policy. By phase, the online model predicts 84--86\% of the executed steps' observations exactly throughout training, whereas the fixed model falls from 81\% to 68\%. The online model also foresees 73--75\% of the steps that end an episode, against about half for the fixed model. Figure~\ref{fig:keepup} shows the same quantities in windows of three iterations.

\begin{table}[!ht]
\centering\small
\caption{World-model predictions of executed steps along training.}
\label{tab:wmq}
\resizebox{\linewidth}{!}{\begin{tabular}{l l l r c c c c}
\toprule
Environment & World model & Iterations & Steps & Exact & Rejections: recall / precision & Endings foreseen \\
\midrule
ScienceWorld & online & 13--24 & 16,602 & 0.84 & 0.99 / 0.99 & 0.74 \\
 &  & 25--36 & 18,066 & 0.86 & 0.99 / 0.99 & 0.73 \\
 &  & 37--48 & 17,036 & 0.86 & 0.99 / 0.99 & 0.74 \\
 &  & 49--60 & 16,452 & 0.85 & 0.99 / 0.99 & 0.75 \\
 & fixed & 13--24 & 5,114 & 0.81 & 0.97 / 0.98 & 0.51 \\
 &  & 25--36 & 4,543 & 0.80 & 0.96 / 0.98 & 0.52 \\
 &  & 37--48 & 3,959 & 0.74 & 0.96 / 0.97 & 0.50 \\
 &  & 49--60 & 4,308 & 0.68 & 0.92 / 0.89 & 0.49 \\
\bottomrule
\end{tabular}
}
\end{table}

\FloatBarrier
\subsection{Recomputed advantages}
\label{app:advrec}

Simulator feedback makes accepted and rejected commands a direct probe of how learning signals distinguish candidates at the same state. A rejected command usually leaves the task recoverable and rarely determines its final outcome. The following measurements compare advantages and candidate preferences; Section~\ref{app:rejected} reports the critics' comparisons during training and at evaluation, and Section~\ref{app:behavior} reports the policy's command choices during training.

To measure how the advantages separate useful from useless steps, we recomputed the raw tree-GAE advantages of ScienceWorld iterations 15, 30, 45 and 60 for the three runs each of PPO-160, \wmppo{}, \wmppo{} (policy channel only) and Wide-Tree-PPO. The recomputation uses the critic of training (saved at the end of the preceding iteration) and the training code's estimator ($\gamma=1$, $\lambda=0.95$). The means and standard deviations of the recomputed advantages match the training logs to four decimals in all 48 (run, iteration) cases, and so do the counts of auxiliary rows of both \wmppo{} variants. Executed non-terminal edges are classified by the true next observation as rejected (the simulator's \emph{No known action matches that input}) or accepted. Table~\ref{tab:adv} summarizes the results, and Table~\ref{tab:advrec} gives the counts.

\begin{figure}[t]
\centering
\includegraphics[width=0.96\linewidth]{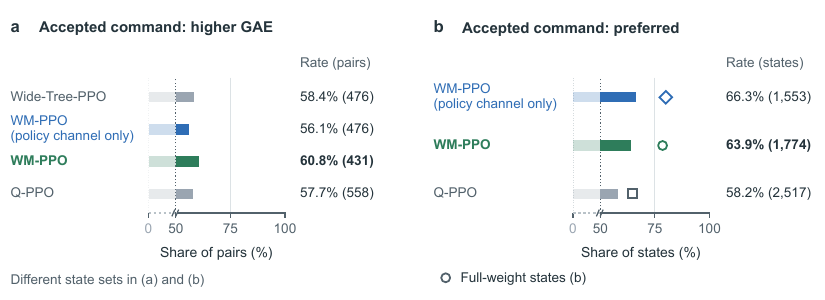}
\caption{\textbf{Which signal prefers the accepted command at the same state?} ScienceWorld, iterations 15, 30, 45 and 60 of all runs of each method. \textbf{(a)}~Advantages: at states where an accepted and a rejected command were both executed, the share of pairs in which the tree-GAE advantage computed with the method's own critic is higher for the accepted command (number of pairs). \textbf{(b)}~Comparisons: at evaluated states with nonzero weight and both kinds of candidate, the share of states at which the comparison prefers an accepted candidate (number of states; open marker: states at full weight). Both axes are compressed below 50\%. The two rates are computed on different sets of states. \qppo{} compares candidates by its learned action value. Acceptance of an unexecuted candidate is read from one fixed world model for all methods; each run's own world model gives nearly the same rates (Table~\ref{tab:adv}). PPO-160 plays one episode per task and has no such states. Advantages on all steps: Table~\ref{tab:adv} in Appendix~\ref{app:advrec}.}
\label{fig:same-state}
\end{figure}

\begin{table}[!ht]
\centering\small
\caption{\textbf{Do the advantages tell useful from useless steps?} Advantages of executed non-terminal steps, recomputed with the critics used in training (iterations 15, 30, 45 and 60 on ScienceWorld; 15, 24, 30 and 36 on ALFWorld; all runs of each method, except Wide-Tree-PPO on ALFWorld, first run only). Useful / useless: accepted / rejected commands on ScienceWorld; steps that take, process or place the task object / wasted steps (``Nothing happens'', or returning to a searched location before the object is found) on ALFWorld. AUC: over all executed steps, the probability that a useful step has the higher advantage; reinforced: share of useless steps with a positive advantage. Same state: at states with one useful and one useless executed step, the share at which the useful one has the higher advantage (number of such pairs; PPO-160 plays one episode per task and has none). For the comparisons of the two \wmppo{} variants: among the states with nonzero weight and both kinds of candidate, the share at which the comparison prefers a useful candidate (at full weight on ScienceWorld, 80\% for the policy-channel-only variant and 78\% for \wmppo{}; on ALFWorld, the variant with the policy channel only prefers a useless candidate at 21\% of the states and a third, neutral candidate at 14\%, the full method a useless one at 26\%). Acceptance of an unexecuted candidate is read from each run's own world model here, and from one fixed world model for all methods in Figure~\ref{fig:same-state}, which gives 66\% and 64\% on ScienceWorld. Counts in Table~\ref{tab:advrec}.}
\label{tab:adv}
\setlength{\tabcolsep}{4pt}
\begin{tabular}{@{}l ccc ccc@{}}
\toprule
& \multicolumn{3}{c}{ScienceWorld} & \multicolumn{3}{c}{ALFWorld} \\
\cmidrule(lr){2-4}\cmidrule(l){5-7}
& AUC & Reinforced & Same state & AUC & Reinforced & Same state \\
\midrule
PPO-160, tree GAE & 0.51 & 39\% & -- & 0.57 & 34\% & -- \\
Wide-Tree-PPO, tree GAE & 0.51 & 45\% & 58\% (476) & 0.62 & 31\% & 86\% (21) \\
\multicolumn{7}{@{}l}{\wmppo{} (policy channel only)} \\
\quad tree GAE & 0.48 & 40\% & 56\% (476) & 0.65 & 29\% & 74\% (43) \\
\quad comparisons & -- & -- & 67\% (1,560) & -- & -- & 66\% (116) \\
\multicolumn{7}{@{}l}{\wmppo{}} \\
\quad tree GAE & 0.51 & 43\% & 61\% (431) & 0.61 & 20\% & 75\% (52) \\
\quad comparisons & -- & -- & 64\% (1,789) & -- & -- & 59\% (219) \\
\bottomrule
\end{tabular}
\end{table}

The first term of the advantage, $\V(s_{t+1})-\V(s_t)$, averages $-0.003$ for rejected and $+0.002$ for accepted commands under PPO-160's critics ($-0.005$ and $+0.003$ for Wide-Tree-PPO, $-0.005$ and $+0.003$ for \wmppo{}, $-0.006$ and $+0.003$ for the policy-only variant). At states where both kinds were executed, the accepted command has the higher advantage in 58\% of 476 pairs for Wide-Tree-PPO, in 61\% of 431 pairs for \wmppo{} and in 56\% of 476 pairs for the policy-only variant; in 77\%, 84\% and 81\% of these pairs, the two advantages differ by less than 0.25 standard deviations. The auxiliary comparisons were examined at the states with nonzero weight at which candidates were predicted to be rejected and accepted (prediction precision 0.99): 1,789 states for \wmppo{} and 1,560 for the policy-only variant. At these states, the comparison prefers an accepted candidate in 64\% for \wmppo{} (78\% at full weight and 60\% at partial weight; mean weight 0.49) and in 67\% for the policy-only variant (80\% and 64\%; mean weight 0.44). In the remaining third, mostly in partially weighted bins, the critic values the predicted successor of the rejected command, which leaves the task where it was, above those of the alternatives. These are comparisons in which the critic values both successors (Table~\ref{tab:branch-split}).

\paragraph{The same measurement on ALFWorld.} For ALFWorld, we recomputed iterations 15, 24, 30 and 36 of the same methods in the same way, and the logs again match to four decimals. Executed non-terminal steps are classified from the true next observation. \emph{Advancing} steps take the task object, heat, cool or clean it, place it in the target receptacle, or use the desk lamp. \emph{Wasted} steps receive \emph{Nothing happens} or go to a location already visited in the episode before the task object has been taken (returning to the target receptacle is not counted). Under PPO-160's critics, the first term of the advantage averages $+0.014$ for advancing steps and $-0.011$ and $-0.009$ for the two kinds of wasted step. The comparisons were examined at the states with nonzero weight at which both an advancing and a wasted candidate were sampled (219 for \wmppo{}, 116 for the policy-only variant); the preferred candidate is advancing at 59\% and 66\% of these states and wasted at 26\% and 21\% (mean weights 0.64 and 0.60). Few states have both kinds of executed step (21--52 pairs per method; Table~\ref{tab:adv}), and these states differ from those of the comparisons. Unlike on ScienceWorld, the advantages already separate the two kinds of step: the gap between their mean advantages is 0.41 standard deviations under PPO-160's critics, against 0.10 on ScienceWorld (Table~\ref{tab:advrec}).

\begin{table}[!ht]
\centering\small
\caption{Counts and mean raw advantages behind Table~\ref{tab:adv}. ScienceWorld: rejected and accepted commands; ALFWorld: wasted (\emph{Nothing happens} / revisits) and advancing steps. Three runs each, except Wide-Tree-PPO on ALFWorld (one run).}
\label{tab:advrec}
\setlength{\tabcolsep}{3pt}
\footnotesize{\begin{tabular}{l r r c c c c}
\toprule
\multicolumn{7}{@{}l}{\emph{ScienceWorld}} \\
Method & Useless & Useful & Mean: useless / useful & Gap & AUC & Useless with $A^{\mathrm{GAE}}>0$ \\
\midrule
PPO-160 & 11,200 & 14,624 & $-0.018$ / $+0.003$ & 0.10 & 0.51 & 39\% \\
Wide-Tree-PPO & 13,167 & 13,739 & $-0.013$ / $+0.004$ & 0.08 & 0.51 & 45\% \\
\wmppo{} & 11,387 & 14,665 & $-0.008$ / $+0.010$ & 0.11 & 0.51 & 43\% \\
\quad policy channel only & 11,410 & 17,150 & $+0.004$ / $+0.006$ & 0.02 & 0.48 & 40\% \\
\midrule
\multicolumn{7}{@{}l}{\emph{ALFWorld} (wasted = \emph{Nothing happens} + revisits)} \\
PPO-160 & 2,909 + 11,288 & 5,007 & $-0.051$, $-0.074$ / $+0.032$ & 0.41 & 0.57 & 34\% \\
Wide-Tree-PPO & 1,154 + 3,826 & 1,769 & $-0.041$, $-0.048$ / $+0.017$ & 0.29 & 0.62 & 31\% \\
\wmppo{} & 2,331 + 12,398 & 4,275 & $-0.053$, $-0.069$ / $-0.009$ & 0.35 & 0.61 & 20\% \\
\quad policy channel only & 1,884 + 10,786 & 5,358 & $-0.025$, $-0.070$ / $+0.019$ & 0.33 & 0.65 & 29\% \\
\bottomrule
\end{tabular}}
\end{table}

\FloatBarrier
\subsection{Candidates at evaluation states}
\label{app:truth}

\begin{figure}[!ht]
\centering
\includegraphics[width=\linewidth]{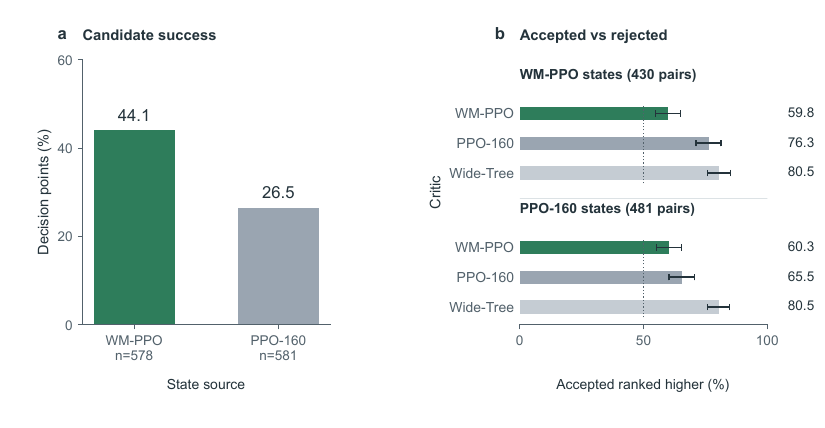}
\caption{\textbf{Candidates at evaluation states} (ScienceWorld, 60 procedural evaluation games, final policies). Every distinct candidate among four samples was executed and then played to the end five times by the same policy; its success rate estimates the value of its next state. \textbf{(a)}~Decision points at which at least one candidate succeeded in one of its five continuations, each policy on its own decision points (not a comparison of the methods). \textbf{(b)}~Pairs of non-terminal candidates at the same state of which the simulator accepted one and rejected the other: share in which a critic values the accepted command's true next state higher (ties count half; 481 pairs on PPO-160's states, 430 on \wmppo{}'s); groups give the policy whose states were used, rows the critic. All critics and states are those of the first run of each method; \wmppo{} is the full method. Horizontal bars: 95\% bootstrap intervals clustered by decision point. The critics' ordering of all candidate pairs is in Figure~\ref{fig:mechanism-b} (Appendix~\ref{app:truth}).}
\label{fig:mechanism}
\end{figure}

\begin{figure}[!ht]
\centering
\includegraphics[width=\linewidth]{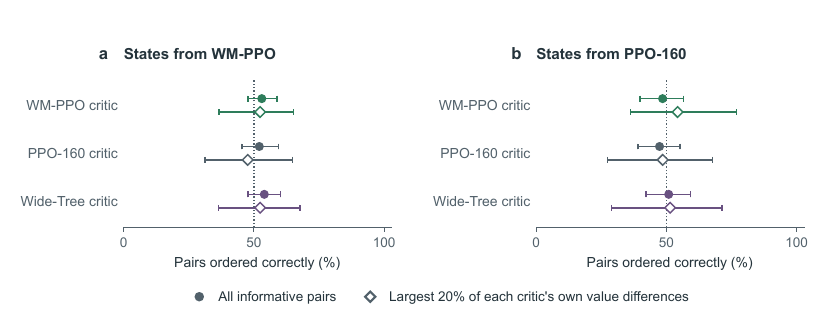}
\caption{\textbf{Critics' ordering of the candidates' true next states} (ScienceWorld, 60 procedural evaluation games, final policies). Agreement with the success rates estimated from five continuations, over candidate pairs whose estimates differ (filled circles; 313 pairs on \wmppo{}'s states, 171 on PPO-160's) and over the 20\% of pairs with the largest gap in each critic's own values (open diamonds; 63 and 35 pairs, selected separately for each critic, so the three diamonds do not refer to the same pairs); \textbf{(a)}~states from \wmppo{}, \textbf{(b)}~states from PPO-160; rows give the critic. Horizontal bars: 95\% bootstrap intervals clustered by decision point.}
\label{fig:mechanism-b}
\end{figure}

At the decision points of the final (iteration-60) PPO-160 and \wmppo{} policies on 60 procedural ScienceWorld evaluation games, every distinct candidate among four samples was executed in a copy of the game, and the same policy then played the episode to the end five times. The success rate estimates the candidate's true value. With four candidates and five continuations each, no candidate led to a success in any continuation at 73.5\% of PPO-160's decision points and at 55.9\% of \wmppo{}'s (Figure~\ref{fig:mechanism}a). At such points, these continuations cannot tell the candidates apart, although the candidates' true values need not be exactly zero. Figure~\ref{fig:mechanism-b} shows how three critics order the true next states of the candidate pairs whose estimates differ. On both sets of decision points, all three critics are close to chance (47--54\% of all such pairs and 48--54\% of the 20\% with the largest value gap). With five continuations, each estimate has a standard error of up to about 0.2.

\paragraph{Rejected commands in ScienceWorld: detection.}\label{app:rejected} During training, on the executed steps of the evaluated decision points, the online world model of \wmppo{} (policy channel only) predicts 98--99\% of the rejections with a precision of 0.99 (Table~\ref{tab:wmq}). On evaluation states, the recall is lower. On 80 procedural evaluation games, the final \wmppo{} and PPO-160 policies played as in training. At every decision point, each distinct candidate among four samples was executed in a copy of the game and also predicted by a final world model. Of 1,717 candidates on \wmppo{}'s decision points, 670 were rejected by the simulator; the world model of the first \wmppo{} run predicted a rejection for 582 candidates, 580 of them correctly (recall 0.87, precision 1.00). On PPO-160's decision points, the world model of the first policy-channel-only run finds the rejections among 664 of 1,760 candidates with recall 0.84 and precision 0.99.

\paragraph{Rejected commands: comparisons during training.} Of the 17,385 real branch pairs of iterations 13--60 in the three runs of \wmppo{} (policy channel only), 3,835 have exactly one action predicted to be rejected and no predicted ending. The critic of that iteration valued the accepted action higher in 67\% of them (59\% in the smallest-margin bin, 73\% in the middle bin and 74\% in the largest). By phase, the rate is 64\%, 70\%, 69\% and 67\% (iterations 13--24, 25--36, 37--48 and 49--60), and in the largest bin of iterations 49--60 it is 78\%. In the three runs of \wmppo{}, 3,971 pairs qualify; the rate is 69\% (60\%, 72\% and 81\% by margin bin; 63\%, 72\%, 74\% and 68\% by phase), and in the largest bin of iterations 49--60 it is 90\% (171 pairs).

\paragraph{Rejected commands: final critics.} On the evaluation states of the true-outcome test (Appendix~\ref{app:truth}; Figure~\ref{fig:mechanism}b), we consider pairs of non-terminal candidates of which the simulator accepted one and rejected the other. On PPO-160's decision points (481 pairs), the iteration-60 critics value the accepted command's true next state higher in 65\% (PPO-160), 80\% (Wide-Tree-PPO) and 60\% (\wmppo{}) of the pairs; on \wmppo{}'s decision points (430 pairs), the rates are 76\%, 80\% and 60\% (53\% for the critic of a second \wmppo{} run). A second run of the test on the 80 procedural evaluation games described above gives the same ordering (PPO-160 72--76\%, Wide-Tree-PPO 80--83\%, \wmppo{} 58--59\%). The critics of PPO-160 and Wide-Tree-PPO separate rejected commands more sharply, while their policies keep issuing them (37.5\% and 40.2\% of procedural commands late in training, means over runs); the policy of \wmppo{} issues fewer (29.5\%). The final critic of \wmppo{} (policy channel only) orders 43--48\% of these pairs correctly (45--47\% in the separate test).

\FloatBarrier
\subsection{Behavior of the trained policies}
\label{app:behavior}

Section~\ref{sec:behavior} takes success rates from the official evaluation and behavior rates from greedy re-runs of the final checkpoints that log every step. Greedy decoding on the inference server is not bit-exact, and the re-runs reproduce the official per-game outcome for 94--99\% of the ALFWorld and 91.5--96.5\% of the ScienceWorld games. The pooled behavior differences of Figure~\ref{fig:endings}b do not hold in every run: one PPO-160 run completes the processing in all 91 of its picked-up episodes and places the object before processing it in 10.

Tables~\ref{tab:beh-sw} and~\ref{tab:beh-alf} are computed from the training chains (the episode's own path, without branches), decoded into actions and observations. Table~\ref{tab:beh-sw} covers procedural ScienceWorld tasks in the first and last windows of training for nine methods, and Table~\ref{tab:beh-alf} covers ALFWorld in iterations 34--36, pooling the runs of each method. One pattern is not discussed method by method in Section~\ref{sec:behavior}: the tuned \sopo{} ends slightly fewer ScienceWorld evaluation episodes with a failure than \wmppo{} (27.1\% against 30.3\%) but more at the step limit (26.4\% against 16.8\%).

\begin{figure}[!ht]
\centering
\includegraphics[width=0.55\linewidth]{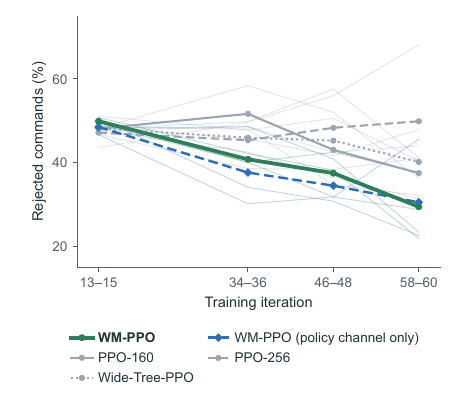}
\caption{\textbf{Rejected commands during training} (ScienceWorld, procedural tasks of the training episodes, chains only, windows of three iterations). Thin gray lines: individual runs; colored lines: mean over the runs of each method. Other methods: Table~\ref{tab:beh-sw}.}
\label{fig:behavior}
\end{figure}

\begin{table}[!ht]
\centering\small
\caption{ScienceWorld, procedural tasks of the training episodes (\% of commands, or of episodes for wrong-focus endings).}
\label{tab:beh-sw}
\resizebox{\linewidth}{!}{\begin{tabular}{l c c c c c c}
\toprule
 & \multicolumn{2}{c}{Rejected commands} & \multicolumn{2}{c}{Repeated commands} & \multicolumn{2}{c}{Ended by a wrong focus} \\
\cmidrule(lr){2-3}\cmidrule(lr){4-5}\cmidrule(lr){6-7}
Method & it.~13--15 & it.~58--60 & it.~13--15 & it.~58--60 & it.~13--15 & it.~58--60 \\
\midrule
\wmppo{} (3 runs) & 49.9 & 29.5 & 65.1 & 54.0 & 64.0 & 28.6 \\
\wmppo{} (policy channel only, 3 runs) & 48.5 & 30.5 & 64.6 & 55.3 & 65.9 & 32.5 \\
PPO-160 (3 runs) & 48.1 & 37.5 & 65.3 & 57.7 & 62.1 & 36.2 \\
PPO-256 (3 runs) & 47.2 & 49.9 & 65.3 & 65.0 & 62.4 & 33.6 \\
OCSD (self teacher) & 51.5 & 43.8 & 65.7 & 66.7 & 60.1 & 28.3 \\
\qppo{} (3 runs) & 50.0 & 40.7 & 66.2 & 52.9 & 62.1 & 32.4 \\
\sopo{} (tuned clip) & 49.6 & 34.4 & 67.3 & 55.2 & 60.5 & 28.6 \\
Wide-Tree-PPO (3 runs) & 48.2 & 40.2 & 64.4 & 57.2 & 63.7 & 44.9 \\
Fixed WM & 50.0 & 17.8 & 65.2 & 52.5 & 63.0 & 62.6 \\
\bottomrule
\end{tabular}
}
\end{table}

\begin{figure}[!ht]
\centering
\includegraphics[width=0.62\linewidth]{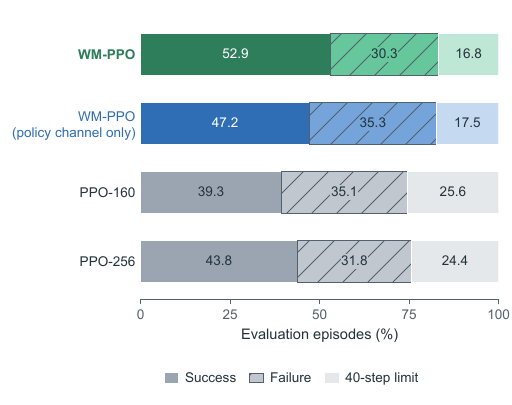}
\caption{\textbf{How ScienceWorld evaluation episodes end} (final checkpoints, iteration 60, greedy, 284 games, mean over three runs): success, failure declared by the simulator, or the 40-step limit. \wmppo{} ends fewer episodes than PPO-160 with a failure and fewer at the step limit.}
\label{fig:endings-sw}
\end{figure}

\begin{table}[!ht]
\centering\small
\caption{ALFWorld training episodes of iterations 34--36. Repeated actions: \% of steps whose action was already taken earlier in the episode; the last column counts \texttt{go to} actions before the first \texttt{take} in successful episodes.}
\label{tab:beh-alf}
\resizebox{\linewidth}{!}{\begin{tabular}{l c c c c c}
\toprule
Method & Success & Steps (successful) & Repeated actions & \texttt{go to} before first \texttt{take} (successful) \\
\midrule
\wmppo{} (ours) & 83.9 & 16.5 (12.0) & 42.4 & 3.4 \\
\wmppo{} (policy channel only) & 86.8 & 16.0 (12.4) & 38.7 & 3.6 \\
PPO-160 & 64.7 & 23.3 (14.1) & 53.2 & 4.5 \\
PPO-256 & 83.2 & 18.1 (13.7) & 42.9 & 3.9 \\
\qppo{} & 86.5 & 16.4 (12.7) & 40.0 & 3.8 \\
Wide-Tree-PPO & 77.3 & 19.5 (13.4) & 47.1 & 4.0 \\
\bottomrule
\end{tabular}
}
\end{table}

\begin{table}[!ht]
\centering\footnotesize
\caption{Further evaluation games (greedy re-runs of the final checkpoints; successes over the three runs of each method).}
\label{tab:cases-extra}
\renewcommand{\arraystretch}{1.12}
\begin{tabularx}{\linewidth}{@{}>{\raggedright\arraybackslash}p{0.24\linewidth} c c >{\raggedright\arraybackslash}X@{}}
\toprule
Game & \wmppo{} & PPO-160 & What differs \\
\midrule
ALFWorld: heat an apple, put it on the dining table & 3/3 & 0/3 & \wmppo{} takes the apple and heats it in the microwave (7 steps); PPO-160 never takes the apple and circles the microwave, fridge and stove burners \\
ALFWorld: clean a cloth, put it in a cabinet & 3/3 & 0/3 & \wmppo{} finds the cloth in the first cabinet it opens and cleans it at the sink (8 steps); PPO-160 alternates between sink and cabinets without taking it \\
ALFWorld: put a pepper shaker in a drawer & 3/3 & 0/3 & \wmppo{} checks the countertops, where the pepper shaker is (7 steps); PPO-160 opens drawers and cabinets in turn and never finds it \\
\bottomrule
\end{tabularx}
\end{table}

\paragraph{Premature focus.} Figure~\ref{fig:endings}d counts, in the greedy re-runs of the 188 procedural ScienceWorld games, the episodes whose first accepted \texttt{focus on} command comes before the state that the answer depends on exists; the definition was fixed per task type before counting. In the chemistry tasks (52 games), the first focus precedes any \texttt{mix} whose feedback reports a product; in the melting-point task (18), the focus on an answer box precedes any thermometer reading; in the find tasks (72), the first focus is on an object that does not qualify, and the simulator ends the episode as failed. The remaining 46 procedural games ask for a focus first or are answered by placing an object, so they cannot contain a premature focus. The re-runs reproduce the official per-game outcome for 89--97\% of the games. Over all procedural games, the premature share is 22.0\% for \wmppo{}, 26.8\% for PPO-256, 28.7\% for PPO-160 and 28.7\% for the policy-channel-only variant (per run 13.3--35.1\%, 23.9--30.9\%, 25.5--33.0\% and 25.5--34.0\%); per game, the share of \wmppo{} is lower than that of PPO-256 in 37 games and higher in 25. Split by task type after this count, it is 6.9\%, 19.0\%, 19.4\% and 21.3\% on the find tasks (per game, \wmppo{} is lower than PPO-256 in 14 games and higher in 1), 64.8\%, 79.6\%, 74.1\% and 83.3\% on the melting-point task, and 47.4\%, 42.9\%, 51.3\% and 45.5\% on the chemistry tasks, where one \wmppo{} run focuses prematurely in all of its chemistry episodes. A premature focus ends in failure in the chemistry and find tasks; in the melting-point task, 20 of 35 such \wmppo{} episodes still name the right box.

\paragraph{Further cases.} Table~\ref{tab:cases-extra} lists further evaluation games from the greedy re-runs in which the decisive difference is an intermediate operation or the direction of search. At iteration 31 of a run with the policy channel only, the policy held a dirty pot and sampled \texttt{go to sinkbasin 1} twice and \texttt{move pot 1 to stoveburner 2} twice. The world model predicted \emph{You arrive at sinkbasin 1\dots} and \emph{You move the pot 1 to the stoveburner 2}, and the critic valued the two predicted successors at 0.952 and 0.950. The bin of this small margin had weight 0.85 in that iteration, so the four samples received auxiliary advantages of $+0.42$, $+0.42$, $-0.42$ and $-0.42$. PPO-160 made the second move in the same game one iteration later and failed. In one further ALFWorld game (heating an apple), the final world model and critic of one run rank the fridge slightly above the microwave (1.051 against 1.044; another run ranks the microwave first), although \wmppo{} succeeds in all three runs. We therefore report these additional cases without assigning decisive-state values to them.

\section{Inputs and outputs of the models}
\label{app:io-sec}
\label{app:io}

In the training example of Figure~\ref{fig:io}, the task requires making orange paint and then focusing on it. The bowl currently contains red and yellow paint, but the orange paint does not yet exist. The world model predicts that `mix bowl' will produce orange paint, whereas `focus on orange paint' will be rejected and leave the bowl unchanged; both predictions match the environment. The critic assigns a higher transition value to the successor in which the paint has been made. Although both real continuations ultimately succeed, these immediate consequences supply a preference for mixing before focusing. Of the four sampled responses, the three `mix bowl' samples each receive $+\tfrac14$ and the `focus on orange paint' sample receives $-\tfrac34$, times the bin's weight. These auxiliary advantages raise the log-probability of the mixing responses relative to the focus response (Equation~\eqref{eq:aux-grad}).

\begin{figure}[H]
\centering\scriptsize
\setlength{\fboxsep}{3pt}
\begin{tabularx}{\linewidth}{@{}>{\raggedright\arraybackslash}p{0.17\linewidth} >{\raggedright\arraybackslash}X >{\raggedright\arraybackslash}X@{}}
\toprule
& \textcolor{cpol}{\textbf{Candidate}} \texttt{mix bowl} & \textcolor{cpol}{\textbf{Candidate}} \texttt{focus on orange paint} \\
\midrule
\textcolor{creal}{Current observation} (both) & \multicolumn{2}{>{\raggedright\arraybackslash}p{0.775\linewidth}}{\texttt{You pour the contents of the wood cup into the bowl.} \texttt{[Current room]} \dots\ \texttt{a table. On the table is: a bowl (containing red paint, yellow paint).} \dots\ \texttt{[score: 40]}} \\
\addlinespace
\textcolor{cwm}{World model predicts} $\hat o_{t+1}(a)$ & \texttt{Yellow and red paint mix to produce orange paint.} \texttt{[Current room]} \dots\ \texttt{a bowl (containing orange paint).} \dots\ \texttt{[score: 50]} & \texttt{No known action matches that input.} \texttt{[Current room]} \dots\ \texttt{a bowl (containing red paint, yellow paint).} \dots\ \texttt{[score: 40]} \\
\addlinespace
\textcolor{ccri}{Critic reads} $\hat s_{t+1}(a)$ & \multicolumn{2}{>{\raggedright\arraybackslash}p{0.775\linewidth}}{task and rules (copied); step counter and last five observation--action pairs, now ending with the real current observation and the candidate (updated by the program); $\hat o_{t+1}(a)$ as the current observation (generated); referable objects from before the step, still without orange paint (copied)} \\
\addlinespace
Comparison & \multicolumn{2}{l}{$\qhat_t(\texttt{mix bowl})-\qhat_t(\texttt{focus on orange paint})=+0.153$, in the large-margin bin} \\
\bottomrule
\end{tabularx}
\caption{\textbf{From predicted changes to transition values at a training decision point} (ScienceWorld, iteration 59 of a run with the policy channel only). Predictions are reproduced verbatim (``\dots'' marks identical room lines); each matches the observation returned when that action was executed in the chain or a branch. The world model predicts the whole next observation, including the progress score. Neither prediction ends the episode, so both transition values come from the critic's evaluation of the predicted successors. The complete inputs and outputs of two decision points follow below.}
\label{fig:io}
\end{figure}

\begin{table}[H]
\centering\scriptsize
\caption{\textbf{What each model reads and writes.} Generated: produced by the world model; copied: taken unchanged from $s_t$; program: updated deterministically by the prompt constructor $C$.}
\label{tab:fields}
\begin{tabularx}{\linewidth}{@{}>{\raggedright\arraybackslash}p{0.15\linewidth} >{\raggedright\arraybackslash}X >{\raggedright\arraybackslash}X >{\raggedright\arraybackslash}p{0.16\linewidth}@{}}
\toprule
 & ALFWorld & ScienceWorld & Source \\
\midrule
Policy and critic input $s_t$ & task; step counter; last 5 observation--action pairs; current observation; admissible actions & task and rules; step counter; last 5 observation--action pairs; current observation (action feedback, room, inventory, score); action templates; referable objects & real interaction \\
World-model input $(x_t, a)$ & task; initial observation; all previous observations and actions; current observation; admissible actions; candidate & task; last 8 actions, without observations; current observation; referable objects; candidate & real interaction \\
World-model output $\hat o_{t+1}$ & next observation as the environment prints it & action feedback, \texttt{[Current room]}, \texttt{[Inventory]}, \texttt{[score: S]}, and \texttt{[task completed]} or \texttt{[task failed]} when the simulator ends the task & generated \\
Successor state $\hat s_{t+1}=C(s_t,a,\hat o_{t+1})$ & policy format: window of recent pairs ends with (real current observation, candidate); $\hat o_{t+1}$ as current observation; admissible actions from before the step & the same; referable objects from before the step & program (window, step counter); copied (task, lists); generated ($\hat o_{t+1}$) \\
Predicted ending & none (observations report no endings) & an end marker in $\hat o_{t+1}$ replaces the critic's value by 1 (completed) or 0 (failed) & parsed from $\hat o_{t+1}$ \\
Progress score & none & printed in every observation; read by the policy and the critic; part of the world model's input and of its next-observation target; not used as a reward or as a separate training target & environment text \\
Reward & 1 if completed within 40 actions, else 0 & 1 if the simulator reports completion within 40 actions, else 0 & environment \\
\bottomrule
\end{tabularx}
\vspace{-2pt}
\end{table}

Table~\ref{tab:fields} distinguishes the simulator's internal state, which no model sees; the prompt state $s_t$ read by the policy and critic; and the next observation $\hat o_{t+1}$ generated by the world model. Two evaluated decision points from the training logs follow verbatim. The world-model input is not logged as text; we rebuilt it from the logged transition record with the training prompt function, which reproduces the logged policy prompt exactly. The prediction and the critic input are decoded from the logged token ids, and re-assembling the critic input from the decoded prediction reproduces those ids. The raw world-model generation and the critic's individual values of predicted successors are not logged; the trainer logs only the value difference of each branch pair.

\subsubsection*{ALFWorld: training iteration 48, step 14}
\begin{sloppypar}
Run 1 of \wmppo{} (policy channel only); models of iteration 47; training game \texttt{train/\allowbreak{}pick\_\allowbreak{}cool\_\allowbreak{}then\_\allowbreak{}place\_\allowbreak{}in\_\allowbreak{}recep-\allowbreak{}Pot-\allowbreak{}None-\allowbreak{}CounterTop-\allowbreak{}24/\allowbreak{}trial\_\allowbreak{}T20190907\_\allowbreak{}002332\_\allowbreak{}305243}. The policy sampled \texttt{move pan 1 to fridge 1}, \texttt{cool pan 1 with fridge 1}, \texttt{cool pan 1 with fridge 1}, \texttt{cool pan 1 with fridge 1}; \texttt{move pan 1 to fridge 1} was executed in the chain and \texttt{cool pan 1 with fridge 1} in a branch from the same state. Logged difference of transition values $\qhat(\text{chain})-\qhat(\text{branch})=+0.137$ (margin bin 3 of 3); real returns 1 (chain) and 0 (branch). Both predictions equal the observations the environment printed; this is not representative of predictions that involve content the agent has not seen.
\end{sloppypar}

\noindent\emph{(i) World-model input $(x_t,a)$ for candidate \texttt{move pan 1 to fridge 1} (rebuilt token-exactly from the logged record; the input for the other candidate differs only in its last line)}
\begin{lstlisting}[basicstyle=\fontsize{6.5}{7.6}\selectfont\ttfamily]
<|im_start|>system
You are a world model of the ALFRED text household environment.<|im_end|>
<|im_start|>user
Task: put a cool pot in countertop.
Initial observation: You are in the middle of a room. Looking quickly around you, you see a cabinet 23, a cabinet 22, a cabinet 21, a cabinet 20, a cabinet 19, a cabinet 18, a cabinet 17, a cabinet 16, a cabinet 15, a cabinet 14, a cabinet 13, a cabinet 12, a cabinet 11, a cabinet 10, a cabinet 9, a cabinet 8, a cabinet 7, a cabinet 6, a cabinet 5, a cabinet 4, a cabinet 3, a cabinet 2, a cabinet 1, a coffeemachine 1, a countertop 3, a countertop 2, a countertop 1, a diningtable 1, a drawer 4, a drawer 3, a drawer 2, a drawer 1, a fridge 1, a garbagecan 1, a microwave 1, a sinkbasin 1, a stoveburner 4, a stoveburner 3, a stoveburner 2, a stoveburner 1, and a toaster 1.
History of steps so far: [Step 1: observation 'You are in the middle of a room. Looking quickly around you, you see a cabinet 23, a cabinet 22, a cabinet 21, a cabinet 20, a cabinet 19, a cabinet 18, a cabinet 17, a cabinet 16, a cabinet 15, a cabinet 14, a cabinet 13, a cabinet 12, a cabinet 11, a cabinet 10, a cabinet 9, a cabinet 8, a cabinet 7, a cabinet 6, a cabinet 5, a cabinet 4, a cabinet 3, a cabinet 2, a cabinet 1, a coffeemachine 1, a countertop 3, a countertop 2, a countertop 1, a diningtable 1, a drawer 4, a drawer 3, a drawer 2, a drawer 1, a fridge 1, a garbagecan 1, a microwave 1, a sinkbasin 1, a stoveburner 4, a stoveburner 3, a stoveburner 2, a stoveburner 1, and a toaster 1.', action 'go to countertop 1'] [Step 2: observation 'You arrive at countertop 1. On the countertop 1, you see a bread 1, and a spatula 1.', action 'go to countertop 2'] [Step 3: observation 'You arrive at countertop 2. On the countertop 2, you see a pan 1.', action 'go to countertop 3'] [Step 4: observation 'You arrive at countertop 3. On the countertop 3, you see a fork 3, a knife 1, and a mug 1.', action 'go to diningtable 1'] [Step 5: observation 'You arrive at diningtable 1. On the diningtable 1, you see a apple 1, a butterknife 1, a cup 1, a egg 2, a knife 2, a lettuce 1, a mug 2, a papertowelroll 1, a potato 3, a potato 2, a potato 1, a soapbottle 2, a spatula 2, a spoon 3, a spoon 1, and a tomato 1.', action 'go to fridge 1'] [Step 6: observation 'You arrive at fridge 1. The fridge 1 is closed.', action 'open fridge 1'] [Step 7: observation 'You open the fridge 1. The fridge 1 is open. In it, you see nothing.', action 'go to microwave 1'] [Step 8: observation 'You arrive at microwave 1. The microwave 1 is closed.', action 'open microwave 1'] [Step 9: observation 'You open the microwave 1. The microwave 1 is open. In it, you see a egg 1.', action 'go to countertop 1'] [Step 10: observation 'You arrive at countertop 1. On the countertop 1, you see a bread 1, and a spatula 1.', action 'go to countertop 2'] [Step 11: observation 'You arrive at countertop 2. On the countertop 2, you see a pan 1.', action 'take pan 1 from countertop 2'] [Step 12: observation 'You pick up the pan 1 from the countertop 2.', action 'go to countertop 3'] [Step 13: observation 'You arrive at countertop 3. On the countertop 3, you see a fork 3, a knife 1, and a mug 1.', action 'go to fridge 1']
Current observation: You arrive at fridge 1. The fridge 1 is open. In it, you see nothing.
Admissible actions: ['close fridge 1', 'cool pan 1 with fridge 1', 'examine fridge 1', 'examine pan 1', 'go to cabinet 1', 'go to cabinet 10', 'go to cabinet 11', 'go to cabinet 12', 'go to cabinet 13', 'go to cabinet 14', 'go to cabinet 15', 'go to cabinet 16', 'go to cabinet 17', 'go to cabinet 18', 'go to cabinet 19', 'go to cabinet 2', 'go to cabinet 20', 'go to cabinet 21', 'go to cabinet 22', 'go to cabinet 23', 'go to cabinet 3', 'go to cabinet 4', 'go to cabinet 5', 'go to cabinet 6', 'go to cabinet 7', 'go to cabinet 8', 'go to cabinet 9', 'go to coffeemachine 1', 'go to countertop 1', 'go to countertop 2', 'go to countertop 3', 'go to diningtable 1', 'go to drawer 1', 'go to drawer 2', 'go to drawer 3', 'go to drawer 4', 'go to garbagecan 1', 'go to microwave 1', 'go to sinkbasin 1', 'go to stoveburner 1', 'go to stoveburner 2', 'go to stoveburner 3', 'go to stoveburner 4', 'go to toaster 1', 'inventory', 'look', 'move pan 1 to fridge 1']
Action taken now: move pan 1 to fridge 1

Predict the environment's next observation exactly as the environment would print it, inside <obs> </obs>.<|im_end|>
<|im_start|>assistant
\end{lstlisting}

\par\noindent\begin{minipage}{\linewidth}
\noindent\emph{(ii) World-model output $\hat o_{t+1}(a)$ for \texttt{move pan 1 to fridge 1} (generated; identical to the true next observation)}
\begin{lstlisting}[basicstyle=\fontsize{6.5}{7.6}\selectfont\ttfamily]
You move the pan 1 to the fridge 1.
\end{lstlisting}
\end{minipage}

\par\noindent\begin{minipage}{\linewidth}
\noindent\emph{(iii) World-model output $\hat o_{t+1}(a)$ for \texttt{cool pan 1 with fridge 1} (generated; identical to the true next observation)}
\begin{lstlisting}[basicstyle=\fontsize{6.5}{7.6}\selectfont\ttfamily]
You cool the pan 1 using the fridge 1.
\end{lstlisting}
\end{minipage}

\noindent\emph{(iv) Critic input $\hat s_{t+1}(a)=C(s_t,a,\hat o_{t+1}(a))$ for \texttt{move pan 1 to fridge 1} (decoded from the logged ids; the value head reads the last token)}
\begin{lstlisting}[basicstyle=\fontsize{6.5}{7.6}\selectfont\ttfamily]
<|im_start|>system
You are an expert agent operating in the ALFRED embodied household environment.<|im_end|>
<|im_start|>user
Your task is to: put a cool pot in countertop.
Prior to this step, you have already taken 14 step(s). Below are the most recent 5 observations and the corresponding actions you took: [Observation 10: 'You arrive at countertop 1. On the countertop 1, you see a bread 1, and a spatula 1.', Action 10: 'go to countertop 2'] [Observation 11: 'You arrive at countertop 2. On the countertop 2, you see a pan 1.', Action 11: 'take pan 1 from countertop 2'] [Observation 12: 'You pick up the pan 1 from the countertop 2.', Action 12: 'go to countertop 3'] [Observation 13: 'You arrive at countertop 3. On the countertop 3, you see a fork 3, a knife 1, and a mug 1.', Action 13: 'go to fridge 1'] [Observation 14: 'You arrive at fridge 1. The fridge 1 is open. In it, you see nothing.', Action 14: 'move pan 1 to fridge 1']
You are now at step 15 and your current observation is: You move the pan 1 to the fridge 1.
Your admissible actions of the current situation are: ['close fridge 1', 'cool pan 1 with fridge 1', 'examine fridge 1', 'examine pan 1', 'go to cabinet 1', 'go to cabinet 10', 'go to cabinet 11', 'go to cabinet 12', 'go to cabinet 13', 'go to cabinet 14', 'go to cabinet 15', 'go to cabinet 16', 'go to cabinet 17', 'go to cabinet 18', 'go to cabinet 19', 'go to cabinet 2', 'go to cabinet 20', 'go to cabinet 21', 'go to cabinet 22', 'go to cabinet 23', 'go to cabinet 3', 'go to cabinet 4', 'go to cabinet 5', 'go to cabinet 6', 'go to cabinet 7', 'go to cabinet 8', 'go to cabinet 9', 'go to coffeemachine 1', 'go to countertop 1', 'go to countertop 2', 'go to countertop 3', 'go to diningtable 1', 'go to drawer 1', 'go to drawer 2', 'go to drawer 3', 'go to drawer 4', 'go to garbagecan 1', 'go to microwave 1', 'go to sinkbasin 1', 'go to stoveburner 1', 'go to stoveburner 2', 'go to stoveburner 3', 'go to stoveburner 4', 'go to toaster 1', 'inventory', 'look', 'move pan 1 to fridge 1'].

Now it's your turn to take an action. First reason briefly (at most two sentences) about the current situation inside <think> </think> tags. Then choose one admissible action for the current step and present it inside <action> </action> tags.<|im_end|>
<|im_start|>assistant
\end{lstlisting}

\begin{center}\scriptsize
\begin{tabularx}{\linewidth}{@{}>{\raggedright\arraybackslash}p{0.42\linewidth} >{\raggedright\arraybackslash}X@{}}
\toprule
Field of the critic input & Source \\
\midrule
\texttt{system message} & fixed (prompts.SYSTEM) \\
\texttt{'Your task is to: \textless{}task\textgreater{}'} & copied from h (same task string) \\
\texttt{'Prior to this step, you have already taken \textless{}t+1\textgreater{} step(s)'} & updated by the program (len(history) after appending (obs, a)) \\
\texttt{'Below are the most recent \textless{}k\textgreater{} observations ...: [Observation i: ..., Action i: ...]'} & updated by the program: last 5 pairs of history + (real current observation, candidate action); all texts copied, none generated \\
\texttt{'You are now at step \textless{}t+2\textgreater{} and your current observation is: \textless{}o\_hat\textgreater{}'} & step number: program; \textless{}o\_hat\textgreater{}: GENERATED by the WM \\
\texttt{'Your admissible actions of the current situation are: [...]'} & COPIED from h (the list before the action; not predicted, not updated) \\
\texttt{instruction sentence} & fixed template \\
\bottomrule
\end{tabularx}
\end{center}

\subsubsection*{ScienceWorld: training iteration 59, step 24}
\begin{sloppypar}
Run 1 of \wmppo{} (policy channel only); models of iteration 58; training game \texttt{chemistry-\allowbreak{}mix-\allowbreak{}paint-\allowbreak{}secondary-\allowbreak{}color|\allowbreak{}16}. The policy sampled \texttt{mix bowl}, \texttt{mix bowl}, \texttt{mix bowl}, \texttt{focus on orange paint}; \texttt{mix bowl} was executed in the chain and \texttt{focus on orange paint} in a branch from the same state. Logged difference of transition values $\qhat(\text{chain})-\qhat(\text{branch})=+0.153$ (margin bin 3 of 3); real returns 1 (chain) and 1 (branch). Both predictions equal the observations the environment printed; this is not representative of predictions that involve content the agent has not seen.
\end{sloppypar}

\noindent\emph{(i) World-model input $(x_t,a)$ for candidate \texttt{mix bowl} (rebuilt token-exactly from the logged record; the input for the other candidate differs only in its last line)}
\begin{lstlisting}[basicstyle=\fontsize{6.5}{7.6}\selectfont\ttfamily]
<|im_start|>system
You are a world model of the ScienceWorld text environment.<|im_end|>
<|im_start|>user
Task: use chemistry to create orange paint. When you are done, focus on the orange paint.
Recent actions: [Step 1: action 'teleport to hallway'] [Step 2: action 'teleport to art studio'] [Step 3: action 'mix cup containing red paint and cup containing yellow paint'] [Step 4: action 'mix cup containing red paint and cup containing yellow paint'] [Step 5: action 'mix cup containing red paint and cup containing yellow paint'] [Step 6: action 'mix cup containing red paint and cup containing yellow paint'] [Step 7: action 'pour cup containing red paint into bowl'] [Step 8: action 'pour cup containing yellow paint into bowl']
Current observation:
You pour the contents of the wood cup into the bowl.
[Current room]
This room is called the art studio. In it, you see: 
    a large cupboard. The large cupboard door is closed. 
    a substance called air
    a table. On the table is: a bowl (containing red paint, yellow paint).
    a wood cup (containing blue paint)
    a wood cup (containing nothing)
    a wood cup (containing nothing)
    the agent
You also see:
    A door to the hallway (that is open)
[Inventory]
In your inventory, you see:
    an orange
[score: 40]
Objects the agent can refer to: [agent, air, art studio, blue paint, bowl, cup containing blue paint, cup containing nothing in art studio, cup on room, cupboard, door, hallway, red paint, table, yellow paint]
Action taken now: mix bowl

Predict the environment's next observation exactly as the environment would print it (action feedback, [Current room], [Inventory] and [score] lines), inside <obs> </obs>.<|im_end|>
<|im_start|>assistant
\end{lstlisting}

\noindent\emph{(ii) World-model output $\hat o_{t+1}(a)$ for \texttt{mix bowl} (generated; identical to the true next observation)}
\begin{lstlisting}[basicstyle=\fontsize{6.5}{7.6}\selectfont\ttfamily]
Yellow and red paint mix to produce orange paint.
[Current room]
This room is called the art studio. In it, you see: 
    a large cupboard. The large cupboard door is closed. 
    a substance called air
    a table. On the table is: a bowl (containing orange paint).
    a wood cup (containing blue paint)
    a wood cup (containing nothing)
    a wood cup (containing nothing)
    the agent
You also see:
    A door to the hallway (that is open)
[Inventory]
In your inventory, you see:
    an orange
[score: 50]
\end{lstlisting}

\noindent\emph{(iii) World-model output $\hat o_{t+1}(a)$ for \texttt{focus on orange paint} (generated; identical to the true next observation)}
\begin{lstlisting}[basicstyle=\fontsize{6.5}{7.6}\selectfont\ttfamily]
No known action matches that input.
[Current room]
This room is called the art studio. In it, you see: 
    a large cupboard. The large cupboard door is closed. 
    a substance called air
    a table. On the table is: a bowl (containing red paint, yellow paint).
    a wood cup (containing blue paint)
    a wood cup (containing nothing)
    a wood cup (containing nothing)
    the agent
You also see:
    A door to the hallway (that is open)
[Inventory]
In your inventory, you see:
    an orange
[score: 40]
\end{lstlisting}

\noindent\emph{(iv) Critic input $\hat s_{t+1}(a)=C(s_t,a,\hat o_{t+1}(a))$ for \texttt{mix bowl} (decoded from the logged ids; the value head reads the last token)}
\begin{lstlisting}[basicstyle=\fontsize{6.5}{7.6}\selectfont\ttfamily]
<|im_start|>system
You are an expert agent operating in the ScienceWorld environment, a text-based simulator of elementary science experiments.<|im_end|>
<|im_start|>user
Your task is to: use chemistry to create orange paint. When you are done, focus on the orange paint.
Rules: 'focus on OBJ' declares which object the task is about and cannot be undone; focusing on a wrong object ends the task as a failure, so explore first and focus only on the object the task asks about. OBJ must be an object you can refer to now (listed below); LOC is a room name (teleport moves you there directly). Answering questions such as 'which one ...' is done by focusing on the answer object.
Prior to this step, you have already taken 24 step(s). Below are the most recent 5 observations and the corresponding actions you took:
[Observation 20: 'No known action matches that input.
[Current room]
This room is called the art studio. In it, you see: 
    a large cupboard. The large cupboard door is closed. 
    a substance called air
    a table. On the table is: a bowl (containing nothing).
    a wood cup (containing blue paint)
    a wood cup (containing red paint)
    a wood cup (containing yellow paint)
    the agent
You also see:
    A door to the hallway (that is open)
[Inventory]
In your inventory, you see:
    an orange
[score: 30]', Action 20: 'mix cup containing red paint and cup containing yellow paint']
[Observation 21: 'No known action matches that input.
[Current room]
This room is called the art studio. In it, you see: 
    a large cupboard. The large cupboard door is closed. 
    a substance called air
    a table. On the table is: a bowl (containing nothing).
    a wood cup (containing blue paint)
    a wood cup (containing red paint)
    a wood cup (containing yellow paint)
    the agent
You also see:
    A door to the hallway (that is open)
[Inventory]
In your inventory, you see:
    an orange
[score: 30]', Action 21: 'mix cup containing red paint and cup containing yellow paint']
[Observation 22: 'No known action matches that input.
[Current room]
This room is called the art studio. In it, you see: 
    a large cupboard. The large cupboard door is closed. 
    a substance called air
    a table. On the table is: a bowl (containing nothing).
    a wood cup (containing blue paint)
    a wood cup (containing red paint)
    a wood cup (containing yellow paint)
    the agent
You also see:
    A door to the hallway (that is open)
[Inventory]
In your inventory, you see:
    an orange
[score: 30]', Action 22: 'pour cup containing red paint into bowl']
[Observation 23: 'You pour the contents of the wood cup into the bowl.
[Current room]
This room is called the art studio. In it, you see: 
    a large cupboard. The large cupboard door is closed. 
    a substance called air
    a table. On the table is: a bowl (containing red paint).
    a wood cup (containing blue paint)
    a wood cup (containing nothing)
    a wood cup (containing yellow paint)
    the agent
You also see:
    A door to the hallway (that is open)
[Inventory]
In your inventory, you see:
    an orange
[score: 30]', Action 23: 'pour cup containing yellow paint into bowl']
[Observation 24: 'You pour the contents of the wood cup into the bowl.
[Current room]
This room is called the art studio. In it, you see: 
    a large cupboard. The large cupboard door is closed. 
    a substance called air
    a table. On the table is: a bowl (containing red paint, yellow paint).
    a wood cup (containing blue paint)
    a wood cup (containing nothing)
    a wood cup (containing nothing)
    the agent
You also see:
    A door to the hallway (that is open)
[Inventory]
In your inventory, you see:
    an orange
[score: 40]', Action 24: 'mix bowl']
You are now at step 25 and your current observation is:
Yellow and red paint mix to produce orange paint.
[Current room]
This room is called the art studio. In it, you see: 
    a large cupboard. The large cupboard door is closed. 
    a substance called air
    a table. On the table is: a bowl (containing orange paint).
    a wood cup (containing blue paint)
    a wood cup (containing nothing)
    a wood cup (containing nothing)
    the agent
You also see:
    A door to the hallway (that is open)
[Inventory]
In your inventory, you see:
    an orange
[score: 50]
Action templates: [look around, look at OBJ, look in OBJ, inventory, task, teleport to LOC, go to LOC, open OBJ, close OBJ, activate OBJ, deactivate OBJ, pick up OBJ, put down OBJ, move OBJ to OBJ, pour OBJ in OBJ, dunk OBJ in OBJ, mix OBJ, use OBJ on OBJ, connect OBJ to OBJ, disconnect OBJ, eat OBJ, flush OBJ, read OBJ, focus on OBJ, wait, wait1].
Objects you can refer to: [agent, air, art studio, blue paint, bowl, cup containing blue paint, cup containing nothing in art studio, cup on room, cupboard, door, hallway, red paint, table, yellow paint].

Now it's your turn to take an action. First reason step by step (at most two sentences) inside <think> </think>, then give exactly one action inside <action> </action>.<|im_end|>
<|im_start|>assistant
\end{lstlisting}

\begin{center}\scriptsize
\begin{tabularx}{\linewidth}{@{}>{\raggedright\arraybackslash}p{0.42\linewidth} >{\raggedright\arraybackslash}X@{}}
\toprule
Field of the critic input & Source \\
\midrule
\texttt{system message} & fixed (prompts.SYSTEM) \\
\texttt{'Your task is to: \textless{}task\textgreater{}'} & copied from h \\
\texttt{'Rules: ...'} & fixed (prompts.RULES) \\
\texttt{'Prior to this step, you have already taken \textless{}t+1\textgreater{} step(s) ...'} & updated by the program \\
\texttt{'[Observation i: ..., Action i: ...]' (last 5)} & updated by the program: window shifted by one; newest pair = (real current observation, candidate action); copied texts \\
\texttt{'You are now at step \textless{}t+2\textgreater{} and your current observation is:\textbackslash{}n\textless{}o\_hat\textgreater{}'} & step number: program; \textless{}o\_hat\textgreater{} (action feedback, [Current room], [Inventory], [score: S], optional [task completed]/[task failed]): GENERATED by the WM \\
\texttt{'Action templates: [...]'} & fixed (prompts.TEMPLATES) \\
\texttt{'Objects you can refer to: [...]'} & COPIED from h (referable objects before the action; not predicted, not updated) \\
\texttt{instruction sentence} & fixed template \\
\bottomrule
\end{tabularx}
\end{center}

\end{document}